\documentclass[10pt,twocolumn,letterpaper]{article}

\usepackage{cvpr}              
\usepackage[toc,page,header]{appendix}

\usepackage{graphicx}
\usepackage{amsmath}
\usepackage{amssymb}
\usepackage{float}
\usepackage{booktabs}
\usepackage{mathtools}
\usepackage{color}
\usepackage{colortbl}
\usepackage{comment}
\usepackage{verbatim}
\usepackage{tabularx}
\usepackage{multirow}
\usepackage{makecell}
\usepackage{amsmath}
\usepackage{amssymb}
\usepackage{booktabs}
\usepackage{tabularx}
\usepackage{minitoc}
\noptcrule 

\usepackage{tocloft}

\usepackage[accsupp]{axessibility}  
\usepackage[table]{xcolor}  
\usepackage{arydshln}
\usepackage{float}
\usepackage{pifont}    
\usepackage{listings}  
\usepackage{xcolor}    

\definecolor{shcolor}{rgb}{0.1, 0.0, 1.0}

\definecolor{jhcolor}{rgb}{0.0, 0.44, 0.0}

\definecolor{swcolor}{rgb}{0.607,0.149,0.713}

\definecolor{darkred}{rgb}{0.6, 0.1, 0.1}

\definecolor{todocolor}{rgb}{0.5, 0.0, 1.0}

\usepackage[breaklinks,colorlinks]{hyperref}

\usepackage[capitalize]{cleveref}
\crefname{section}{Sec.}{Secs.}
\Crefname{section}{Section}{Sections}
\Crefname{table}{Table}{Tables}
\crefname{table}{Tab.}{Tabs.}

\newcommand{\figref}[1]{Fig. \ref{#1}}
\newcommand{\tabref}[1]{Tab. \ref{#1}}

\newcommand{\rom}[1]{\uppercase\expandafter{\romannumeral #1\relax}}
\newcommand{\hdline}{\hdashline\addlinespace[0.6ex]}

\newcommand*\samethanks[1][\value{footnote}]{\footnotemark[#1]}

\doparttoc 
\faketableofcontents 

\newcommand{\paragrapht}[1]{\noindent\textbf{#1}}  

\def\cvprPaperID{6416} 
\def\confName{CVPR}
\def\confYear{2023}

\begin{document}

\title{LANIT: Language-Driven Image-to-Image Translation for Unlabeled Data}

\author {
    Jihye Park \thanks{Equal contribution} \,\textsuperscript{\rm 1},
    Sunwoo Kim \samethanks   \, \textsuperscript{\rm 1},
    Soohyun Kim \samethanks   \, \textsuperscript{\rm 1},
    Seokju Cho
    \textsuperscript{\rm 1}, \\
    Jaejun Yoo \textsuperscript{\rm 2}, 
    Youngjung Uh \textsuperscript{\rm 3}, 
    Seungryong Kim \thanks{Corresponding author}\, \textsuperscript{\rm 1} \\ \\
    \textsuperscript{\rm 1} Korea University, Seoul, Korea \hspace{5pt}
    \textsuperscript{\rm 2} UNIST, Ulsan, Korea \hspace{5pt}
    \textsuperscript{\rm 3} Yonsei University, Seoul, Korea \\
     {\tt\small \textsuperscript{\rm 1}\{ghp1112,sw-kim,shkim1211,seokju\_cho,seungryong\_kim\}@korea.ac.kr } \\[-3pt]
    {\tt\small \textsuperscript{\rm 2}jaejun.yoo@unist.ac.kr
    \textsuperscript{\rm 3}yj.uh@yonsei.ac.kr}}
\maketitle

\begin{abstract}

Existing techniques for image-to-image translation commonly have suffered from two critical problems: heavy reliance on per-sample domain annotation and/or inability to handle multiple attributes per image.
Recent truly-unsupervised methods adopt clustering approaches to easily provide per-sample one-hot domain labels. However, they cannot account for the real-world setting: one sample may have multiple attributes. In addition, the semantics of the clusters are not easily coupled to human understanding.
To overcome these, we present LANguage-driven Image-to-image Translation model, dubbed LANIT.
We leverage easy-to-obtain candidate attributes given in texts for a dataset: the similarity between images and attributes indicates per-sample domain labels. This formulation naturally enables multi-hot labels so that users can specify the target domain with a set of attributes in language.
To account for the case that the initial prompts are inaccurate, we also present prompt learning. We further present domain regularization loss that enforces translated images to be mapped to the corresponding domain. Experiments on several standard benchmarks demonstrate that LANIT achieves comparable or superior performance to existing models. The code is available at \href{https://github.com/KU-CVLAB/LANIT}{github.com/KU-CVLAB/LANIT}.
\end{abstract}

\section{Introduction}

Unpaired image-to-translation frameworks aim to translate an image from one domain to another~\cite{lee2018diverse,huang2018multimodal,park2020contrastive}. They are typically trained on many images with their respective domain labels. It has been widely studied for a decade in numerous computer vision applications, such as image manipulation~\cite{gonzalez2018image,zhou2021cocosnet,zheng2021spatially}, multi-modal translation~\cite{huang2018multimodal,choi2018stargan,liu2019few,choi2020starganv2,baek2021rethinking}, and instance-aware translation~\cite{bhattacharjee2020dunit,jeong2021memory,kim2022instaformer}.

Conventional methods~\cite{huang2018multimodal,choi2018stargan,choi2020starganv2} required at least \textit{per-sample} domain supervision based on the strict assumption that each image should be described by a one-hot domain label, which is frequently violated; e.g., a face can have multiple attributes. In addition, annotating domain label for each image may become considerably labor-intensive and error-prone, and subjectivity may lead to inconsistency across different annotators.

To mitigate such a heavy reliance on per-sample domain annotation, some recent methods formulate the problem in a few-shot setting~\cite{liu2019few,saito2020coco} or semi-supervised learning setting~\cite{wang2020semi}. More recently, TUNIT~\cite{baek2021rethinking} and Kim \textit{et al.}~\cite{kim2022style} present a framework for jointly learning domain clustering and translation within a unified model, thus enabling a \textit{truly-unsupervised} learning of image-to-image translation. Although these methods ease the burden of per-sample domain annotation and show relatively competitive performance, they still inherit the limitation of using one-hot domain labels to train the image translation model. In addition, these truly-unsupervised learning methods only learn domain clusters that are dominant in the dataset, which sometimes do not reflect semantic meanings. Such limitations make users confused in understanding the semantic meaning of each learned domain cluster.

\begin{figure*}[t]
  \begin{subfigure}{0.32\linewidth}
  \centering
	{\includegraphics[width=1\linewidth]{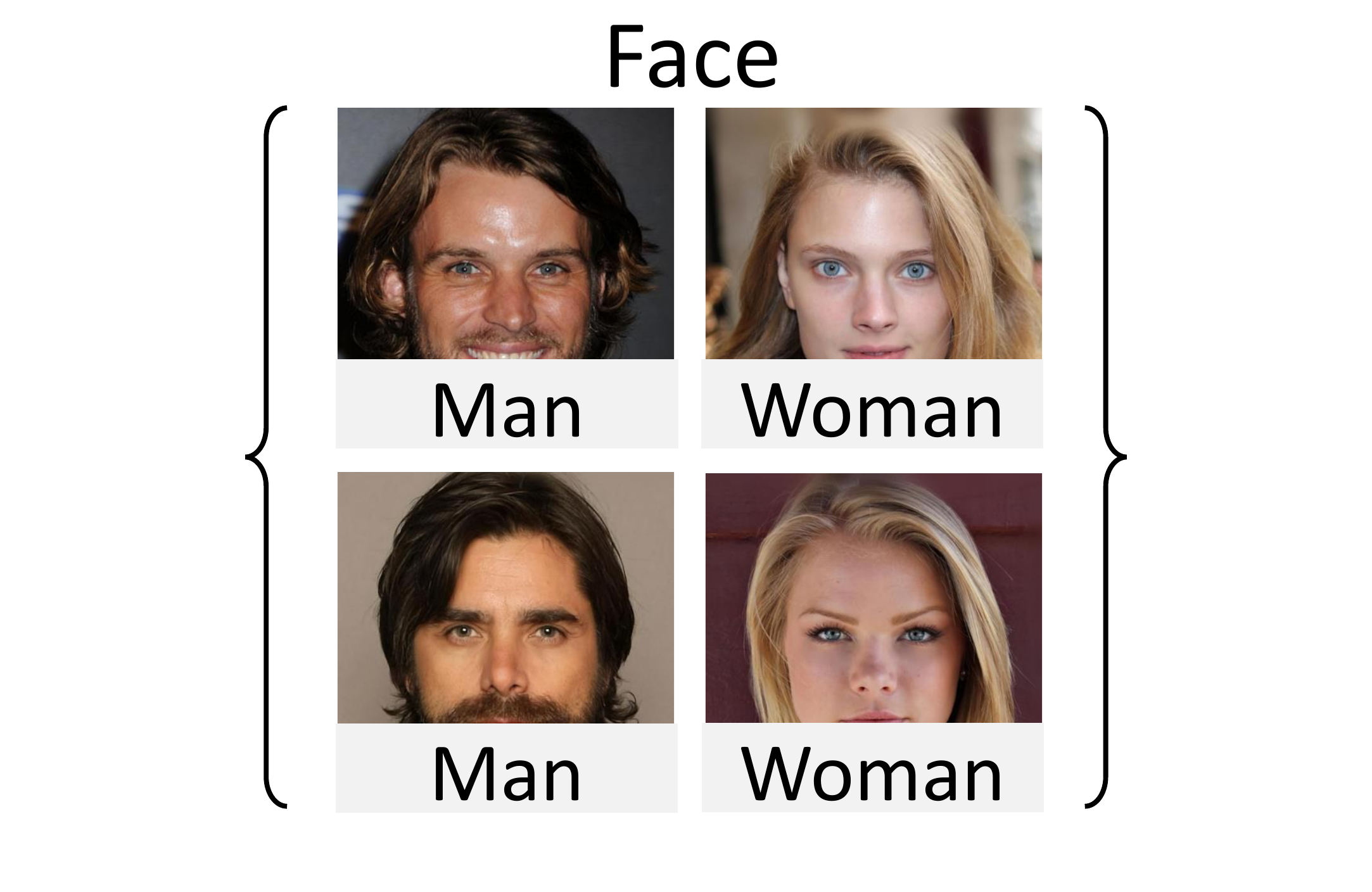}}\hfill
    \caption{Per-sample-level}
  \end{subfigure}
   \begin{subfigure}{0.32\linewidth}
  \centering
	{\includegraphics[width=1\linewidth]{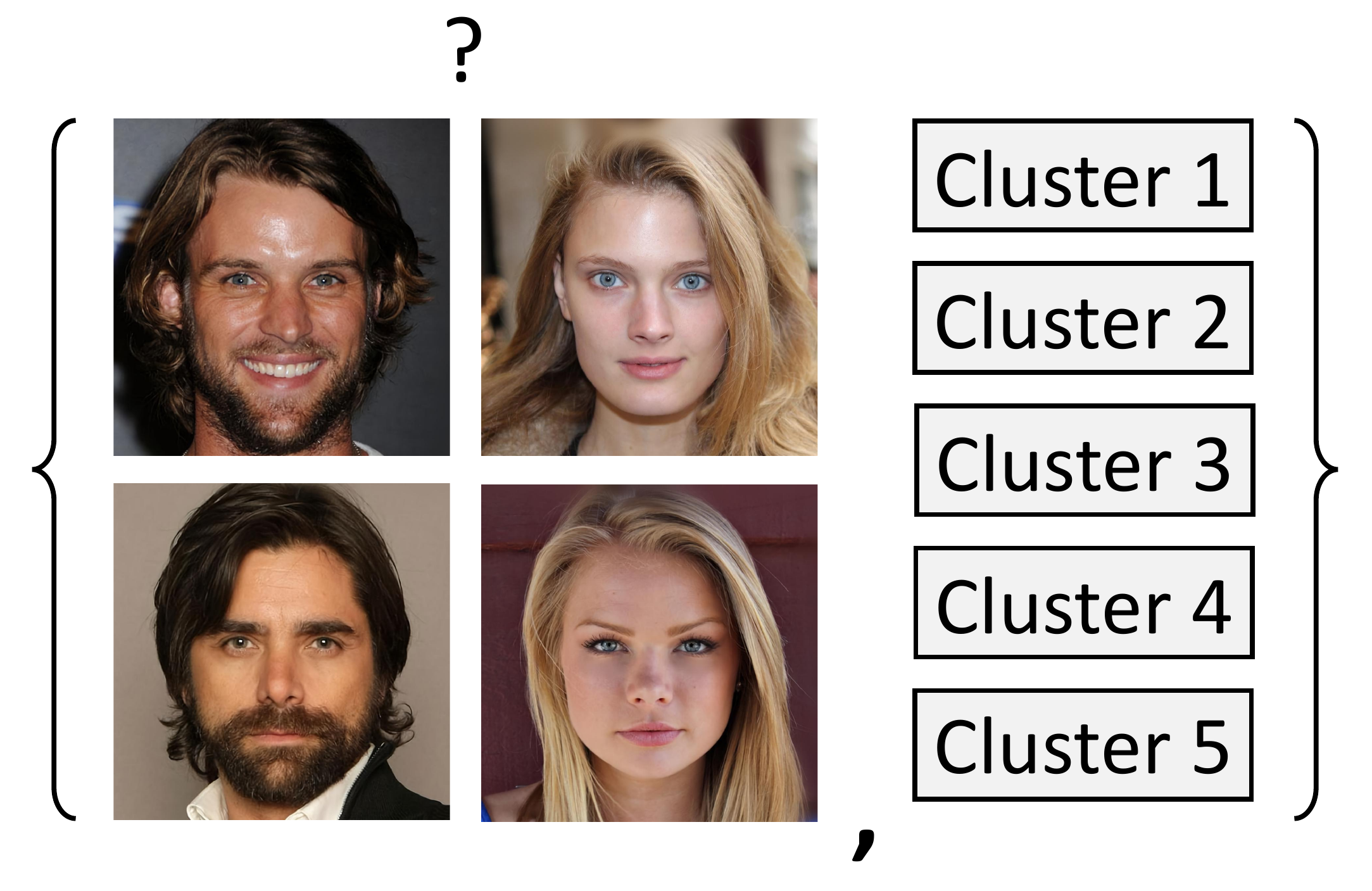}}\hfill
    \caption{Unsupervised}
  \end{subfigure}
    \begin{subfigure}{0.32\linewidth}
  \centering
	{\includegraphics[width=1\linewidth]{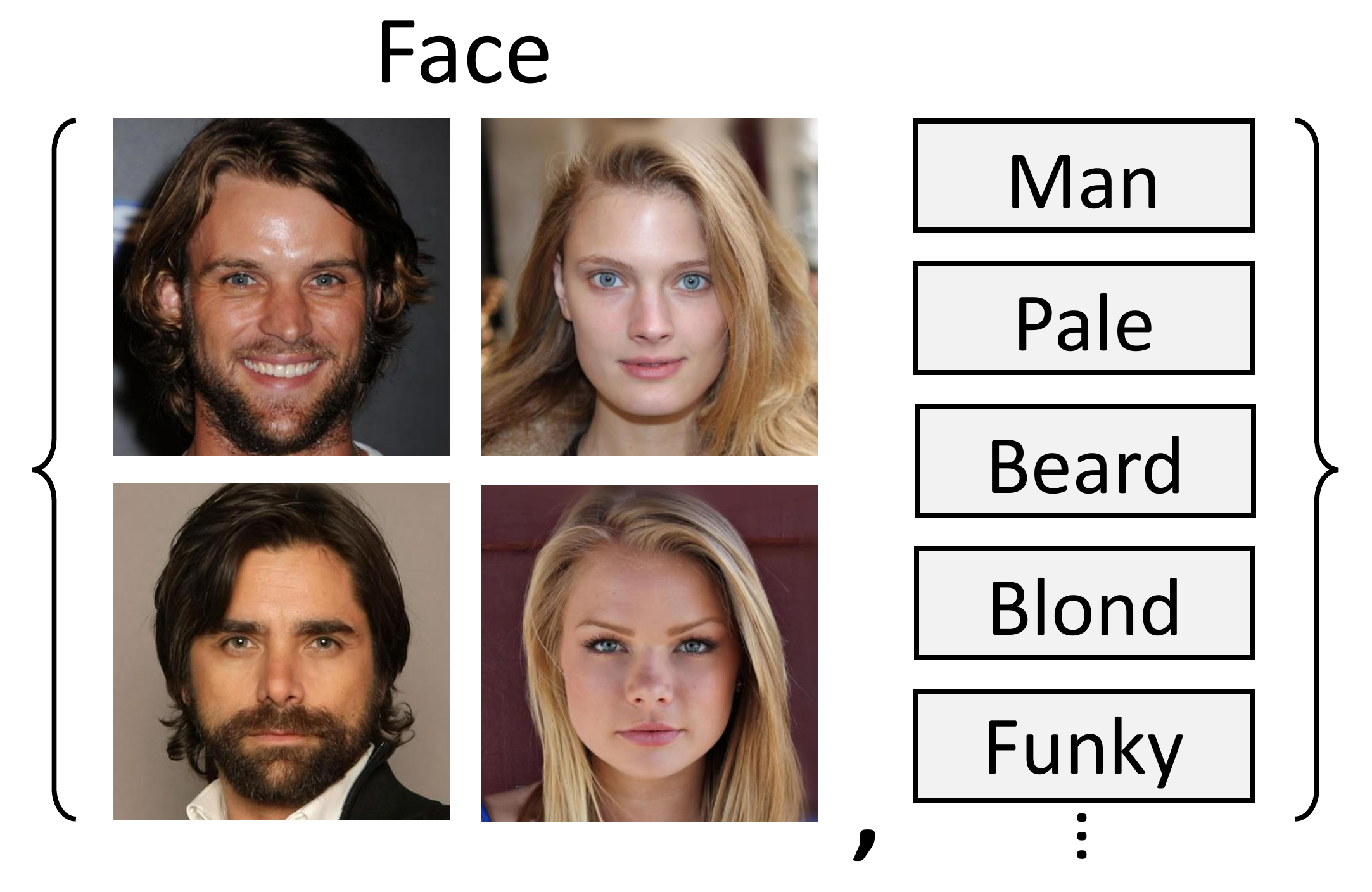}}\hfill
    \caption{Dataset-level (Ours)}
  \end{subfigure}\\
  \vspace{-10pt}
  \caption{\textbf{Levels of supervision.} For unpaired image-to-image translation, (a) conventional methods~\cite{huang2018multimodal,liu2019few,choi2020starganv2,wang2020semi} require at least \textit{per-sample}-level domain supervision, which is often hard to collect. To overcome this, (b) unsupervised learning methods~\cite{baek2021rethinking,kim2022style} learn image translation model using a dataset itself without any supervision, but it shows limited performance and lacks the semantic understanding of each cluster, limiting its applicability. Unlike them, (c) we present a novel framework that requires a dataset with possible textual domain descriptions (i.e., \textit{dataset-level} annotation), which achieves comparable or even better performance than previous methods.} 
\label{fig:level_sup}\vspace{-10pt}
\end{figure*}

\begin{figure}[t]
\centering
\includegraphics[width=1\linewidth]{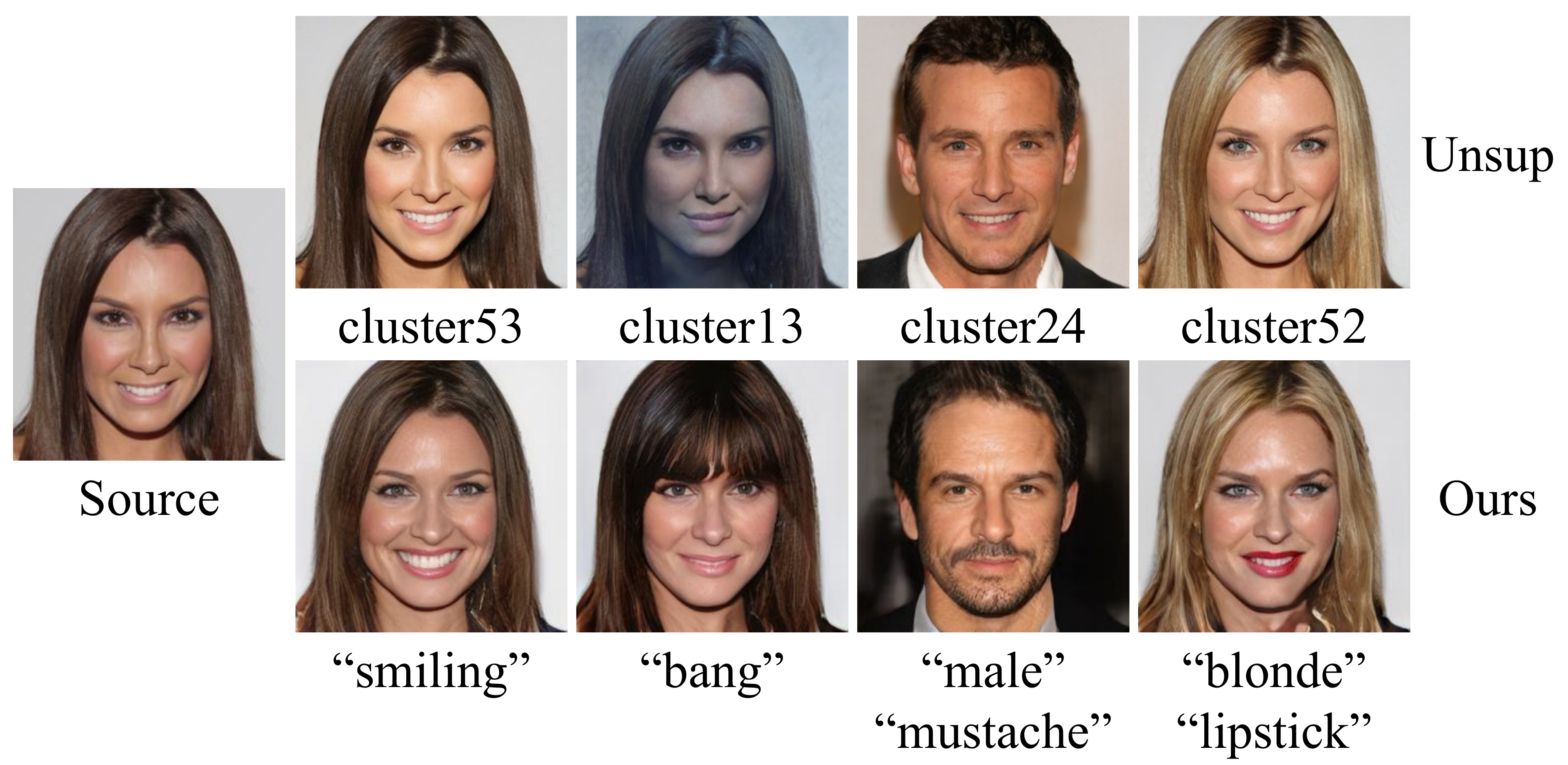}\\
\vspace{-5pt}
    \caption{\textbf{Examples of semantic encoding.} The existing unsupervised method~\cite{baek2021rethinking} allows users to translate one of several clusters. However, the learned clusters lack semantic meaning and sometimes some attributes do not appear in any clusters. Unlike this, our framework can select and train the domains that have explicit semantic meaning, which is more applicable.  
    }
    \label{fig:application}\vspace{-10pt}
\end{figure}

In this paper, to overcome the problems in per-sample-level and unsupervised learning methods, we propose using the \textit{dataset-level} annotation for the first time as exemplified in \figref{fig:level_sup}. In practice, it could be much easier to list candidate textual domain descriptions that describe images in the dataset. Using the candidate attributes given in texts for a dataset, we specify the target domain given a reference image according to its similarity to the attributes in vision-language embedding~\cite{radford2021learning}. Then, a generator translates a source image to the target domain using the target style code of the reference image. As similarities naturally lead to a multi-hot label, we can aggregate the style codes of multiple domains to represent the style code for a joint domain. In addition, to account for the fact that the initial prompts may be sometimes inaccurate, we also propose prompt learning and encourage the translated images to have corresponding labels with our proposed domain regularization loss.

Our proposed framework just requires some textual domain descriptions for defining domains and it can ease the burden on the per-sample annotating process. Thanks to the proposed dataset-level annotation, users can explicitly define the semantic meaning of the target domain with textual descriptions, as exemplified in \figref{fig:application}. We evaluate our method on several standard benchmarks~\cite{bossard2014food,liu2019few,karras2019style}. Experimental results show that the proposed model trained only with dataset-level supervision is comparable or even outperforms the latest methods, including StarGAN2~\cite{choi2020starganv2} trained with per-sample supervision. We also provide intensive ablation studies and user study results.

\section{Related Work}
\paragrapht{Image-to-Image Translation.}
While early methods for image-to-image translation considered a paired setting~\cite{isola2017image}, most recent state-of-the-arts focus on an unpaired setting~\cite{hoffman2018cycada,park2020contrastive,wu2019relgan,zhou2021cocosnet,baek2021rethinking,zheng2021spatially,gabbay2021scaling,liu2021smoothing}. This trend was initiated by CycleGAN~\cite{CycleGAN2017}, of which many variants were proposed, formulating uni-modal models~\cite{liu2017unsupervised,yi2017dualgan,zhang2020cross} and multi-modal models~\cite{huang2018multimodal,lee2018diverse,choi2018stargan,choi2020starganv2}. Although these methods overcome the inherent limitation of the paired setting, they still require at least \textit{per}-\textit{sample} domain annotation, which is often labor intensive or even impossible to achieve in some cases. In addition, this per-sample supervision does not consider that some images may require multi-hot domain labels such as facial attributes, e.g., blond and young.

To mitigate the reliance on per-sample supervision, FUNIT~\cite{liu2019few} and CoCo-FUNIT~\cite{saito2020coco} proposed a few-shot setting, but still required a few annotations. SEMIT~\cite{wang2020semi} utilizes a pseudo labeling technique. S$^3$GAN~\cite{luvcic2019high}, Self-conditioned GAN~\cite{liu2020diverse}, and \cite{bahng2020exploring} adopt clustering methods or a classifier. Recently, TUNIT~\cite{baek2021rethinking} and Kim \textit{et al}.~\cite{kim2022style} have proposed the \textit{truly-unsupervised} frameworks with clustering approach or prototype-based learning, which do not need any per-sample domain supervision. However, their performance was limited in that it is challenging to understand the semantic meaning of each pseudo domain, which in turn limits their applicability. Note that the aforementioned methods also inherit the problem of the one-hot domain labeling assumption.
\begin{figure*}[t]
\centering

\includegraphics[width=1\linewidth]{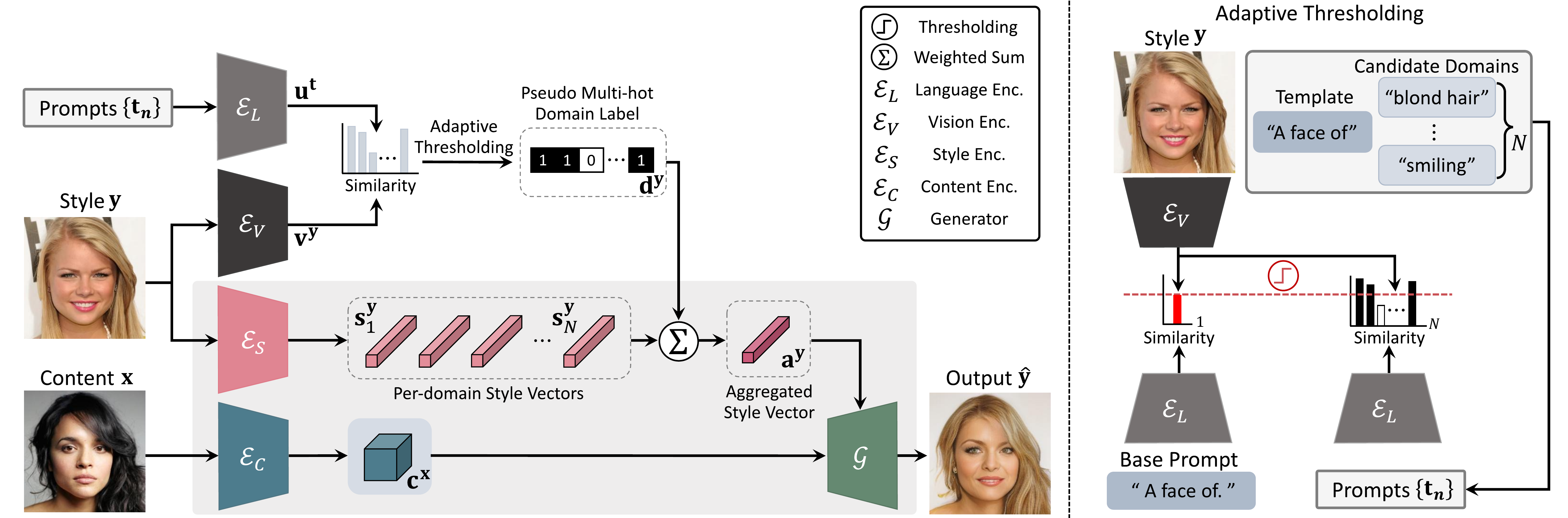}\caption{\textbf{Network configuration.} Our model consists of content and style encoders ($\mathcal{E}_{C}$, $\mathcal{E}_{S}$), vision-language encoders ($\mathcal{E}_{V}$, $\mathcal{E}_{L}$), and generator ($\mathcal{G}$). We extract content and style vectors from content $\mathbf{x}$ and style $\mathbf{y}$, respectively. By leveraging vision-language features and the proposed adaptive thresholding technique, we measure the pseudo domain label $\mathbf{d}^\mathbf{y}$ of $\mathbf{y}$. We generate $\hat{\mathbf{y}}$ with content vector $\mathbf{c}^\mathbf{x}$ and aggregated style vector 
  $\mathbf{a}^\mathbf{y}$ through the generator.}
\label{fig:lunit_arch}\vspace{-10pt}
\end{figure*}

\vspace{3pt}
\paragrapht{Vision-Language Model in Image Manipulation.}
Many large-scale vision-language models~\cite{radford2021learning,jia2021scaling} have been popularly adopted in computer vision problems. Especially, CLIP~\cite{radford2021learning} has been employed in image synthesis and manipulation with GANs~\cite{wei2021hairclip,patashnik2021styleclip,abdal2021clip2stylegan,liu2021fusedream}. For instance, StyleCLIP~\cite{patashnik2021styleclip} and HairCLIP~\cite{wei2021hairclip} manipulate the latent space of  StyleGAN~\cite{Karras2019stylegan2}, through the guidance of CLIP, but they heavily rely on pre-trained StyleGAN. CLIP2StyleGAN extends such an approach in an unsupervised manner that eliminates the need for target style descriptions.

More recently, some works~\cite{kim2021diffusionclip, nichol2021glide, liu2021more, avrahami2022blended, hertz2022prompt, gal2022image, ruiz2022dreambooth} combine a diffusion model~\cite{ho2020denoising} with CLIP~\cite{radford2021learning} for text-driven image manipulation with high fidelity. However, they show limited performance in realistically translating in the real image domain. Furthermore, they rely on pre-trained diffusion model~\cite{ho2020denoising}, and they are only available on text-conditioned manipulation. On the other hand, our LANIT is not dependent on pre-trained StyleGAN~\cite{Karras2019stylegan2} and also learns style manifold from given image dataset and a few numbers of prompts that designate explicit semantics to be learned. 

\vspace{3pt}
\paragrapht{Prompt Learning.}
Since the success of pretrained large-scale language models such as GPT-3~\cite{brown2020language} in NLP, various methods for optimizing the prompt in a text format have been proposed~\cite{petroni2019language,shin2020autoprompt,li2021prefix,liu2021pre}. Inspired by these works, some works~\cite{zhou2021learning,yao2021cpt,ge2022domain,zhou2022conditional,huang2022unsupervised,li2022ordinalclip,ju2022prompting} using CLIP~\cite{radford2021learning} also attempted to optimize the input prompt. For instance, CoCoOp~\cite{zhou2022conditional} showed that using the learned continuous prompt could surpass the manually-designed discrete prompt-based method on zero-shot image classification. However, they require class supervision. 
Recently, several works~\cite{huang2022unsupervised,zhou2022prompt,shu2022test} propose unsupervised prompt tuning. In this work, we explore combining image translation with prompt learning in an unsupervised way for the first time.

\section{Methodology}
\subsection{Motivation and Problem Formulation}
\vspace{-5pt}
We leverage a list of dataset-level domain descriptions in the text form and their similarity with the unlabeled images in CLIP~\cite{radford2021learning} embedding space to measure a pseudo domain label of style image. By doing so, we aim to alleviate difficulty in the labeling process and give explicit semantic meaning in the learned domain for the multi-hot label setting.

One naive approach obtaining the pseudo domain label is by calculating the similarity between image and text features with CLIP~\cite{radford2021learning}. However, we observe that this leads to inaccurate labeling and noisy translation since the image-text feature space in CLIP~\cite{radford2021learning} is not perfectly aligned. To mitigate this issue, we introduce two solutions, an adaptive thresholding with the base prompt for selecting confident pseudo labels, and a prompt learning technique with domain regularization loss for boosting confidence in calculating similarity.

\subsection{Image Translation with Pseudo Domain Label}
\vspace{-5pt}
Our model, consisting of content encoder $\mathcal{E}_{C}$, style encoder $\mathcal{E}_{S}$, mapping encoder $\mathcal{E}_{M}$, and generator $\mathcal{G}$ similar to StarGAN2~\cite{choi2020starganv2}, aims to learn an image-to-image translation with content image $\mathbf{x}$ and style image $\mathbf{y}$ (or between domains $\mathcal{X}$ and $\mathcal{Y}$) to generate a translated image $\hat{\mathbf{y}}$. 

In specific, we first extract the content vector and style vector from $\mathbf{x}\in\mathbb{R}^ {H \times W \times 3}$ and $\mathbf{y}\in\mathbb{R}^ {H \times W \times 3}$ with height $H$ and width $W$ such that $\mathbf{c}^\mathbf{x} = \mathcal{E}_{C}(\mathbf{x})\in \mathbb{R}^ {h_c \times w_c \times c}$ with height $h_c$, width $w_c$ and channel $c$, and $\mathbf{s}^\mathbf{y}_n = \mathcal{E}_{S,n}(\mathbf{y})\in \mathbb{R}^ {1 \times s}$ for $n$-th domain (or ${\tilde{\mathbf{s}}} = \mathcal{E}_{M}(\mathbf{z})$ from a random latent $\mathbf{z}$), respectively. While existing methods, e.g., StarGAN2~\cite{choi2020starganv2}, select a single vector
from $\{\mathbf{s}^\mathbf{y}_1,...,\mathbf{s}^\mathbf{y}_N\}$ of $N$ domains according to ground-truth one-hot domain label, which is inserted to the generator, we aggregate such style vectors $\{\mathbf{s}^\mathbf{y}_1,...,\mathbf{s}^\mathbf{y}_N\}$ with the pseudo domain label $\mathbf{d}^{\mathbf{y}} = [{d}^{\mathbf{y}}_n]_{n=1}^{N}\in \mathbb{R}^ {N \times 1}$ defined as a binary vector, which will be discussed in the following, such that
\begin{equation}
    \mathbf{a}^\mathbf{y} = 
    \frac{1}{M^{\mathbf{y}}} \sum_{n=1}^{N} \mathbf{s}^\mathbf{y}_n {d}^{\mathbf{y}}_n,
\end{equation}
where ${M^{\mathbf{y}}}$ is the number of one values in the pseudo domain label $\mathbf{d}^{\mathbf{y}}$. 
With this aggregated style vector $\mathbf{a}^\mathbf{y}\in \mathbb{R}^ {1 \times s}$ and content vector $\mathbf{c}^\mathbf{x}$, our networks finally generates the output such that $\hat{\mathbf{y}} = \mathcal{G}( \mathbf{c}^\mathbf{x},\mathbf{a}^\mathbf{y})$, containing a style normalization layer. The detailed network architecture is described in the suppl. material. 

\subsection{Language-Driven Domain Labeling}
\vspace{-5pt}
In this section, we explain how to obtain the pseudo multi-hot domain label $\mathbf{d}^{\mathbf{y}}$ for an image ${\mathbf{y}}$ using pretrained vision-language model. Unlike TUNIT~\cite{baek2021rethinking} that clusters the images from the dataset using the visual features only, we use $n$-th domain's textual description as prompt $\mathbf{t}_n\in \mathbb{R}^ {1 \times L}$ which is defined as
\begin{equation}
\mathbf{t}_n = [p_1, p_2, ..., p_{L}, p^\mathrm{domain}_n],
\end{equation}
which consists of one of candidate domains $\{p_{n}^\mathrm{domain}\}_{n=1}^{N}$ and a template $p_l$ ($l \in \{1,...,L\}$) that is shared with all the candidate domains, where $p_l$ is a vector for a word with the same dimension as $p^\mathrm{domain}_n$. For instance, all $p_{l}$ and $p^\mathrm{domain}_n$ can be initialized with a template, e.g., ``a face with'', and given textual domain descriptions, e.g., ``black hair'', respectively. It is utilized to softly cluster the images by measuring the similarity between visual features and language features. In our framework, all the prompts $\mathbf{T} = [ \mathbf{t}_n ]_{n=1}^{N} \in \mathbb{R}^ {N \times (L+1)}$ are initialized by the human-annotated domain descriptions or keywords, namely \textit{dataset-level} supervision.

Based on this definition, we first extract vision and language features $\mathbf{v}^\mathbf{y} = \mathcal{E}_{V}(\mathbf{y}) \in \mathbb{R}^ {1 \times k}$ with $k$ channels, and $[\mathbf{u}^\mathbf{t}_n]_{n=1}^{N} = \mathcal{E}_{L}(\mathbf{T}) \in \mathbb{R}^{N \times k}$ with $k$ channels from style $\mathbf{y}$ and all the prompts $\mathbf{T}$, respectively, by using pretrained vision-language model, i.e., CLIP~\cite{radford2021learning}. Note that any vision-language models with vision and language encoders, \textit{i.e.}, ALIGN~\cite{jia2021scaling}, can be used in our framework. We then measure a similarity $\mathbf{f}^{\mathbf{y}} = [f^{\mathbf{y},\mathbf{t}}_n]_{n=1}^{N} \in \mathbb{R}^ {N \times 1}$, computed as:
\begin{equation}
    f^{\mathbf{y},\mathbf{t}}_n = \bar{\mathbf{v}}^{\mathbf{y}} \cdot \bar{\mathbf{u}}^\mathbf{t}_n,
\end{equation}
where $\bar{\mathbf{v}} = \mathbf{v}/\|\mathbf{v}\|$ and $\bar{\mathbf{u}} = \mathbf{u}/\|\mathbf{u}\|$. 

By utilizing the measured similarity $\mathbf{f}^{\mathbf{y}}$, we can obtain the multi-hot pseudo domain label $\mathbf{d}^{\mathbf{y}}$ with a simple candidate selecting method, e.g., top-$K$ operation or thresholding. However, with top-$K$ operation, we can just select $K$ number of attributes  ignoring the other attributes that would exist in the style image. Moreover, a simple thresholding with a predefined hyperparameter would limit the performance. 
To overcome these, we present an adaptive thresholding technique based on the observation that the more specified text descriptions, \textit{e.g}, ``a face with blonde hair'' make higher similarity with corresponding images than the base prompt as a neutral text description, \textit{e.g.}, ``a face with'', that represent each dataset. We set the base prompt $\textbf{p}$ as the template, meaning that they are updated simultaneously, and thus ${\mathbf{u}}^\mathbf{p} = \mathcal{E}_{L}(\mathbf{p}) \in \mathbb{R}^{1 \times k}$. 
We then define the domain label $\mathbf{d}^\mathbf{y} = [ {d}^{\mathbf{y}}_n]_{n=1}^{N} \in \mathbb{R}^ {N \times 1}$ as follows:
\begin{equation}
\begin{aligned}
  {d}^{\mathbf{y}}_n = \begin{cases}
        1, \quad \mathrm{if} ,\ f^{\mathbf{y},\mathbf{t}}_n > \bar{\mathbf{v}}^{\mathbf{y}} \cdot \bar{\mathbf{u}}^\mathbf{p},  \\
        0, \quad \mathrm{otherwise}.
  \end{cases} 
\end{aligned}
\end{equation}

\begin{figure}[t]
\centering
{\includegraphics[width=1.0\linewidth]{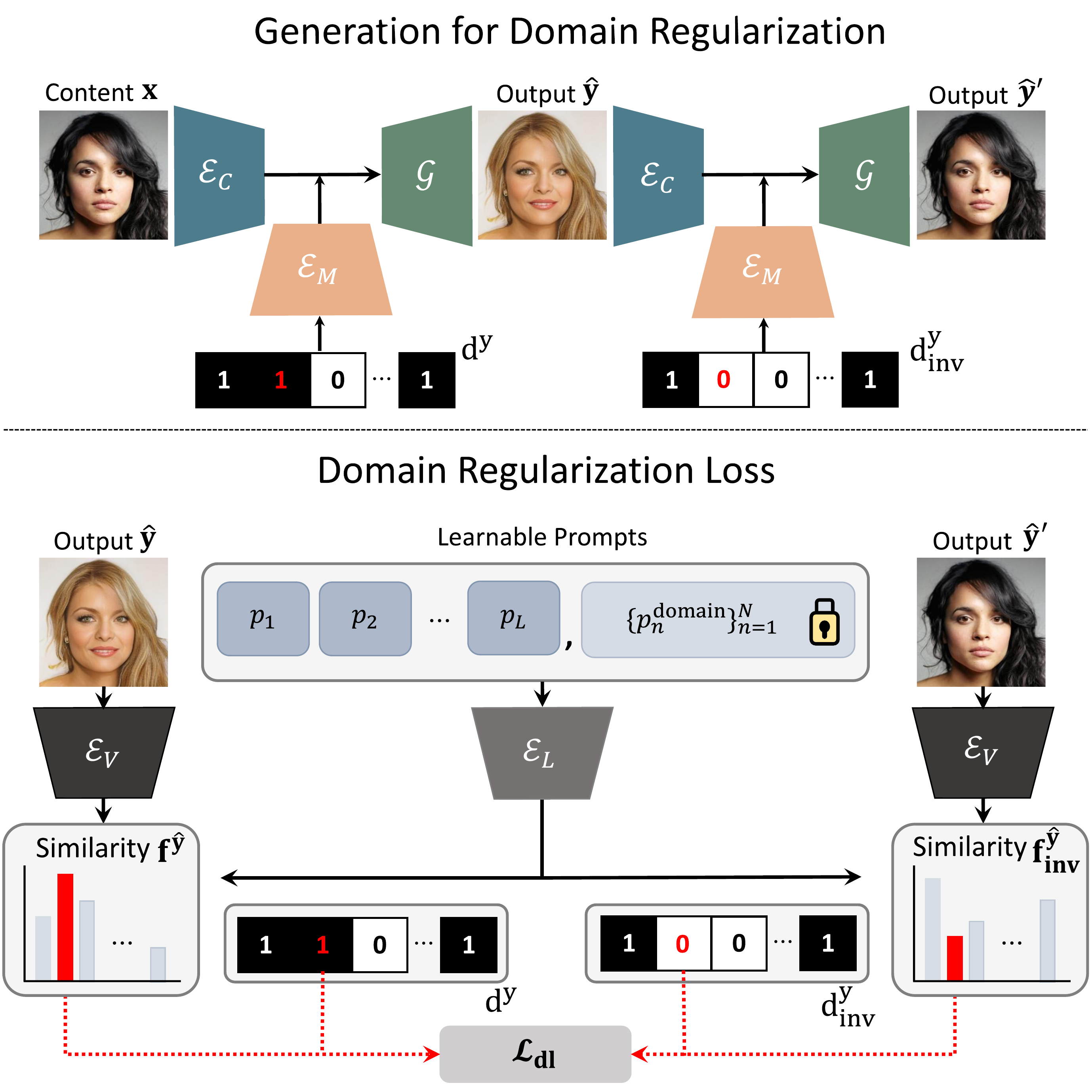}}\hfill\\\vspace{-5pt}
  \caption{\textbf{Illustration of domain regularization loss.} We define a domain regularization loss $\mathcal{L}_\mathrm{dl}$ utilizing two outputs $\hat{\mathbf{y}}$ and $\hat{\mathbf{y}}'$ that have the opposite styles at $n$-th domain, and minimize the loss function not to only learn the optimal prompt but also better learn the translation process.}
  \vspace{-15pt}
\label{fig:prompt_learning}
\end{figure}

\subsection{Prompt Learning}
\vspace{-5pt}
So far, we have discussed the method for language-driven domain labeling and image translation with the pseudo domain label. However, since the pre-defined templates may not be optimal to describe all the images in the dataset, the performance would be noisy and limited. To overcome these, we present a prompt learning technique as shown in \figref{fig:prompt_learning}.

Similar to~\cite{zhou2021learning,zhou2022conditional,huang2022unsupervised}, we set the prompt $\mathbf{t}_n$ as learnable except $p^\mathrm{domain}_n$. In other words, the templates $[p_l]_{l}$ are set as the learnable parameters in $\mathbf{t}_n$.
By allowing the template to be updated, more optimal prompts can be optimized, which in turn boosts the translation performance. Note that the input domain descriptions should be given in our framework, but they can also be obtained by using a predefined dictionary~\cite{radford2021learning,abdal2021clip2stylegan} in a manner that the highly relevant texts to the dataset can be selected based on the similarity between an image and candidate texts. 
\vspace{-5pt}

\subsection{Loss Functions} 
\vspace{-5pt}
\paragrapht{Adversarial Loss.}
We adopt an adversarial loss to guide the translated images $\hat{\mathbf{y}}$ to arrive at the training distribution of the target domain. Following StarGAN2~\cite{choi2020starganv2} and TUNIT~\cite{baek2021rethinking}, we also adopt the multi-domain discriminators, but we use the outputs of multi-domain discriminators weighted by a multi-hot domain label $\mathbf{d}^{\mathbf{y}}$:
\begin{equation}
\begin{split}
     &\mathcal{L}_\mathrm{adv}\\ 
     &= \mathbb{E}_{\mathbf{x}, \mathbf{y}} \sum_{n=1}^{N}\left[\log{\mathcal{D}_{n}(\mathbf{y}})  {{d}^{\mathbf{y}}_n} + {  \log(1 -\mathcal{D}_{n}(\mathcal{G}(\mathbf{x},\mathbf{a}^\mathbf{y}))  {{d}^{\mathbf{y}}_n})} \right],
\end{split}
\end{equation}
where $\mathcal{D}_{n}(\cdot)$ denotes $n$-th discriminator output and ${d}^{\mathbf{y}}_n$ denotes $n$-th element of ${\mathbf{d}^{\mathbf{y}}}$. It should be noted that if we set ${\mathbf{d}^{\mathbf{y}}}$ as a one-hot domain label by ground-truth or pseudo-label, this loss function becomes the same adversarial loss in~\cite{choi2020starganv2,baek2021rethinking}.   

\vspace{3pt}
\paragrapht{Domain Regularization Loss.} 
We further present a domain regularization loss that encourages the networks to generate an image ${\hat{\mathbf{y}}}$ following the domain label of input style image ${\mathbf{y}}$ more precisely. To this end, perhaps one of the simplest functions is to use the dot product between $\mathbf{d}^{{\mathbf{y}}}$ and $\mathbf{f}^{\hat{\mathbf{y}}}$ such that $\mathcal{L} = \mathbf{d}^{{\mathbf{y}}} \cdot \mathbf{f}^{\hat{\mathbf{y}}}$.  Minimizing this, however, can induce erroneous solutions, since considering an inactive domain label, i.e., ${d}^{\mathbf{y}}_n=0$, may degrade the performance further. To overcome this, we present a novel loss function. We first make a domain label pair, $\mathbf{d}^{\mathbf{y}}$ and $\mathbf{d}_\mathrm{inv}^{\mathbf{y}}(n)$, which has same labels with $\mathbf{d}^{\mathbf{y}}$ except the $n$-th label by replacing $n$-th domain label ${d}^{\mathbf{y}}_n$ to $d_{\mathrm{inv},n}^{\mathbf{y}}(n) = 1-{d}^{\mathbf{y}}_n$. We then generate $\hat{\mathbf{y}}' = \mathcal{G}(\mathbf{c}^{\hat{\mathbf{y}}},\mathbf{a}_\mathrm{inv}^{\hat{\mathbf{y}}})$ where $\mathbf{a}_\mathrm{inv}^{\hat{\mathbf{y}}}$ is obtained using $\hat{\mathbf{y}}$ and $\mathbf{d}_\mathrm{inv}^{\mathbf{y}}(n)$. We then encourage not only $\mathbf{f}^{\hat{\mathbf{y}}}$ to follow $\mathbf{d}^{\mathbf{y}}$ at $n$-th label, but also $\mathbf{f}^{\hat{\mathbf{y}}_\mathrm{inv}}$ to follow $\mathbf{d}_\mathrm{inv}^{\mathbf{y}}(n)$ at $n$-th label, such that 
\begin{equation}
\label{eq:loss_dc}
\mathcal{L}_\mathrm{dl} =
\mathcal{H}(d^{\mathbf{y}}_n,f^{\hat{\mathbf{y}}}_{n})
+ \mathcal{H}(d_{\mathrm{inv},n}^{\mathbf{y}}(n), f^{\hat{\mathbf{y}}}_{\mathrm{inv},n}),
\end{equation}
where $\mathcal{H}(\cdot,\cdot)$ denotes a binary cross-entropy.

\vspace{3pt}
\paragrapht{Cycle-Consistency Loss.}
To make the translated image similar to its original image, which also regularizes the ill-posed translation problem, we adopt a cycle-consistency loss such that
    \begin{equation}
        \mathcal{L}_\mathrm{cyc} =
        \mathbb{E}_{\mathbf{x}, \mathbf{y}} \left[ {\|\mathbf{x} - \mathcal{G}((\hat{\mathbf{y}}),
        \mathbf{a}^\mathbf{x})\|}_{1} \right],
    \end{equation}
where $\mathbf{c}^{\hat{\mathbf{y}}}$ denotes the content from $\hat{\mathbf{y}}$, and $\mathbf{a}^\mathbf{x}$ denotes the style vector of input $\mathbf{x}$. By encouraging the generator $\mathcal{G}$ to reconstruct the input image $\mathbf{x}$ with estimated style vector $\mathbf{a}^\mathbf{x}$, $\mathcal{G}$ learns to preserve the original characteristics of $\mathbf{x}$. 

\vspace{3pt}
\paragrapht{Style Reconstruction Loss.} 
While the generator is able to synthesize realistic images with the losses above, the synthesized results are not guaranteed to be style-consistent with $\mathbf{y}$. In order to better learn style representation, we compute $l$-1 loss between the style vector from the translated image and style image such that
\begin{equation}
    \mathcal{L}_\mathrm{sty} =
    \mathbb{E}_\mathbf{x,y}[\|{\mathbf{s^{y}}-\,\mathcal{E}_{S}(\hat{\mathbf{y}})}\|_{1}].
\end{equation}

\vspace{3pt}
\paragrapht{Style Diversification Loss.}
To learn the generator to produce diverse images, we employ a diversity sensitive loss~\cite{mao2019mode, dsganICLR2019} defined as:
\begin{equation}
    \mathcal{L}_\mathrm{ds} =
    \mathbb{E}_\mathbf{x,y}[\|\mathcal{G}(\mathbf{x}, \mathcal{E}_{M}(\mathbf{z}_{1})) - \mathcal{G}(\mathbf{x}, \mathcal{E}_{M}(\mathbf{z}_{2}))\|_{1}], 
\end{equation}
where $\mathbf{z}_1$ and $\mathbf{z}_2$ are random latent vectors from Gaussian distribution. Note that we only adopt the diversification loss using random vectors, not the style vector $\mathbf{a}^\mathbf{y}$ from an image. Also, we utilize the domain label $\mathbf{d}^{\mathbf{y}}$ from the style image $\mathbf{y}$ as a pseudo-label for the latent vector.

\vspace{3pt}
\paragrapht{Overall Objective.}
Full loss functions are as follows: 
\begin{equation}
\begin{split}
    \mathcal{L}_\mathrm{total} = &\lambda_\mathrm{adv}\mathcal{L}_\mathrm{adv} +  \lambda_\mathrm{dl}\mathcal{L}_\mathrm{dl} +
    \lambda_\mathrm{cyc}\mathcal{L}_\mathrm{cyc} \\
    &+\lambda_\mathrm{sty}\mathcal{L}_\mathrm{sty} -
    \lambda_\mathrm{ds}\mathcal{L}_\mathrm{ds},
\end{split}
\end{equation}
where $\lambda_\mathrm{adv}$, $\lambda_\mathrm{dl}$, $\lambda_\mathrm{cyc}$, $\lambda_\mathrm{sty}$, and $\lambda_\mathrm{ds}$ are hyper-parameters.

\begin{figure*}[t]
\centering
  \begin{subfigure}{0.12\linewidth}
  \centering
	{\includegraphics[width=1\linewidth]{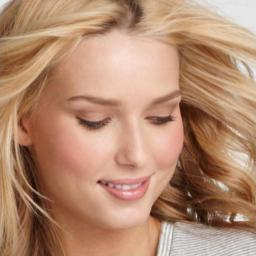}}\hfill
    \end{subfigure}
  \begin{subfigure}{0.12\linewidth}
  \centering
	{\includegraphics[width=1\linewidth]{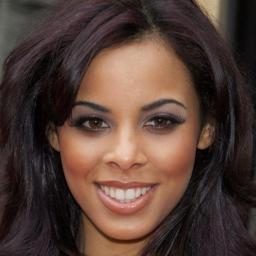}}\hfill
    \end{subfigure}
  \begin{subfigure}{0.12\linewidth}
  \centering
	{\includegraphics[width=1\linewidth]{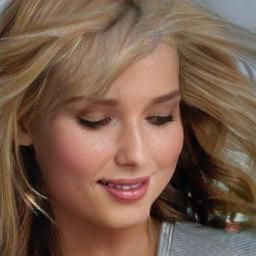}}\hfill
    \end{subfigure}
  \begin{subfigure}{0.12\linewidth}
  \centering
	{\includegraphics[width=1\linewidth]{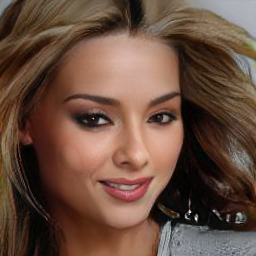}}\hfill
    \end{subfigure}
  \begin{subfigure}{0.12\linewidth}
  \centering
	{\includegraphics[width=1\linewidth]{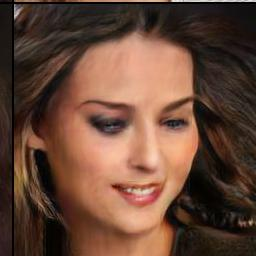}}\hfill
    \end{subfigure}
  \begin{subfigure}{0.12\linewidth}
  \centering
	{\includegraphics[width=1\linewidth]{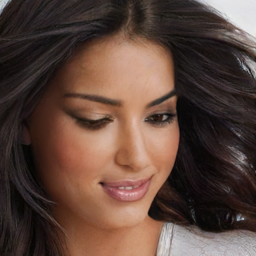}}\hfill
    \end{subfigure}
  \begin{subfigure}{0.12\linewidth}
  \centering
	{\includegraphics[width=1\linewidth]{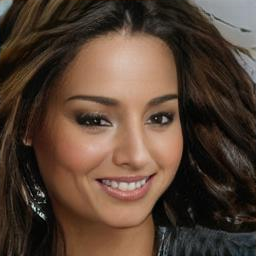}}\hfill\end{subfigure} \\\vspace{0pt}

  \begin{subfigure}{0.12\linewidth}
  \centering
	{\includegraphics[width=1\linewidth]{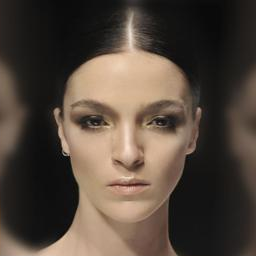}}\hfill
    \end{subfigure}
  \begin{subfigure}{0.12\linewidth}
  \centering
	{\includegraphics[width=1\linewidth]{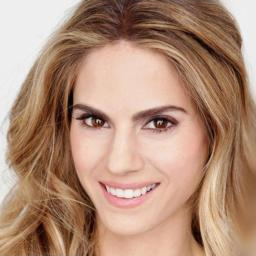}}\hfill
    \end{subfigure}
  \begin{subfigure}{0.12\linewidth}
  \centering
	{\includegraphics[width=1\linewidth]{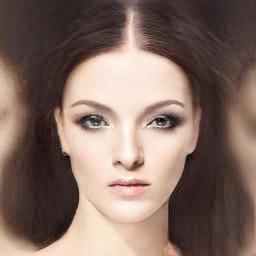}}\hfill
    \end{subfigure}
  \begin{subfigure}{0.12\linewidth}
  \centering
	{\includegraphics[width=1\linewidth]{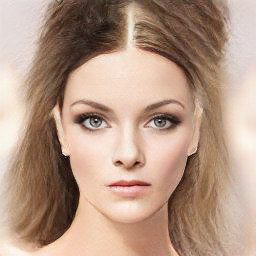}}\hfill
    \end{subfigure}
  \begin{subfigure}{0.12\linewidth}
  \centering
	{\includegraphics[width=1\linewidth]{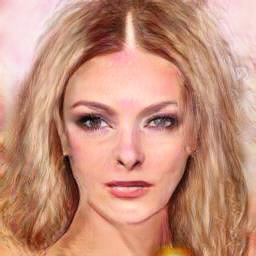}}\hfill
    \end{subfigure}
  \begin{subfigure}{0.12\linewidth}
  \centering
	{\includegraphics[width=1\linewidth]{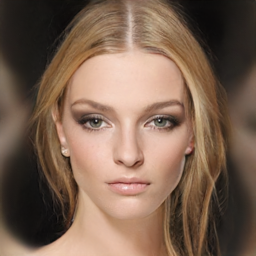}}\hfill
    \end{subfigure}
  \begin{subfigure}{0.12\linewidth}
  \centering
	{\includegraphics[width=1\linewidth]{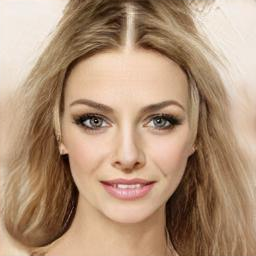}}\hfill\end{subfigure} \\\vspace{0pt}

  \begin{subfigure}{0.12\linewidth}
	{\includegraphics[width=1\linewidth]{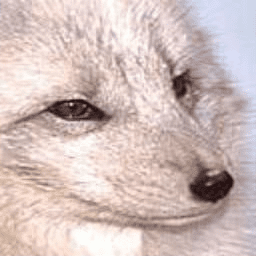}}\hfill
\end{subfigure}
  \begin{subfigure}{0.12\linewidth}
	{\includegraphics[width=1\linewidth]{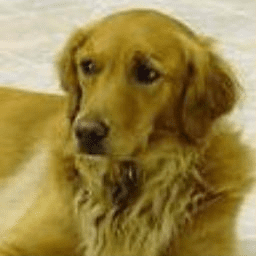}}\hfill
\end{subfigure}
  \begin{subfigure}{0.12\linewidth}
  \centering
	{\includegraphics[width=1\linewidth]{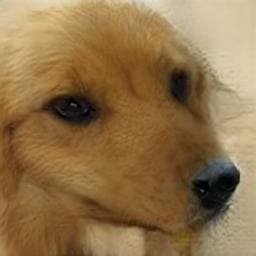}}\hfill
\end{subfigure}
  \begin{subfigure}{0.12\linewidth}
  \centering
	{\includegraphics[width=1\linewidth]{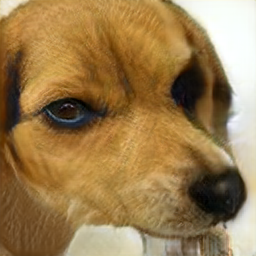}}\hfill
\end{subfigure}
  \begin{subfigure}{0.12\linewidth}
  \centering
	{\includegraphics[width=1\linewidth]{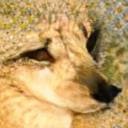}}\hfill
\end{subfigure}
  \begin{subfigure}{0.12\linewidth}
  \centering
	{\includegraphics[width=1\linewidth]{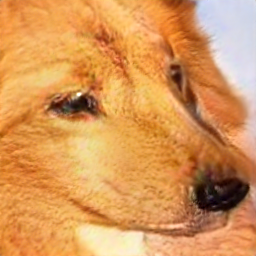}}\hfill
\end{subfigure}
  \begin{subfigure}{0.12\linewidth}
  \centering
	{\includegraphics[width=1\linewidth]{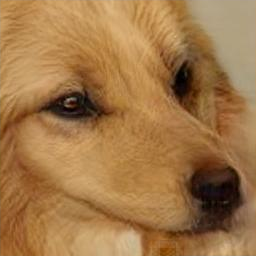}}\hfill\end{subfigure}\\\vspace{0pt}
	
  \begin{subfigure}{0.12\linewidth}
	{\includegraphics[width=1\linewidth]{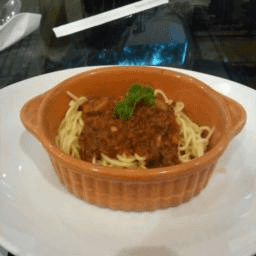}}\caption{Content}\hfill
\end{subfigure}
  \begin{subfigure}{0.12\linewidth}
	{\includegraphics[width=1\linewidth]{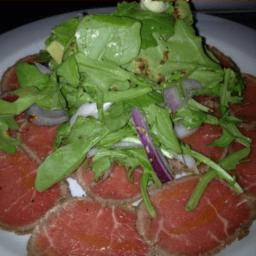}}\caption{Style}\hfill
\end{subfigure}
  \begin{subfigure}{0.12\linewidth}
  \centering
	{\includegraphics[width=1\linewidth]{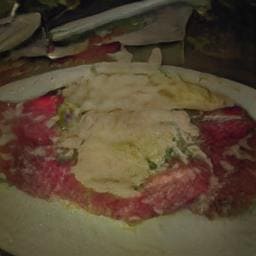}}\caption{StarGANv2~\cite{choi2020starganv2}}\hfill
\end{subfigure}
  \begin{subfigure}{0.12\linewidth}
  \centering
	{\includegraphics[width=1\linewidth]{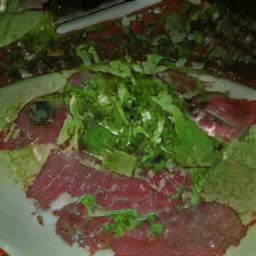}}\caption{Smoothing~\cite{liu2021smoothing}}\hfill
\end{subfigure}
  \begin{subfigure}{0.12\linewidth}
  \centering
	{\includegraphics[width=1\linewidth]{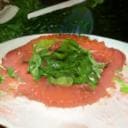}}\caption{TUNIT~\cite{baek2021rethinking}}\hfill
\end{subfigure}
  \begin{subfigure}{0.12\linewidth}
  \centering
	{\includegraphics[width=1\linewidth]{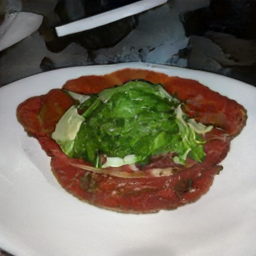}}\caption{Kim \textit{et al}.~\cite{kim2022style}}\hfill
\end{subfigure}
  \begin{subfigure}{0.12\linewidth}
  \centering
	{\includegraphics[width=1\linewidth]{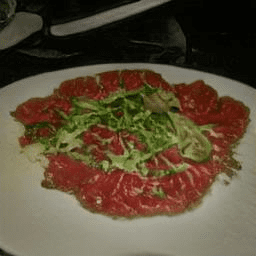}}\caption{Ours}\hfill\end{subfigure}\\

	\vspace{-15pt}
	\caption{\textbf{Qualitative comparison.}}
	\label{fig:main_qual}\vspace{-5pt}
\end{figure*}

\begin{table*}
    \centering
    \begin{tabular}{l c c c c c c }\toprule
         &\multicolumn{2}{c}{CelebA-HQ~\cite{liu2015deep}}
         &\multicolumn{2}{c}{AnimalFaces-10~\cite{liu2019few}}
         &\multicolumn{2}{c}{Food-10~\cite{bossard2014food}}\\
         \cmidrule(lr){2-3} \cmidrule(lr){4-5} \cmidrule(lr){6-7}
        Method &mFID $\downarrow$  &D\&C $\uparrow$ &mFID $\downarrow$ &D\&C $\uparrow$ &mFID $\downarrow$ &D\&C $\uparrow$ \\\midrule 
        StarGAN2~\cite{choi2020starganv2} (sup.) &32.16 &1.22 / \textbf{0.44} &\textbf{33.67} &\textbf{1.54} / \textbf{0.91} &65.03 &1.09 / 0.76 \\
        Smoothing~\cite{liu2021smoothing} (sup.) &35.93 &\textbf{1.25} / 0.43 &38.93 &0.97 / 0.75 &61.13 &0.96 / 0.68 \\ 
        TUNIT~\cite{baek2021rethinking} (unsup.) &61.29 &0.24 / 0.13 &47.70 &1.04 / 0.81 &52.20 &1.08 / \textbf{0.88}  \\
        Kim \textit{et al}.~\cite{kim2022style} (unsup.) &41.33 &0.60 / 0.24 &36.83 & 1.06 / 0.82 &49.34 &1.06 / 0.80  \\ \rowcolor{red!7} 
        LANIT &\textbf{27.96} &0.91 / 0.34 & 34.11 &1.46 / 0.89 &\textbf{48.08} &\textbf{1.24} / 0.86 \\\bottomrule
    \end{tabular}
    \vspace{-1mm}
    \caption{\textbf{Quantitative comparison on CelebA-HQ~\cite{liu2015deep}, Animal Faces-10~\cite{liu2019few} and Food-10~\cite{bossard2014food}} The configurations of StarGAN2 and Smoothing use ground-truth domain labels while TUNIT and Kim \textit{et al}. use pseudo-labels generated from each image. Our LANIT uses only textual domain descriptions.}
    \label{tab:quan_table}
\end{table*}

\section{Experiments}

\paragrapht{Implementation Details}
We adopted StarGAN2~\cite{choi2020starganv2} as the baseline architecture for content, style, mapping encoders, generator, and discriminator. Additional implementation details will be explained in the supplementary material.
\vspace{3pt}

\paragrapht{Setting up Templates and Candidate Domains.}
We set the candidate domains by randomly sampling among the domain or attribute names in each dataset. For a fair comparison, we experiment three times with different domains respectively to measure quantitative results. The selected candidate domains are shown in the supplementary materials. We refer to the templates from~\cite{bar2022text2live,gal2022image} and slightly modify them to align for each dataset. To improve the accuracy of pseudo labels, we also adopt template augmentation in our baseline model.
 
\vspace{3pt}
\paragrapht{Datasets.}
We conduct the experiments on four standard datasets, including Animal Faces-10, Food-10, CelebA-HQ, and LHQ~\cite{bossard2014food,liu2019few,liu2015deep,skorokhodov2021aligning}. In particular, we consider 10 textual domain descriptions among Animal Faces-149 and Food-101, respectively, selected in TUNIT~\cite{baek2021rethinking}, to define the initial prompt of our framework. For CelebA-HQ, we obtain 40 attributes, and sample 10 attributes for three times to define the initial prompt and report the average results. Since LHQ dataset~\cite{skorokhodov2021aligning} does not provide any attributes, we sampled various target domain descriptions and selected 10 domains for experiments.

\vspace{3pt}
\paragrapht{Evaluation Metrics.}
In our experiments, we adopt three quantitative evaluation metrics. First, the mean of class-wise Fr\'enchet Inception Distance (mFID)~\cite{heusel2017gans} is used to evaluate how the translated image reflects the input style. In addition, we use Density and Coverage (D\&C)~\cite{naeem2020reliable}, which measure the fidelity and diversity of the translated images, respectively. The lower the mFID value, the better quality of the image. As fidelity and diversity get better, the D\&C scores are higher or closer to 1. We also measure Acc., namely classification accuracy, between the domain labels with the highest probability and ground-truth domain labels. 

\subsection{Comparisons for Other Models}
\vspace{-5pt}
\paragrapht{Quantitative Results.}
We first report the quantitative comparison of our model with StarGAN2~\cite{choi2020starganv2}, Smoothing~\cite{liu2021smoothing} (supervised), TUNIT~\cite{baek2021rethinking} and Kim \textit{et al.}~\cite{kim2022style} (unsupervised) on CelebA-HQ, AnimalFaces-10 and Food-10 datasets. Note that  StarGAN2 and Smoothing are trained using 10 classes of ground-truth \textit{per-sample} domain labels. Our and truly-unsupervised models use GT labels for calculating quantitative metrics. \tabref{tab:quan_table} shows the quantitative comparison in terms of mFID, D\&C. metrics. LANIT shows the best mFID and D\&C scores, proving that our model outperforms the state-of-the-art methods in terms of image quality, fidelity and diversity. We also observe that LANIT consistently outperforms unsupervised models by a large margin on all three datasets on Acc. metric. Although our LANIT only utilizes unlabeled datasets with \textit{dataset-level} text supervision, it shows competitive results against StarGAN2~\cite{choi2020starganv2} and Smoothing~\cite{liu2021smoothing}, which require ground-truth per-sample domain labels. Note that we do not provide quantitative results on LHQ since it does not provide ground-truth labels.

\begin{figure}
\centering
\newcolumntype{M}[1]{>{\centering\arraybackslash}m{#1}}
\setlength{\tabcolsep}{1pt} 
\renewcommand{\arraystretch}{2} 
\small
\resizebox{\linewidth}{!}{
\begin{tabular}
{M{0.25\linewidth}M{0.006\linewidth}M{0.25\linewidth}M{0.25\linewidth}M{0.25\linewidth}}

Content && \makecell[c]{``mountain'' \\ ``waterfall'' \\ ``summer''} & \makecell[c]{``field'' \\ ``sunset'' \\ ``cloudy''} & \makecell[c]{``ocean'' \\ ``sunset'' \\ ``cloudy''}\\

\hline \\[-20pt]
{\includegraphics[width=1\linewidth]{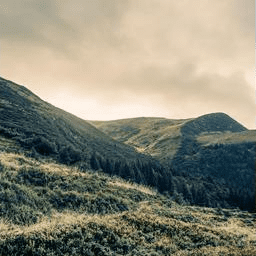}}\hfill & &
{\includegraphics[width=1\linewidth]{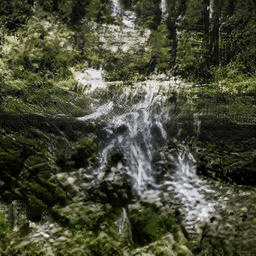}}\hfill  &
{\includegraphics[width=1\linewidth]{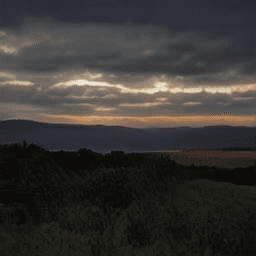}}\hfill  &
{\includegraphics[width=1\linewidth]{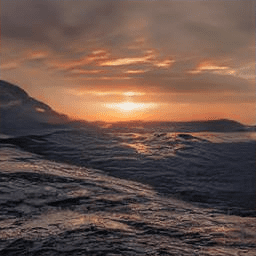}}\hfill  \\[-3pt]
{\includegraphics[width=1\linewidth]{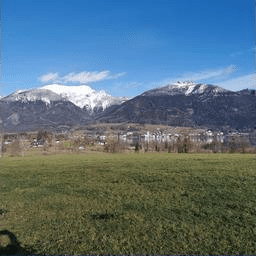}}\hfill & &
{\includegraphics[width=1\linewidth]{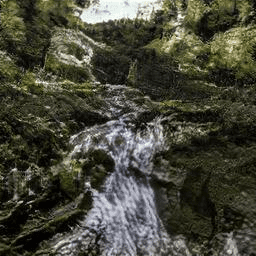}}\hfill  &
{\includegraphics[width=1\linewidth]{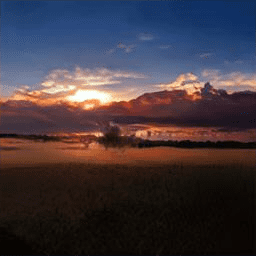}}\hfill  &
{\includegraphics[width=1\linewidth]{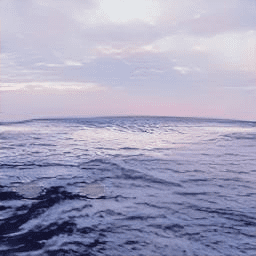}}\hfill   \\
\end{tabular}}
 \vspace{-10pt}  
    \caption{\textbf{Latent-guided diverse image synthesis results by our LANIT on LHQ~\cite{skorokhodov2021aligning}.} Target domain text descriptions are shown above each image.}
    \label{fig:latent}\vspace{-10pt}  
\end{figure}

\vspace{3pt}
\paragrapht{Qualitative Results.}
We show visual results for \tabref{tab:quan_table} in \figref{fig:main_qual}. We observe that the quality differences between our LANIT and TUNIT~\cite{baek2021rethinking} or Kim \textit{et al.}~\cite{kim2022style} are significant. The images generated by the other methods tend to contain artifacts and show limited performance in reflecting diverse attributes from style images, while our method yields stable performance in terms of style representation and semantic consistency. Especially, we show that our model is highly capable of capturing multiple subtle attributes, such as ``{smiling}'' or ``{straight hair}'', while TUNIT~\cite{baek2021rethinking} and Kim \textit{et al.}~\cite{kim2022style} often fail. Interestingly, our LANIT shows competitive results compared to StarGAN2~\cite{choi2020starganv2} and Smoothing~\cite{liu2021smoothing} which are supervised methods, even with better style preservation. We also show domain-to-domain diverse image synthesis results in \figref{fig:latent}, which is guided by target domain labels. Thanks to our proposed prompt learning, the mapping network and style encoder can faithfully produce the style vectors reflecting the target multiple styles.

\begin{table}
    \centering
    \resizebox{\linewidth}{!}{
    \begin{tabular}{l l c c c c}\toprule
         &&\multicolumn{2}{c}{AnimalFaces-10~\cite{liu2019few}}
         &\multicolumn{2}{c}{CelebA-HQ~\cite{liu2015deep}}\\
         \cmidrule(lr){3-4} \cmidrule(lr){5-6}
        N & Method &mFID $\downarrow$ &D\&C $\uparrow$ &mFID $\downarrow$ &D\&C $\uparrow$  \\\midrule 
        \multirow{2}{*}{$4$} & TUNIT &77.7 &0.88 / 0.74 &61.5 &0.24 / 0.12 \\  
        &  \cellcolor{red!7}LANIT  &\cellcolor{red!7}71.6 &\cellcolor{red!7}1.35 / 0.46 & \cellcolor{red!7}49.3 &\cellcolor{red!7}0.33 / 0.14  \\\hdline
        \multirow{2}{*}{$7$}& TUNIT &62.7 &1.02 / 0.73 &54.7 &0.33 / 0.16 \\ 
        &  \cellcolor{red!7}LANIT &\cellcolor{red!7}49.9 &\cellcolor{red!7}1.47 / 0.66 &\cellcolor{red!7} 43.2 &\cellcolor{red!7}0.44 / 0.19  \\\hdline
        \multirow{2}{*}{$10$}& TUNIT &47.7 &1.04 / 0.81 &61.3 &0.24 / 0.13 \\ 
        & \cellcolor{red!7}LANIT &\cellcolor{red!7} \textbf{34.1} &\cellcolor{red!7}1.46 / \textbf{0.89} &\cellcolor{red!7} \textbf{27.9} &\cellcolor{red!7}\textbf{0.91} / \textbf{0.34}  \\\hdline
        \multirow{2}{*}{$13$}& TUNIT &56.8 &0.99 /0.72 &98.9 &0.08 / 0.03  \\ 
        &\cellcolor{red!7}  LANIT &\cellcolor{red!7}30.1 &\cellcolor{red!7}1.43 / 0.85 &\cellcolor{red!7} 34.8 &\cellcolor{red!7}0.58 / 0.21  \\\hdline
        \multirow{2}{*}{$16$}& TUNIT &54.1 &1.09 / 0.78 & 127.7 &0.04 / 0.02  \\
        & \cellcolor{red!7}LANIT &\cellcolor{red!7}35.8 &\cellcolor{red!7}\textbf{1.49} / 0.82 &\cellcolor{red!7} \textbf{27.9} &\cellcolor{red!7}0.76 / 0.23 \\\bottomrule
        
    \end{tabular}}
    \vspace{-1mm}
    \caption{\textbf{Quantitative comparison of LANIT with TUNIT~\cite{baek2021rethinking} by varying the number of domains.}}
    \label{tab:k_comparison}\vspace{-10pt}
\end{table}

\begin{figure}[t!]
\centering
\newcolumntype{M}[1]{>{\centering\arraybackslash}m{#1}}
\setlength{\tabcolsep}{1pt} 
\renewcommand{\arraystretch}{2} 
\footnotesize
\begin{tabular}
{M{0.17\linewidth}M{0.01\linewidth}|M{0.02\linewidth}M{0.17\linewidth}M{0.17\linewidth}M{0.17\linewidth}M{0.17\linewidth}}
Content &&& ``blond hair'' & \makecell[c]{``blond hair''\\``bang''} & \makecell[c]{``blond hair''\\ ``bang'' \\ ``smiling''} & \makecell[c]{``blond hair''\\ ``bang'' \\ ``smiling'' \\ ``lipstick''}\\\hline \\[-15pt]
{\includegraphics[width=\linewidth]{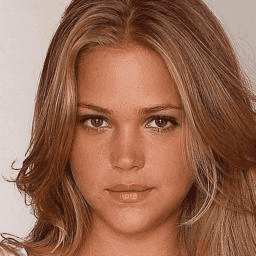}}\hfill &&&
{\includegraphics[width=\linewidth]{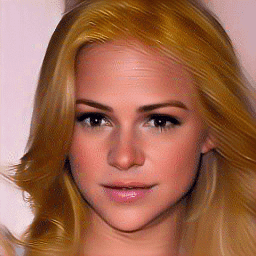}}\hfill  &
{\includegraphics[width=\linewidth]{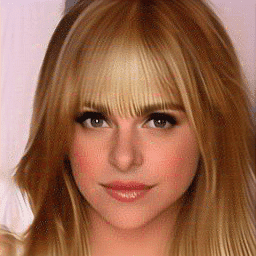}}\hfill  &
{\includegraphics[width=\linewidth]{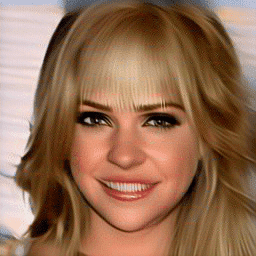}}\hfill &
{\includegraphics[width=\linewidth]{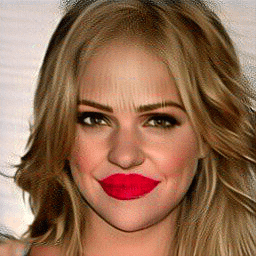}}\hfill\\[-3pt]

&&&
{\includegraphics[width=\linewidth]{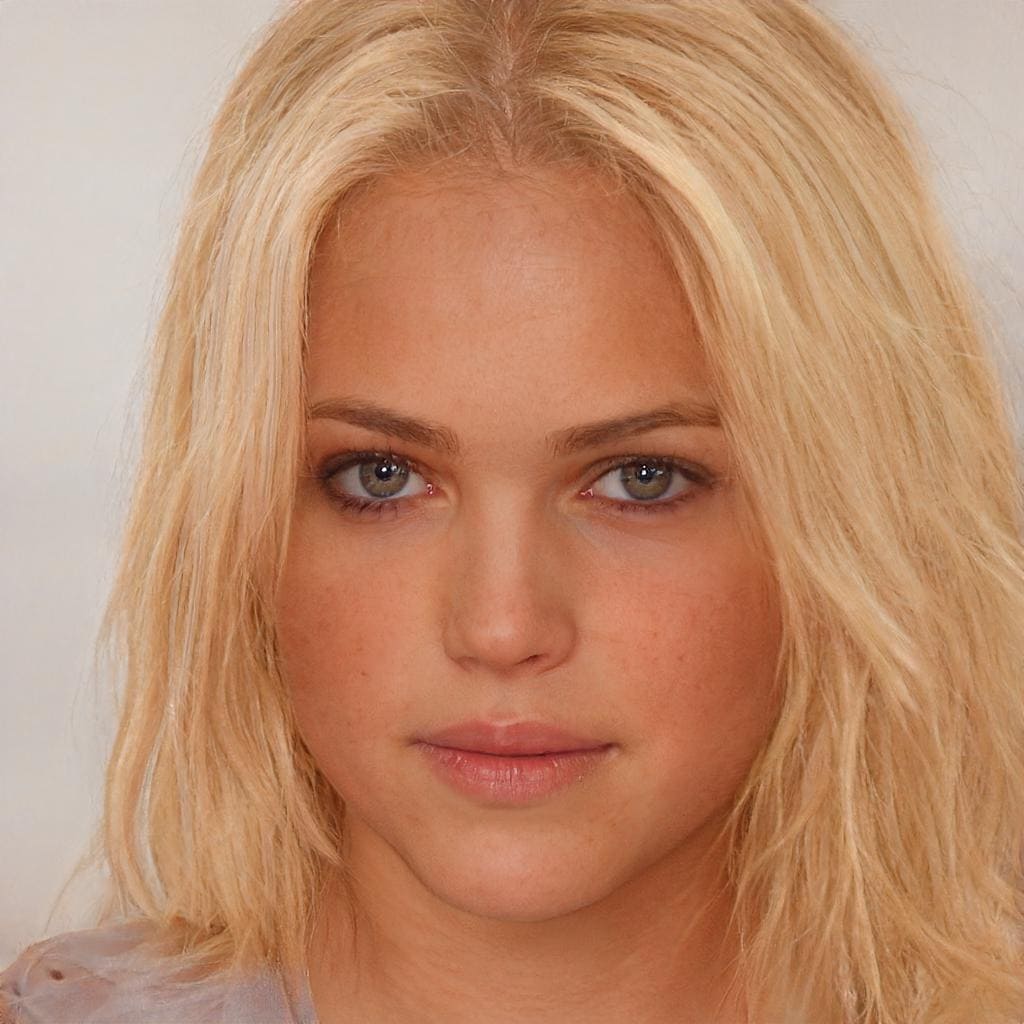}}\hfill  &
{\includegraphics[width=\linewidth]{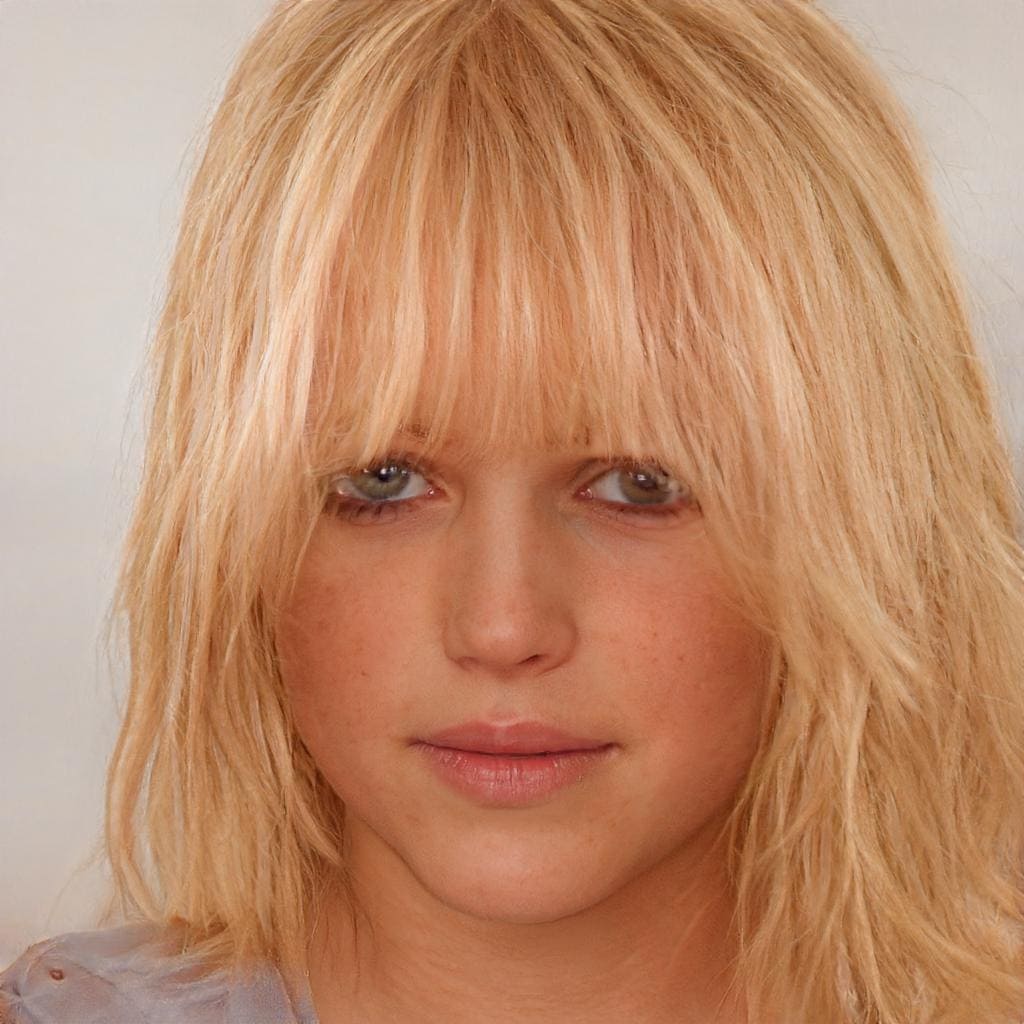}}\hfill  &
{\includegraphics[width=\linewidth]{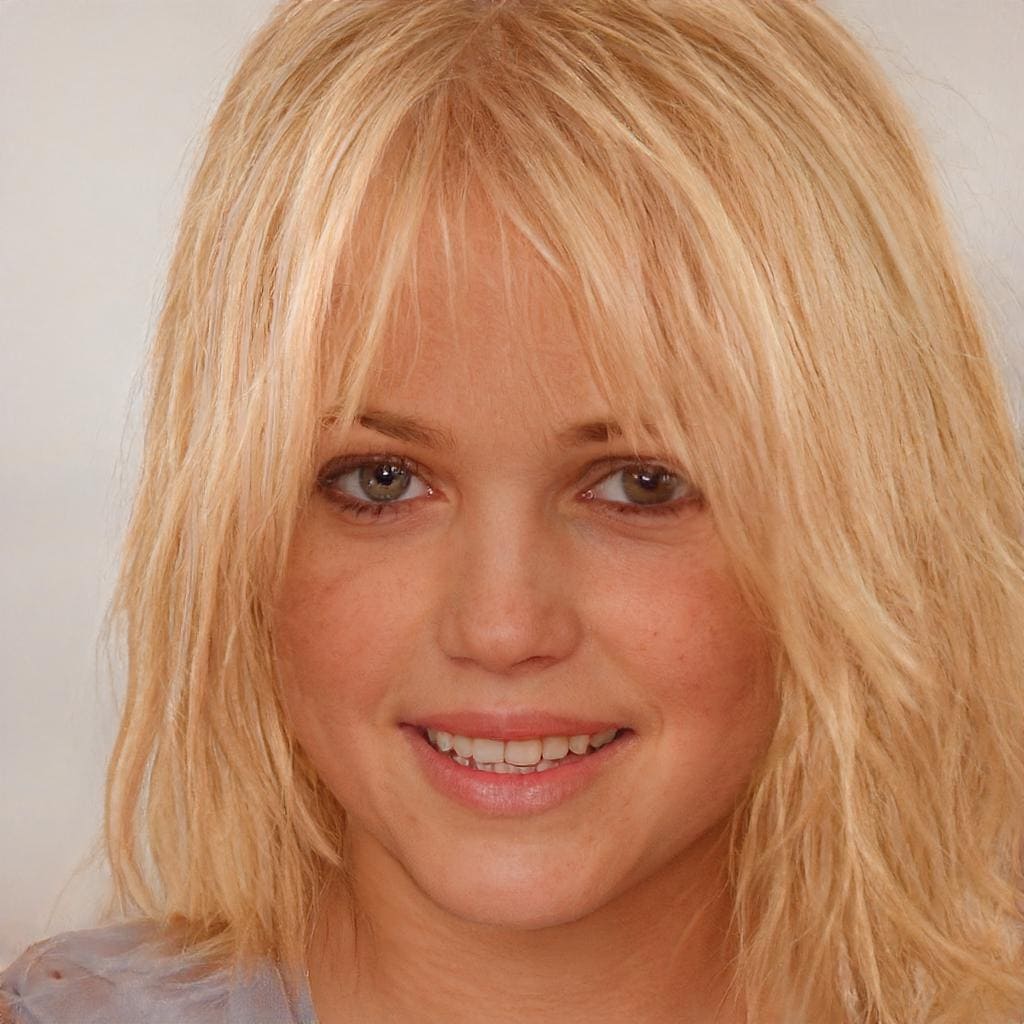}}\hfill &
{\includegraphics[width=\linewidth]{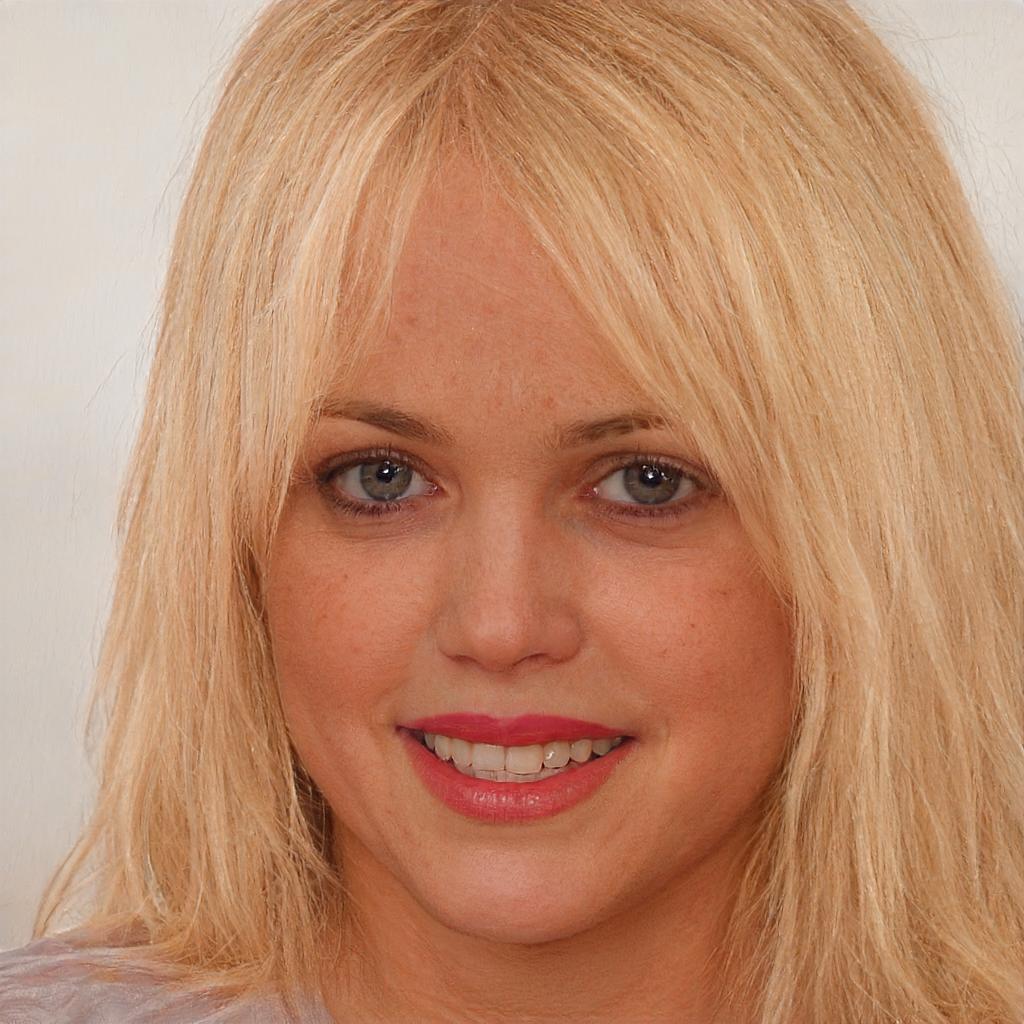}}\hfill \\[-3pt]

&&&
{\includegraphics[width=\linewidth]{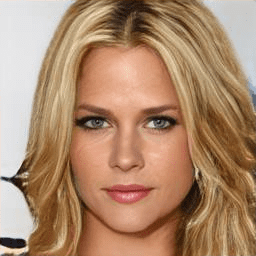}}\hfill  &
{\includegraphics[width=\linewidth]{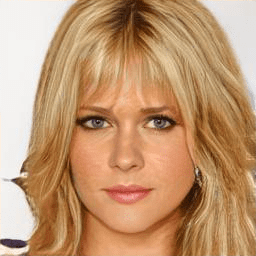}}\hfill  &
{\includegraphics[width=\linewidth]{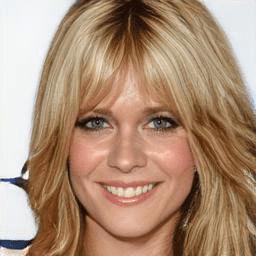}}\hfill &
{\includegraphics[width=\linewidth]{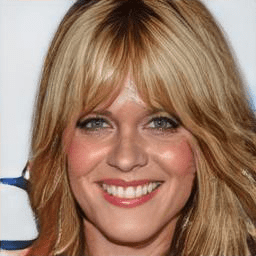}}\hfill \\[-3pt]

\end{tabular}\\[-5pt]
    \caption{\textbf{Additional qualitative results by (from top to bottom) DiffusionCLIP~\cite{kim2021diffusionclip}, StyleCLIP~\cite{patashnik2021styleclip}, and our LANIT}.}
    \label{fig:clip_comparison}\vspace{-10pt}
\end{figure}

\paragrapht{Comparisons to other CLIP-based methods.}
We compare our LANIT with CLIP-based text-guided manipulation models, DiffusionCLIP~\cite{kim2021diffusionclip} and StyleCLIP~\cite{patashnik2021styleclip}, on CelebA-HQ in \figref{fig:clip_comparison}. In the results, DiffusionCLIP and StyleCLIP show a limited capacity to synthesize realistic results with multiple attributes, while our LANIT generates high-quality images, faithfully representing each domain description, regardless of the number of domain descriptions. The reason is that our model focuses on learning style manifold during training from each target domain defined by textual descriptions, while others are designed to just fine-tune the pre-trained generative models on each instance independently. We observe that LANIT also works on a flexible number of K during inference time, which demonstrates the controllability of our model.

\subsection{Ablation Study and Analysis}
\vspace{-5pt}

\begin{table}
    \centering
    \resizebox{\linewidth}{!}{\begin{tabular}{l c c c c c }\toprule
         &\multicolumn{2}{c}{Default}
         &\multicolumn{3}{c}{Dictionary} \\
         \cmidrule(lr){2-3} \cmidrule(lr){4-6}
        Datasets &mFID $\downarrow$ & D\&C $\uparrow$ &mFID $\downarrow$ & D\&C $\uparrow$\\\midrule 
        CelebA-HQ~\cite{he2016deep}      & {27.96}    &{0.91 / 0.34}  &{28.34}     & {0.77 / 0.23}\\
        AnimalFaces-10~\cite{liu2019few} &{34.11} &{1.46 / 0.89}  &{40.48} & {1.01 / 0.78}\\ 
        Food-10~\cite{bossard2014food}  &{48.08}  &{1.24 / 0.86}  &{49.50} & {1.17 / 0.81} \\\bottomrule
    \end{tabular}}
    \vspace{-1mm}
    \caption{\textbf{Quantitative results with pre-defined dictionary.}}
    \label{tab:ablation_dic}\vspace{-10pt}
\end{table}

\begin{figure*}[t]
	\centering
  \begin{subfigure}{0.12\linewidth}
	{\includegraphics[width=1\linewidth]{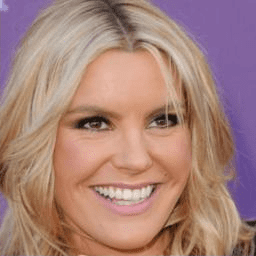}}\hfill
\end{subfigure}  
  \begin{subfigure}{0.12\linewidth}
	{\includegraphics[width=1\linewidth]{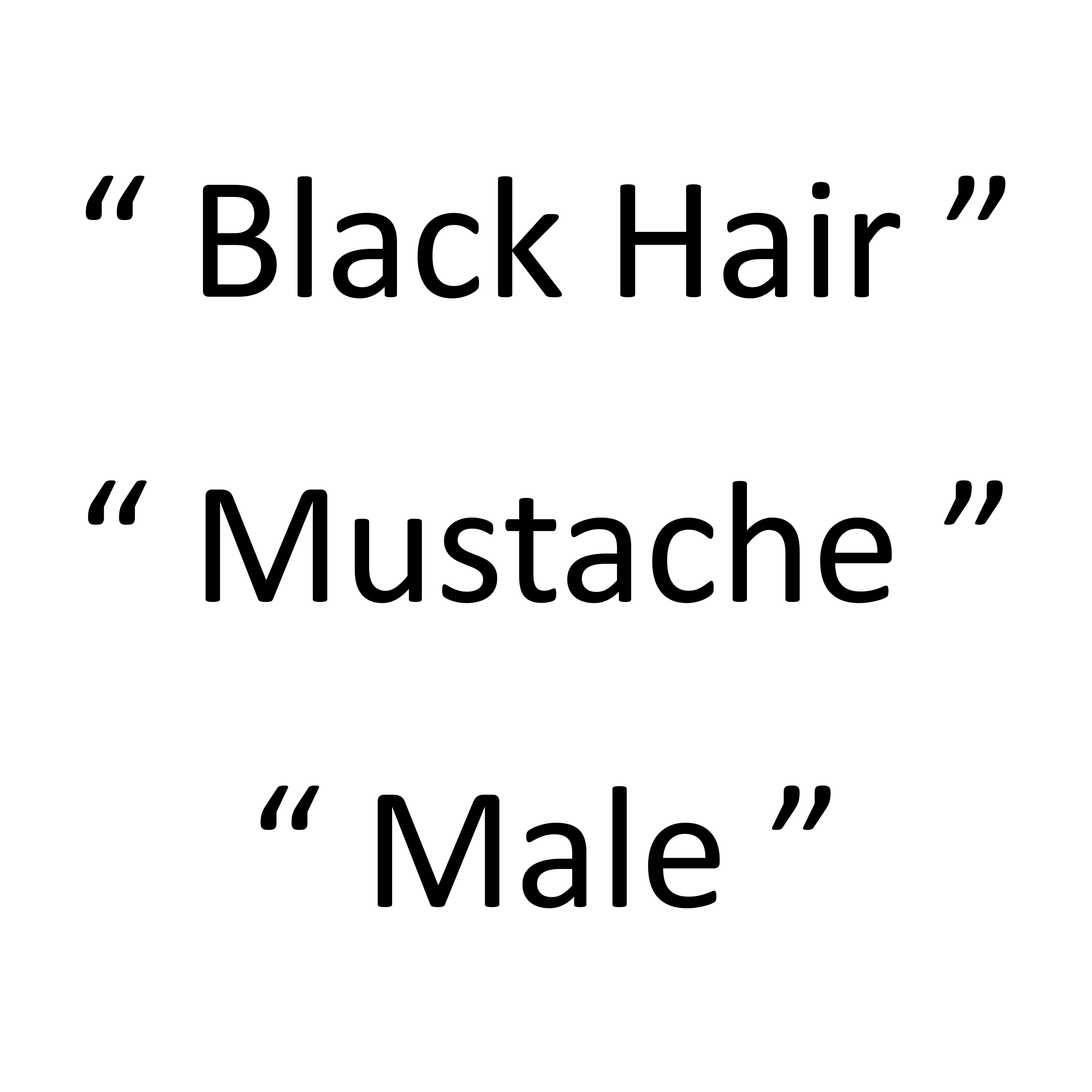}}\hfill
\end{subfigure}  
  \begin{subfigure}{0.12\linewidth}
	{\includegraphics[width=1\linewidth]{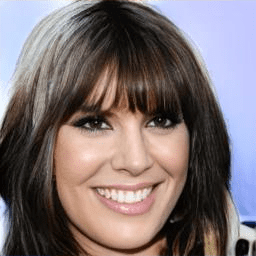}}\hfill
\end{subfigure}  
    \begin{subfigure}{0.12\linewidth}
	{\includegraphics[width=1\linewidth]{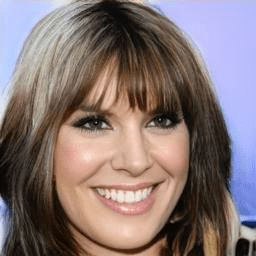}}\hfill
\end{subfigure}
  \begin{subfigure}{0.12\linewidth}
	{\includegraphics[width=1\linewidth]{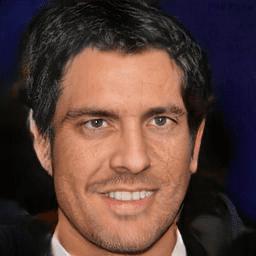}}\hfill
\end{subfigure}
  \begin{subfigure}{0.12\linewidth}
	{\includegraphics[width=1\linewidth]{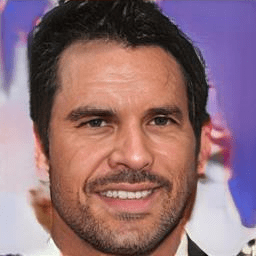}}\hfill
\end{subfigure}
  \begin{subfigure}{0.12\linewidth}
	{\includegraphics[width=1\linewidth]{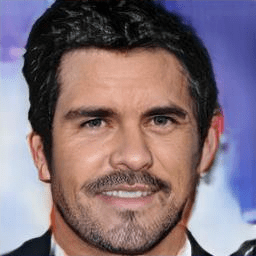}}\hfill
\end{subfigure} \\\vspace{0pt}

  \begin{subfigure}{0.12\linewidth}
	{\includegraphics[width=1\linewidth]{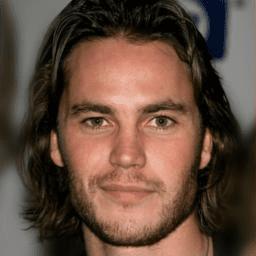}}\hfill\caption{Content}
\end{subfigure}  
  \begin{subfigure}{0.12\linewidth}
	{\includegraphics[width=1\linewidth]{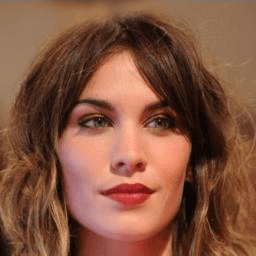}}\hfill\caption{Style}
\end{subfigure} 
  \begin{subfigure}{0.12\linewidth}
	{\includegraphics[width=1\linewidth]{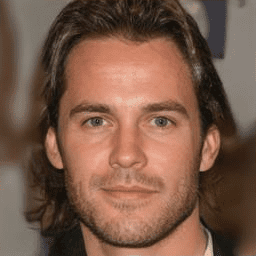}}\hfill\caption{A}
\end{subfigure} 
  \begin{subfigure}{0.12\linewidth}
	{\includegraphics[width=1\linewidth]{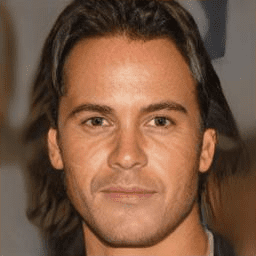}}\hfill\caption{B}
\end{subfigure} 
  \begin{subfigure}{0.12\linewidth}
	{\includegraphics[width=1\linewidth]{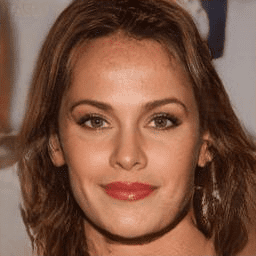}}\hfill\caption{C}
\end{subfigure} 
  \begin{subfigure}{0.12\linewidth}
	{\includegraphics[width=1\linewidth]{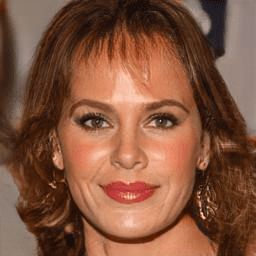}}\hfill\caption{D}
\end{subfigure} 
  \begin{subfigure}{0.12\linewidth}
	{\includegraphics[width=1\linewidth]{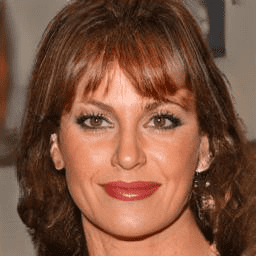}}\hfill\caption{E}
\end{subfigure} \\
	\vspace{-10pt}    
\caption{\textbf{Qualitative results for ablation study.}}
\label{fig:ablation_qual}\vspace{-10pt}
\end{figure*}

\paragrapht{Number of Domain Descriptions $N$.}
In~\tabref{tab:k_comparison}, We validate the effects of the number of candidate domains $N$ compared to TUNIT ~\cite{baek2021rethinking}. Our results show that a different number of $N$ brings minor changes to our model than TUNIT~\cite{baek2021rethinking}, which justifies the robustness of our framework. In animal faces-10, LANIT shows the best performance when there exist 10 candidate domains, whereas in CelebA-HQ, there is little difference in the FID score when $N$ is over 10. 

\vspace{3pt}
\paragrapht{Predefined Dictionary.}
We also evaluate the quantitative results when using a predefined dictionary instead of exploiting specific domain descriptions that we manually provide. For this experiment, we regard the whole class names in each dataset as a dictionary. For sampling candidate domains, we first calculated similarity values on the whole images and target domains, and then only averaged the similarity values assigned to the target domains in the dictionary. Then, we ordered average values in descending order and selected 10 domains. While our LANIT shows a slight drop in both mFID and F1 scores as shown in the~\tabref{tab:ablation_dic} for such a setting in all the datasets, it still records competitive performance compared to the default setting.

\vspace{3pt}
\paragrapht{Adaptive Thresholding with Base Prompt.}
We study the effectiveness of the adaptive thresholding technique using base prompt compared to the simple top-$K$ method in~\tabref{tab:loss_ablation}. We find that the proposed technique brings a large improvement in performance. This indicates that it enables the model to mitigate the influence of ambiguous or noisy data. Additionally, we evaluate our model with single or multiple activated top-$K$ attributes. We can see that the result of Top-3 significantly surpasses Top-1. This demonstrates that an image can be described more precisely when given with multiple keywords or text descriptions, while most prior works including TUNIT, Kim \textit{et al.} and StarGAN2 assume to assign a single label for each image. In addition, we evaluate the qualitative results in ~\figref{fig:ablation_qual} corresponding to the ~\tabref{tab:loss_ablation}. It shows that our LANIT captures more diverse attributes rather than Top-1 and Top-3 models.

\begin{table}
    \centering
    \resizebox{\linewidth}{!}
    {
    \begin{tabular}{lccc}\toprule
         &\multicolumn{3}{c}{CelebA-HQ~\cite{liu2015deep}} \\
         \cmidrule(lr){2-4}
        Method &mFID $\downarrow$ &D\&C $\uparrow$ &F1 $\uparrow$   \\\midrule 
        A \quad LANIT (Top-1)     \quad \quad &49.65  \quad  &0.56 / 0.32  \quad & 0.347\\ 
        B \quad LANIT (Top-3)     \quad \quad  &41.68 \quad &0.68 / 0.30    \quad & 0.564 \\ \hdline
        C \quad Baseline          \quad \quad  &{33.79} \quad &{0.70} / {0.26}   \quad &0.613 \\
        D \quad + DL Loss         \quad \quad  &{29.21} \quad &{0.86} / {0.32}  \quad &0.632\\
        \rowcolor{red!7}E \quad + Prompt Learning \quad \quad  &\textbf{27.96} \quad &\textbf{0.91} / \textbf{0.34} \quad & \textbf{0.670}  \\\bottomrule
    \end{tabular}}
    \vspace{-1mm}
    \caption{\textbf{Ablation Study.}}
    \label{tab:loss_ablation}\vspace{-10pt}
\end{table}

\begin{table}
    \centering
    \resizebox{0.75\linewidth}{!}
    {
    \begin{tabular}{lccc}\toprule
         &\multicolumn{3}{c}{CelebA-HQ~\cite{liu2015deep}} \\
         \cmidrule(lr){2-4} 
        Method &mFID $\downarrow$ &D\&C $\uparrow$ &F1 $\uparrow$  \\\midrule 
        End-to-End &{30.22}& {0.81 / 0.27} & {0.583} \\\hdline \rowcolor{red!7}
        Stage-wise & \textbf{27.96} & \textbf{0.91 / 0.34} & \textbf{0.670}  \\  \bottomrule
    \end{tabular}}
    \vspace{-1mm}
    \caption{\textbf{Ablation Study for PL.}}
    \label{tab:ablation_pl}\vspace{-10pt}
\end{table}

\vspace{3pt}
\paragrapht{Domain Regularization Loss.}
As shown in \tabref{tab:loss_ablation} (\textrm{D} and \textrm{E}), our LANIT improves performance Domain Regularization(DL) loss compared to the baseline on CelebA-HQ. Especially, D\&C metrics are significantly increased by adding our proposed domain regularization loss. In \figref{fig:ablation_qual}, the generated image has the style of reference image more faithfully by using domain regularization loss.

\vspace{3pt}
\paragrapht{Prompt Learning.}
In order to evaluate the effectiveness of prompt learning, we compare the estimated pseudo-domain labels with and without the proposed prompt learning, where the latter can be regarded as an offline clustering. It shows that with prompt learning, the probability of an image/text prompt pair is more accurately estimated. Since prompt learning helps reduce label noise by refining the prompt, the result without prompt learning shows limited clustering performance, which is verified in \tabref{tab:loss_ablation} (\textrm{E}). In addition, we further analyze the effect of when to start prompt learning \tabref{tab:ablation_pl}. The stage-wise learning, where we start prompt learning after convergence of the generative model, shows better performance than end-to-end learning with prompt learning. The reason is that the well-generated image corresponding to a given prompt has a critical role in prompt learning. Note that the base prompt is also updated with prompt learning simultaneously.

\section{Conclusion}
In this paper, we propose LANguage-driven Image-to-image Translation framework for Unlabeled Data (LANIT) that leverages \textit{dataset-level} supervision. By using textual domain descriptions as dataset-level supervision, our model not only mitigates the heavy dependency on per-sample supervision but also considers multi-hot domain labels for an image. The proposed labeling method allows users to explicitly define the semantic meaning of the target domain, which was not possible in the previous truly-unsupervised I2I framework. In addition, as we observe that a naive adoption of a pre-trained vision-language model has limited controllability and even drops the performance of the model, we present some techniques to overcome such limitations: adaptive thresholding, domain regularization loss, and prompt learning. Our experiments show that our LANIT performs better than truly-unsupervised methods in controllability on the application. Furthermore, the quantitative results of our framework have scored competitive or even better values than the existing image-to-image translation models that require per-sample supervision.

\vspace{3pt}
\noindent\textbf{Acknowledgements.}
This research was supported by the MSIT, Korea (IITP-2023-2020-0-01819, ICT Creative Consilience program, No. 2020-0-00368, A Neural-Symbolic Model for Knowledge Acquisition and Inference Techniques), and National Research Foundation of Korea (NRF2021R1C1C1006897).

\newpage

{\small
\bibliographystyle{ieee_fullname}
\bibliography{egbib}
}

\newpage
\onecolumn
\appendix

\onecolumn

\begin{center}
	\textbf{\large LANIT: Language-Driven Image-to-Image Translation for Unlabeled Data \\ - Supplementary Material-}
\end{center}
\vspace{5pt}

\appendix
\renewcommand{\thefigure}{\arabic{figure}}
\renewcommand{\theHfigure}{A\arabic{figure}}
\renewcommand{\thetable}{\arabic{table}}
\renewcommand{\theHtable}{A\arabic{table}}
\setcounter{figure}{0}
\setcounter{equation}{0}
\setcounter{table}{0}

In this document, we describe network architecture, details about domain descriptions, additional ablation study, additional experimental results, and user study for``LANIT: Language-Driven Image-to-Image Translation for Unlabeled Data".

\addcontentsline{toc}{section}{Appendix} 
\part{\Large } 
\parttoc 

\newpage

\section{More Implementation Details}
\subsection{Network architecture of LANIT.} 
We summarize the detailed network architecture of our LANIT in Table 1. We basically follow the content encoder, style encoder, mapping network, and generator architecture from StarGAN2~\cite{choi2020starganv2}. 

\vspace{-5pt}

	\begin{table}[h] 
	\centering
	\small{
	 \scalebox{0.78}{
		
			\begin{tabular}{>{\centering}m{0.25\linewidth} >{\centering}m{0.2\linewidth} 
					>{\centering}m{0.25\linewidth}>{\centering}m{0.30\linewidth}}
					\textbf{Content Encoder}&&& \tabularnewline
				\hline
				Layer & Resample & Norm & Output shape $(C \times H \times W)$ \tabularnewline
				\hline
				Conv1$\times$1  & - & - & $(64,256,256)$  \tabularnewline
				Resblock & AvgPool & InstanceNorm  & $(128,128,128)$  \tabularnewline	
				Resblock & AvgPool & InstanceNorm  & $(256,64,64)$  \tabularnewline
				Resblock & AvgPool & InstanceNorm  & $(512,32,32)$  \tabularnewline
				Resblock & AvgPool & InstanceNorm  & $(512,16,16)$  \tabularnewline
				\hline
                &&\tabularnewline 
                \textbf{Mapping Network}&&& \tabularnewline \hline
                Layer & Type & Activation & Output shape $(C)$  \tabularnewline\hline
				Latent & Shared & - & $16$ \tabularnewline
				Linear & Shared & ReLU & $512$ \tabularnewline
				Linear & Shared & ReLU & $512$ \tabularnewline
				Linear & Shared & ReLU & $512$ \tabularnewline
				Linear & Shared & ReLU & $512$ \tabularnewline
				Linear & Unshared & ReLU & $512$ \tabularnewline
				Linear & Unshared & ReLU & $512$ \tabularnewline
				Linear & Unshared & ReLU & $512$ \tabularnewline
				Linear & Unshared & - & $64$ \tabularnewline

				\hline
                &&\tabularnewline 
                \textbf{Generator}&&& \tabularnewline \hline
				Layer & Resample & Norm & Output shape $(C \times H \times W)$ \tabularnewline
				\hline
                Resblock & - & InstanceNorm & $(512,16,16)$  \tabularnewline
                Resblock & - & InstanceNorm & $(512,16,16)$  \tabularnewline
                Resblock & - & AdaptiveInstanceNorm & $(512,16,16)$  \tabularnewline
                Resblock & - & AdaptiveInstanceNorm & $(512,16,16)$  \tabularnewline
                Resblock & Upsample & AdaptiveInstanceNorm & $(512,32,32)$  \tabularnewline
                Resblock & Upsample & AdaptiveInstanceNorm & $(256,64,64)$  \tabularnewline
                Resblock & Upsample & AdaptiveInstanceNorm & $(128,128,128)$  \tabularnewline
                Resblock & Upsample & AdaptiveInstanceNorm & $(64,256,256)$  \tabularnewline
                Conv1$\times$1  & - & - & $(3,256,256)$  \tabularnewline
				\hline
				
                &&\tabularnewline 
                \textbf{Style Encoder and}&\textbf{Discriminator}&& \tabularnewline \hline
				Layer & Type & Activation & Output shape $(C \times H \times W)$ \tabularnewline
				\hline
				Conv1$\times$1  & - & - & $(64,256,256)$  \tabularnewline
				Resblock & AvgPool & InstanceNorm  & $(128,128,128)$  \tabularnewline	
				Resblock & AvgPool & InstanceNorm  & $(256,64,64)$  \tabularnewline
				Resblock & AvgPool & InstanceNorm  & $(512,32,32)$  \tabularnewline
				Resblock & AvgPool & InstanceNorm  & $(512,16,16)$  \tabularnewline\hline\hline
				Resblock & AvgPool & -  & $(512,8,8)$  \tabularnewline
				Resblock & AvgPool & -  & $(512,4,4)$  \tabularnewline
				LReLU & - & -  & $(512,4,4)$  \tabularnewline
				Conv4$\times$4 & - & -  & $(512,1,1)$  \tabularnewline
				LReLU & - & -  & $(512,1,1)$  \tabularnewline\hline\hline
                Linear(Unshared) & - & - & $(64,1,1)$ \tabularnewline
				\hline
			\end{tabular}}
			\caption{\textbf{Network architecture of our LANIT.}}
		}\label{tab:1}
	\end{table} 

\subsection{Additional experimental setup.}
We employ an Adam optimizer, where $\beta_{1} = 0.0$ and $\beta_{2} = 0.99$, for 100,000 iterations using a step decay learning rate scheduler. We also set a batch size of 8, an initial learning rate of 1e-4 for the encoder, generator, and discriminator, and 1e-6 for the prompt. All coefficients for the losses are set to 1. The training images are resized to 256$\times$256. We conduct experiments using a single 24GB RTX 3090 GPU.

\clearpage
\subsection{Template augmentation.}
In addition, we utilize text-augmentation to boost the accuracy of pseudo labels. which are
``a [domain] photo with [candidate domain].'',
``a [domain] photo of the [candidate domain].'',
``the [domain] photo of the [candidate domain].'',
``a good [domain] photo of the [candidate domain].'',
``high quality [domain] photo of [candidate domain].'',
``a [domain] image of [candidate domain].'',
``the [domain] image of [candidate domain].'',
``high quality [domain] image of [candidate domain].'',
``a high quality [domain] image of [candidate domain].''
.
The [domain] means the kind of text that represents the specific dataset such as ``face'', ``animal'', ``food'' etc. The [candidate domain] indicates the texts that can describe the specific domain as shown in~\tabref{tab:description}

As shown in the~\tabref{tab:f1_score}, we can show that the template augmentation technique can boost the performance of pseudo labels.

\subsection{Details of the latent-guided image-to-image translation.}
\begin{figure*}[h]
\centering
\includegraphics[width=0.9\linewidth]{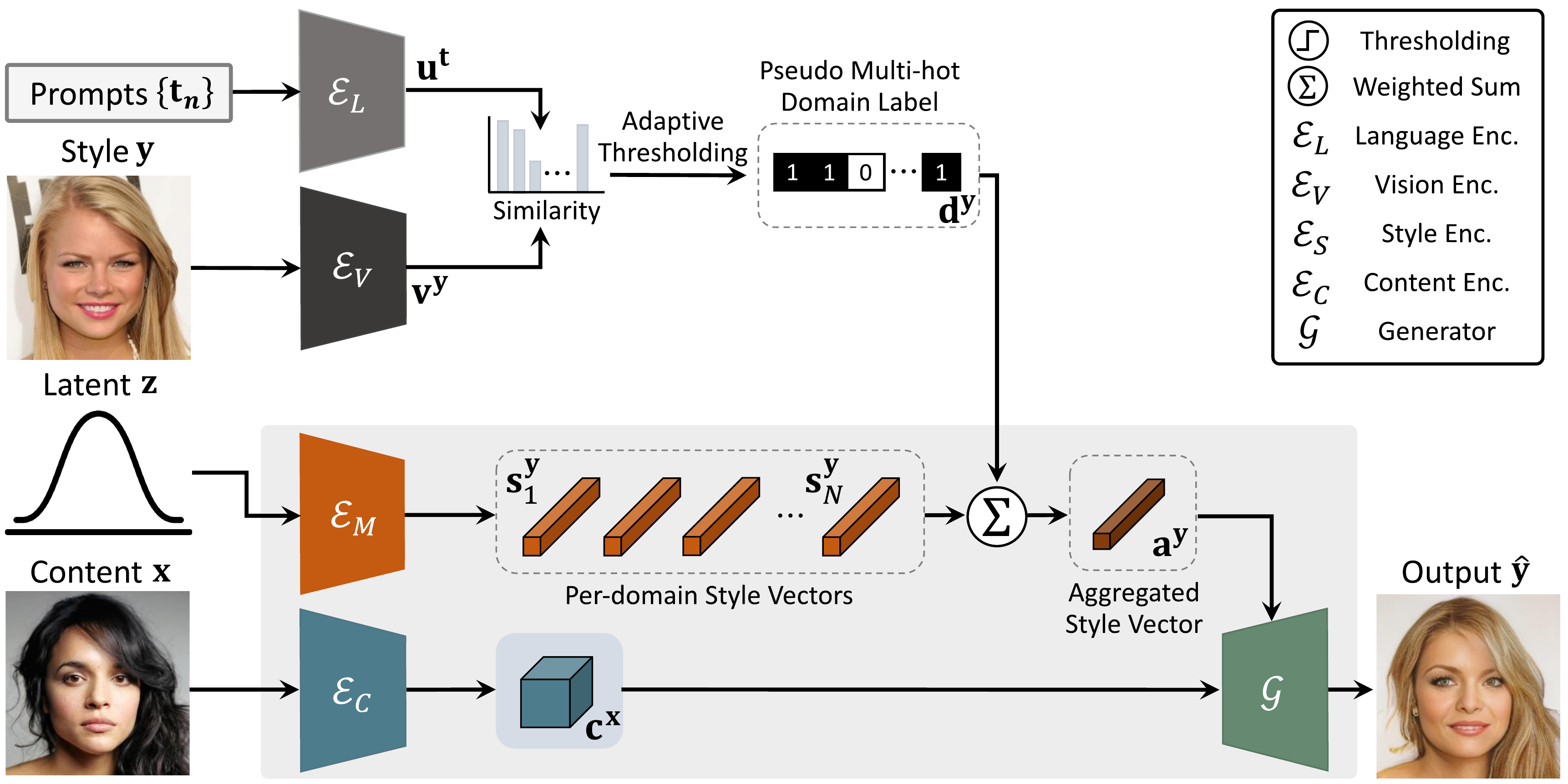}\caption{\textbf{Latent-guided image translation network.}}
\label{fig:lat_arch}\vspace{-10pt}
\end{figure*}

As shown in ~\figref{fig:lat_arch}, instead of utilizing reference images and a style encoder, we generated style vectors by inputting latent vectors sampled from a Gaussian distribution into the mapping network. Then style vectors were aggregated with pseudo labels provided by users.

\clearpage
\subsection{Details of templates and candidate domains.}
\tabref{tab:description} describes additional details of the candidate domains.
We follow 10 pre-defined domains from TUNIT~\cite{baek2021rethinking} for Animal Faces-10 and Food-10. Likewise, for CelebA-HQ~\cite{liu2015deep}, we obtain 40 pre-defined textual attributes from CelebA-HQ~\cite{liu2015deep}, and we have mainly shown the results using 10 domain descriptions, which are randomly selected for 3 times, and then report the average results. Please note that \textit{we do not use per-sample domain labels} in all cases and the domain labels work as the dataset-level candidates.

\begin{table}[h]
\setlength{\tabcolsep}{6pt} 
\centering
\resizebox{1.0\textwidth}{!}{
\normalsize{
\begin{tabular}{|c|c|c|c|}
\hline
Datasets & Template(default)	& \textit{N} &	Candidate Domains\\

\hline\hline

\multirow{3}{*}{\makecell[c]{CelebA-HQ~\cite{liu2015deep}}}
&\multirow{3}{*}{A face of }
& 4 & `blond hair', `bang', `smiling', `wearing lipstick'\\  \cline{3-4}
&& 7 & \makecell[c]{`blond hair', `black hair' , `smiling', `wearing lipstick', \\  `arched eyebrows', `bangs',`mustache'}\\ \cline{3-4}
&& 10 & \makecell[c]{`bangs', `blond hair', `black hair' ,`smiling', `arched eyebrows', \\ `heavy makeup',`mustache', `straight hair', `wearing lipstick', `male’}
\\ \hline

\multirow{3}{*}{Animal Faces-10~\cite{liu2019few}}
&\multirow{3}{*}{\makecell[c]{A photo of \\ animal face with} }
& 4 & `beagle', `golden retriever', `tabby cat', `tiger'\\ \cline{3-4}
&& 7 & \makecell[c]{`beagle', `dandie dinmont terrier', `golden retriever', `white fox',\\ `tabby cat', `snow leopard', `tiger'} \\  \cline{3-4}
&& 10 & \makecell[c]{`appenzeller sennenhund', `beagle', `dandie dinmont terrier', `golden retriever', \\ 
`malinois', `white fox', `tabby cat', `snow leopard', `lion', `tiger'}
\\ \hline

\multirow{3}{*}{Food-10~\cite{bossard2014food}} 
&\multirow{3}{*}{A photo of food with}
& 4 & `baby back ribs', `beignets', `dumplings', `edamame'\\ \cline{3-4}
&& 7 & \makecell[c]{`baby back ribs', `beef carpaccio', `beignets', `clam chowder',\\ `dumplings', `edamame', `strawberry shortcake'}  \\  \cline{3-4}
&& 10 & \makecell[c]{`baby back ribs', `beef carpaccio', `beignets', `bibimbap', `caesar salad', \\
`clam chowder', `dumplings', `edamame', `spaghetti bolognese', `strawberry shortcake'}
\\ \hline

LHQ~\cite{skorokhodov2021aligning}
&{A photo of scene}
& 10 & \makecell[c]{`with mountain', `with field', `with lake' ,`with ocean', `with waterfall', `in summer',\\
`in winter', `on sunny day', `on cloudy day', `at sunset'}
\\ \hline

MetFace
&{A portrait with}
& 10 & \makecell[c]{`oil painting', `grayscale', `black hair', `wavy hair', `male', \\
`mustache', `smiling', `gray hair', `blonde hair', `sculpture'}
\\ \hline

Anime~\cite{chao2019/online}
&{\makecell[c]{A photo of \\anime with}}
& 10 & \makecell[c]{ `brown hair', `red hair', `black hair', `purple hair', `blond hair', \\
 `blue hair', `pink hair', `silver hair', `green hair', `white hair'}
\\ \hline

LSUN-Car~\cite{yu2015lsun}
&{A car painted with}
& 10 & \makecell[c]{`red color', `orange color', `gray color', `blue color', `yellow color',\\
`white color', `black color', `silver color', `green color', `pink color'}
\\ \hline

LSUN-Church~\cite{yu2015lsun} 
&{A church}
& 7 & \makecell[c]{`at night', `with sunset', `in winter', `on cloudy day',\\
`on sunny day', `with trees', `with a river'}
\\ \hline

\end{tabular}}}
\caption{\textbf{Examples of Templates and Candidate Domains for Each Dataset.}}
\label{tab:description}
\vspace{-5pt}

\end{table}

\clearpage
\subsection{Details of the predefined dictionary.}
\tabref{tab:dictionary_description} describes additional details on candidate domains on a predefined dictionary. We used a pre-defined dictionary containing various domain descriptions for each dataset: CelebA-HQ, AnimalFaces-149, and Food-101 labels. As mentioned in the main paper, we selected 10 domains that have the highest similarity values within the whole image and target domains. We utilize the images of 10 classes used in the main paper and define the dictionary as the bundle of class names in the whole dataset. As can be seen in the table below, the selected candidate domains are mostly confident that have higher similarity values than other target domains. 

\begin{table}[h]
\setlength{\tabcolsep}{6pt} 
\centering
\resizebox{1.0\textwidth}{!}{
\normalsize{
\begin{tabular}{|c|c|}
\hline
Datasets &	Selected Candidate Domains\\

\hline\hline

\multirow{1}{*}{\makecell[c]{CelebA-HQ~\cite{liu2015deep}}}
& \makecell[c]{‘blond hair’, ‘bald’, ‘wavy hair’,
‘black hair’, ‘smiling’,\\ ‘straight hair’, ‘eyeglasses’, ‘goatee’, ‘bangs’, ‘arched eyebrows’}
\\ \hline

\multirow{1}{*}{Animal Faces-149~\cite{liu2019few}}
& \makecell[c]{`dandie dinmont terrier',`malinois',`appenzeller sennenhund', `white fox', `tabby cat', \\ `snow leopard', `lion', `bengal tiger', `grey fox', `german shepherd dog'}
\\ \hline

\multirow{1}{*}{Food-101~\cite{bossard2014food}} 
& \makecell[c]{`french toast', `beef carpaccio', `beignets', `seaweed salad', `caesar salad', \\
`clam chowder', `dumplings', `edamame', `spaghetti bolognese', `strawberry shortcake'}
\\ \hline

\end{tabular}}}
\caption{\textbf{Selected Candidate Domains in the Predefined Dictionary.}}
\label{tab:dictionary_description}
\vspace{-5pt}

\end{table}

\clearpage
\section{Additional Ablation Study}

\subsection{Number of domain descriptions $N$.}
\begin{figure*}[h]
\centering
\newcolumntype{M}[1]{>{\centering\arraybackslash}m{#1}}
\setlength{\tabcolsep}{1pt} 
\renewcommand{\arraystretch}{2} 
\begin{tabular}
{M{0.13\linewidth}M{0.02\linewidth}M{0.13\linewidth}M{0.13\linewidth}M{0.13\linewidth}M{0.13\linewidth}M{0.13\linewidth}M{0.13\linewidth}}
Content && Style & \textit{N} = 4 & \textit{N} = 7 & \textit{N} = 10 & \textit{N} = 13 & \textit{N} = 16 \\ \hline \\[-15pt]
{\includegraphics[width=\linewidth]{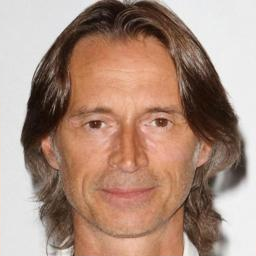}}\hfill & &
{\includegraphics[width=\linewidth]{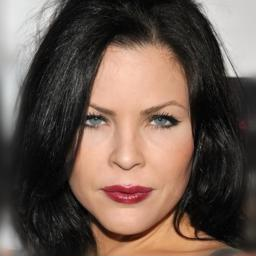}}\hfill  &
{\includegraphics[width=\linewidth]{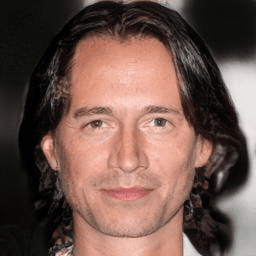}}\hfill  &
{\includegraphics[width=\linewidth]{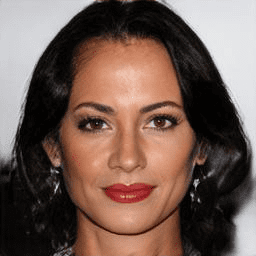}}\hfill  &
{\includegraphics[width=\linewidth]{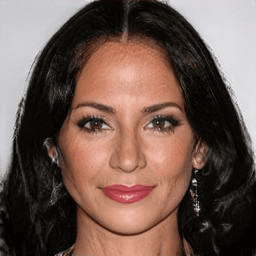}}\hfill  &
{\includegraphics[width=\linewidth]{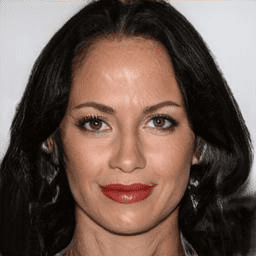}}\hfill  &
{\includegraphics[width=\linewidth]{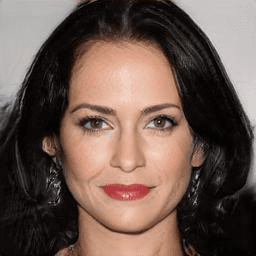}}\hfill \\[-3pt]
{\includegraphics[width=\linewidth]{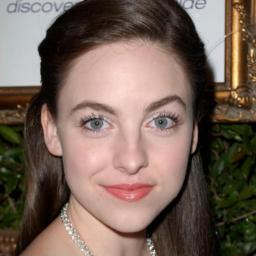}}\hfill & &
{\includegraphics[width=\linewidth]{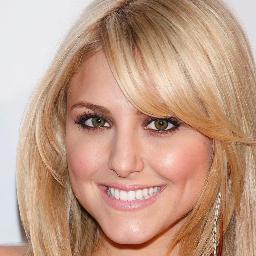}}\hfill  &
{\includegraphics[width=\linewidth]{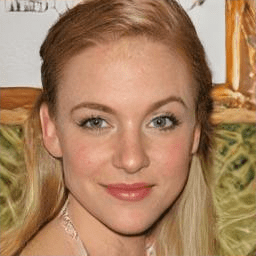}}\hfill  &
{\includegraphics[width=\linewidth]{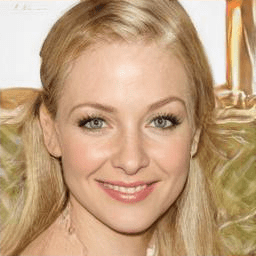}}\hfill  &
{\includegraphics[width=\linewidth]{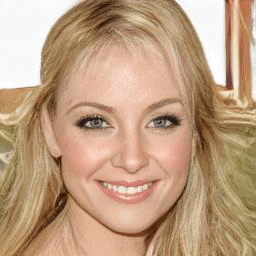}}\hfill  &
{\includegraphics[width=\linewidth]{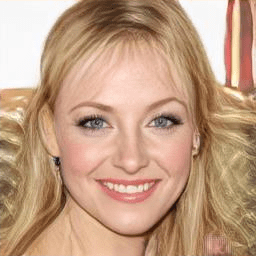}}\hfill  &
{\includegraphics[width=\linewidth]{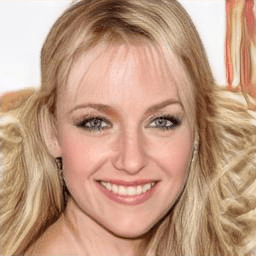}}\hfill\\[-3pt]
\end{tabular}
    \caption{\textbf{Qualitative results by varying the number of domain description $N$.}}
    \label{fig:sup_4710}
\end{figure*}

In the main paper, we have examined the impacts of the number of candidate domains, base prompt thresholding, and prompt learning. In this section, we additionally validate the effects of the different number of domain descriptions, shown in \figref{fig:sup_4710} on CelebA-HQ. While our model consistently generates impressive outputs, results with a larger number of domain descriptions tend to faithfully represent diverse attributes. In addition, we evaluate the F1 score by varying the number of domains in \tabref{tab:f1_score}. Our model shows the highest F1 score when N is 10 in both datasets.

\begin{table}[h]
    \centering
    \normalsize{
    \scalebox{1}{
    \begin{tabular}{l ccccc ccccc}\toprule
         &\multicolumn{5}{c}{AnimalFaces-10~\cite{liu2019few}}
         &\multicolumn{5}{c}{CelebA-HQ~\cite{liu2015deep}} \\
         \cmidrule(lr){2-6} \cmidrule(lr){7-11}
        N & Top-1 & Top-3 & Baseline & TextAug & Prompt learning     & Top-1 & Top-3 & Baseline & TextAug & Prompt learning \\\midrule 
        4 & {0.762} & {0.672} & {0.678} & {0.796} & \textbf{0.832} &         {0.372} & {0.421} & {0.423} & {0.435} & \textbf{0.481}   \\ \hdline
        7 & {0.903} & {0.701} & {0.688} & {0.862} & \textbf{0.893} &         {0.423} & {0.613} & {0.610} & {0.631} & \textbf{0.652} \\\hdline
        10 & {0.956} & {0.723} & {0.693} & {0.835} & \textbf{0.880} &         {0.355} & {0.562} & {0.610} & {0.638} & \textbf{0.670}  \\\hdline
        13 & {0.826} & {0.654} & {0.606} & {0.785} & \textbf{0.801} &         {0.293} & {0.533} & {0.591} & {0.612} & \textbf{0.639}   \\\hdline
        16 &  {0.753} & {0.630} & {0.601} & {0.753} & \textbf{0.783} &         {0.300} & {0.522} & {0.562} & {0.613} & \textbf{0.641}  \\\bottomrule
        
    \end{tabular}}}
    \vspace{-1mm}
    \caption{\textbf{F1 score by varying number of domains.}}
    \label{tab:f1_score}
\end{table}

\clearpage
\section{Additional Comparisons}
\subsection{Additional comparisons to other truly-unsupervised methods on reference-guided translation.}
\begin{figure*}[h]
 \centering	
  \begin{subfigure}{0.13\linewidth}
	{\includegraphics[width=1\linewidth]{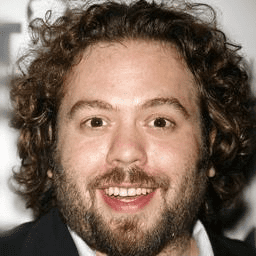}}\hfill
\end{subfigure}
  \begin{subfigure}{0.13\linewidth}
	{\includegraphics[width=1\linewidth]{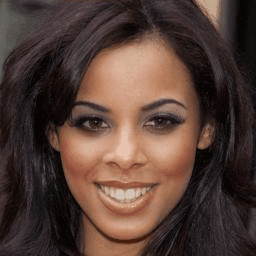}}\hfill
\end{subfigure}
  \begin{subfigure}{0.13\linewidth}
  \centering
	{\includegraphics[width=1\linewidth]{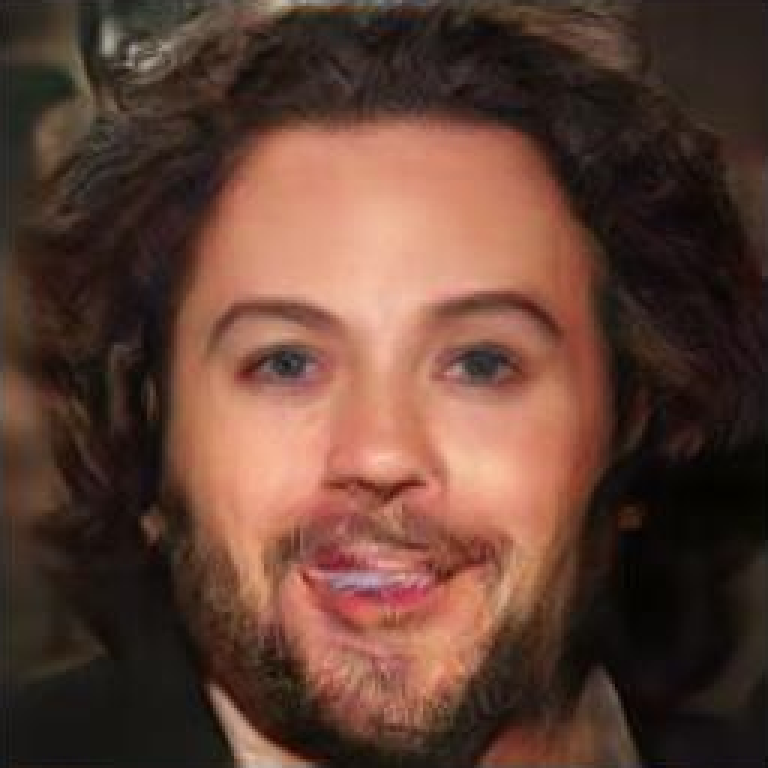}}\hfill
\end{subfigure}
  \begin{subfigure}{0.13\linewidth}
  \centering
	{\includegraphics[width=1\linewidth]{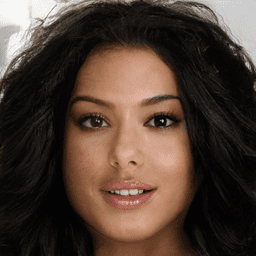}}\hfill
\end{subfigure}
  \begin{subfigure}{0.13\linewidth}
  \centering
	{\includegraphics[width=1\linewidth]{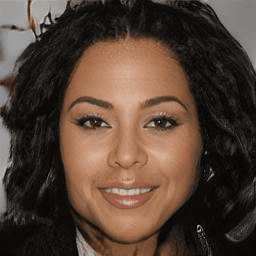}}\hfill
\end{subfigure}
  \begin{subfigure}{0.13\linewidth}
  \centering
	{\includegraphics[width=1\linewidth]{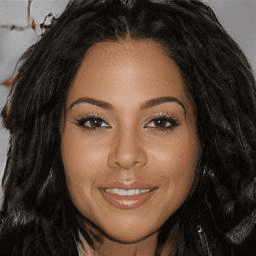}}\hfill
\end{subfigure}
  \begin{subfigure}{0.13\linewidth}
  \centering
	{\includegraphics[width=1\linewidth]{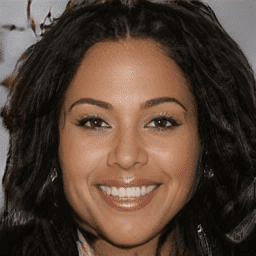}}\hfill\end{subfigure}\\\vspace{0pt}
  \begin{subfigure}{0.13\linewidth}
	{\includegraphics[width=1\linewidth]{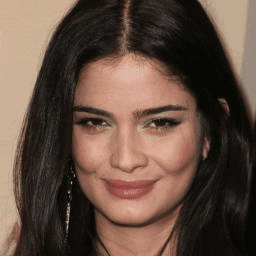}}\caption{Content}\hfill
\end{subfigure}
  \begin{subfigure}{0.13\linewidth}
	{\includegraphics[width=1\linewidth]{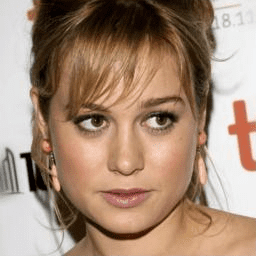}}\caption{Style}\hfill
\end{subfigure}
  \begin{subfigure}{0.13\linewidth}
  \centering
	{\includegraphics[width=1\linewidth]{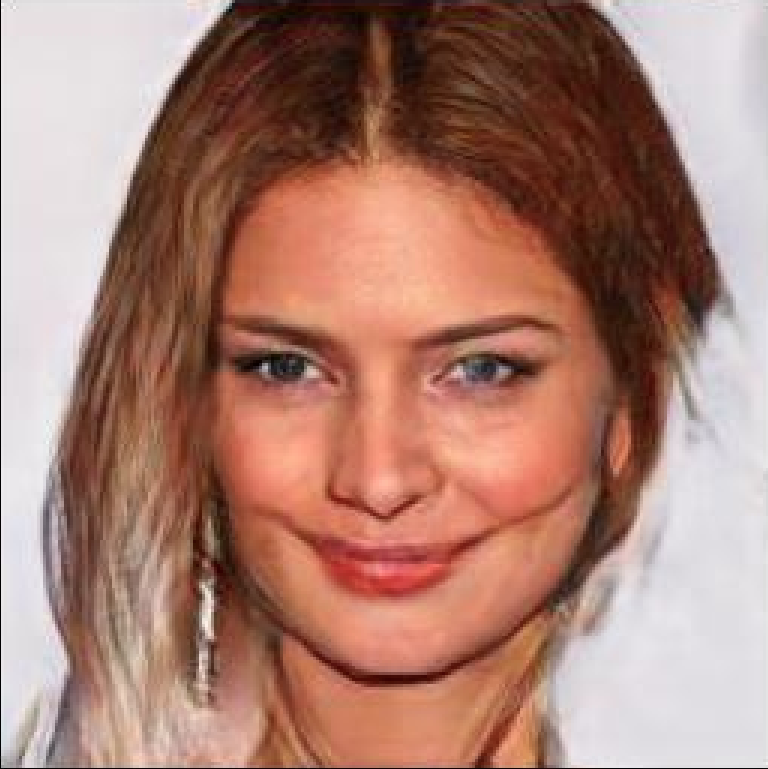}}\caption{TUNIT~\cite{baek2021rethinking}}\hfill
\end{subfigure}
  \begin{subfigure}{0.13\linewidth}
  \centering
	{\includegraphics[width=1\linewidth]{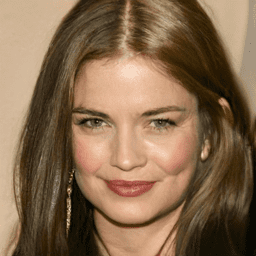}}\caption{Kim \textit{et al}.~\cite{kim2022style}}\hfill
\end{subfigure}
  \begin{subfigure}{0.13\linewidth}
  \centering
	{\includegraphics[width=1\linewidth]{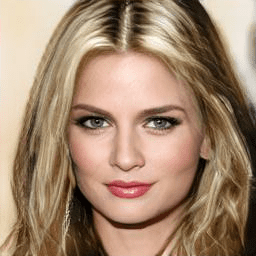}}\caption{Ours(K=1)}\hfill
\end{subfigure}
  \begin{subfigure}{0.13\linewidth}
  \centering
	{\includegraphics[width=1\linewidth]{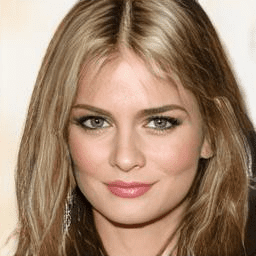}}\caption{Ours(K=2)}\hfill
\end{subfigure}
  \begin{subfigure}{0.13\linewidth}
  \centering
	{\includegraphics[width=1\linewidth]{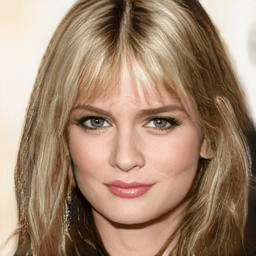}}\caption{Ours(K=3)}\hfill\end{subfigure}\\
	\vspace{-15pt}
	\caption{\textbf{Reference-guided translation results.}}
	\label{fig:reference_guided}\vspace{-5pt}
\end{figure*}

We additionally provide reference-guided image translation results, compared to existing fully-unsupervised I2I methods in \figref{fig:reference_guided}: TUNIT~\cite{baek2021rethinking} and Kim \textit{et al.}~\cite{kim2022style}. Specifically, we visualize our LANIT with a different number of K (K=1,2,3). The results with more K have better capability to represent more diverse attributes. Meanwhile, our method is able to generate robust results with high fidelity regardless of K.
On contrary, TUNIT and Kim \textit{et al.} are limited to faithfully represent the style attributes from the reference images.

\clearpage
\section{Additional Results of LANIT}
\begin{figure}
    \centering
\begin{center}
\begin{tabular}{ccc}
& \hspace{-10pt} \raisebox{0.11\height}{\rotatebox{90}{\hspace{13pt} Style}} 
& \includegraphics[width=0.13\textwidth]{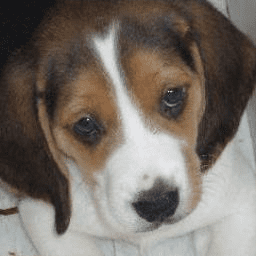}
\includegraphics[width=0.13\textwidth]{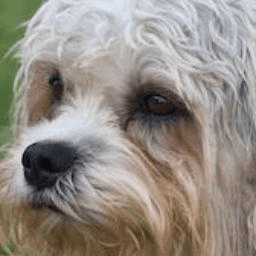}
\includegraphics[width=0.13\textwidth]{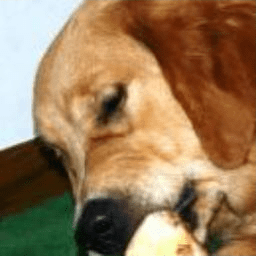}
\includegraphics[width=0.13\textwidth]{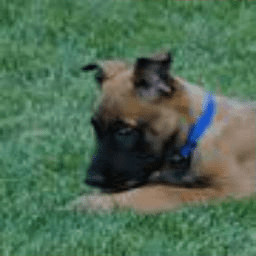}
\includegraphics[width=0.13\textwidth]{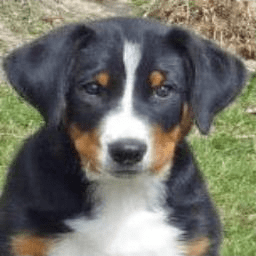} \\[-7pt]
Content \\ \hline \\[-5pt]
\includegraphics[width=0.13\textwidth]{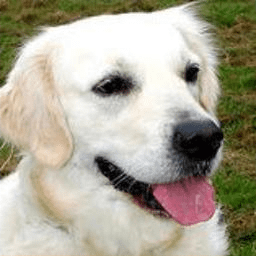} & &
\includegraphics[width=0.13\textwidth]{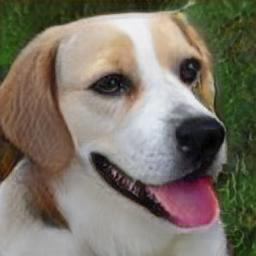}
\includegraphics[width=0.13\textwidth]{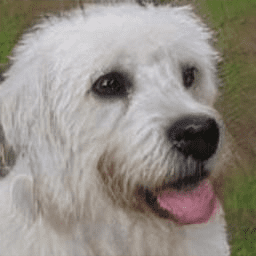}
\includegraphics[width=0.13\textwidth]{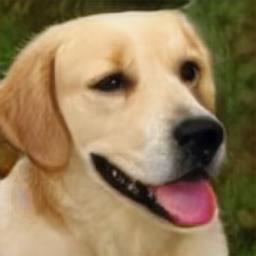}
\includegraphics[width=0.13\textwidth]{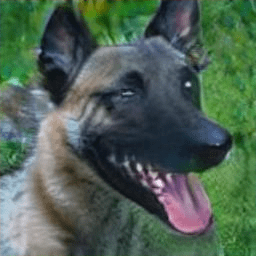}
\includegraphics[width=0.13\textwidth]{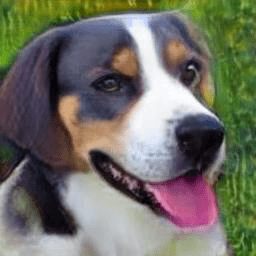} \\
\includegraphics[width=0.13\textwidth]{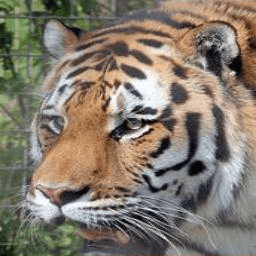} & &
\includegraphics[width=0.13\textwidth]{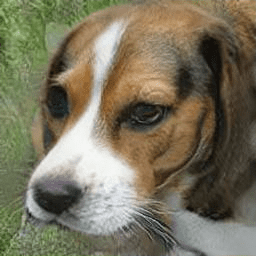}
\includegraphics[width=0.13\textwidth]{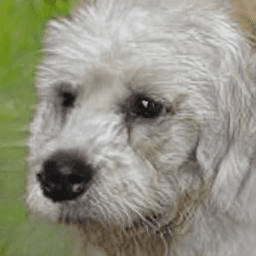}
\includegraphics[width=0.13\textwidth]{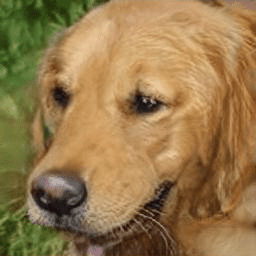}
\includegraphics[width=0.13\textwidth]{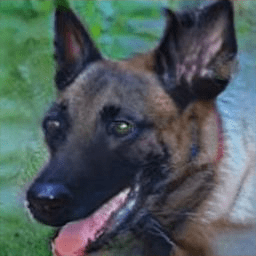}
\includegraphics[width=0.13\textwidth]{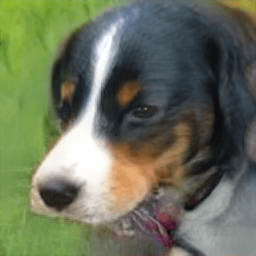} \\
\includegraphics[width=0.13\textwidth]{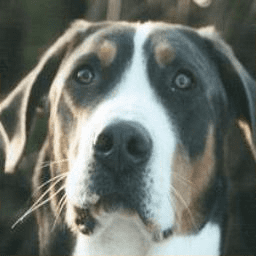} & &
\includegraphics[width=0.13\textwidth]{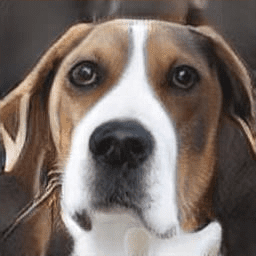}
\includegraphics[width=0.13\textwidth]{figure_sup/af_ref/fake/10_00.png}
\includegraphics[width=0.13\textwidth]{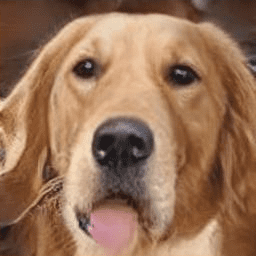}
\includegraphics[width=0.13\textwidth]{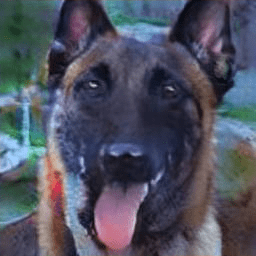}
\includegraphics[width=0.13\textwidth]{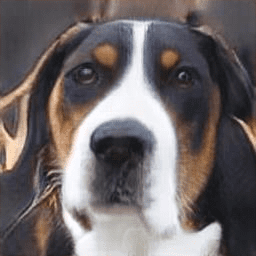} \\[10pt]

& \hspace{-10pt} \raisebox{0.11\height}{\rotatebox{90}{\hspace{13pt} Style}} 
& \includegraphics[width=0.13\textwidth]{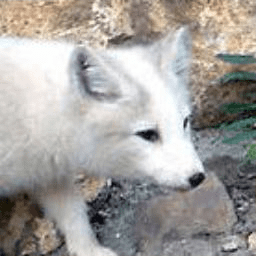}
\includegraphics[width=0.13\textwidth]{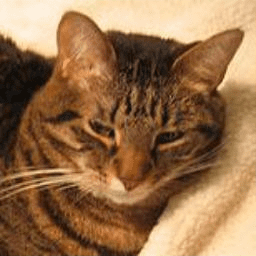}
\includegraphics[width=0.13\textwidth]{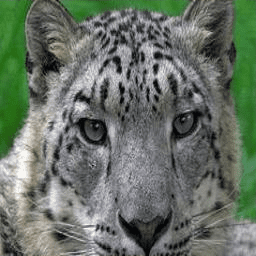}
\includegraphics[width=0.13\textwidth]{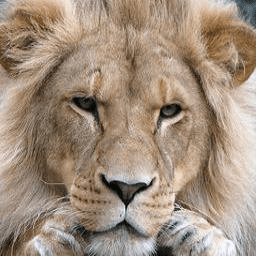}
\includegraphics[width=0.13\textwidth]{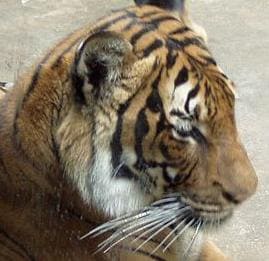} \\[-7pt]
Content \\ \hline \\[-5pt]
\includegraphics[width=0.13\textwidth]{figure_sup/af_ref/con/3.png} & &
\includegraphics[width=0.13\textwidth]{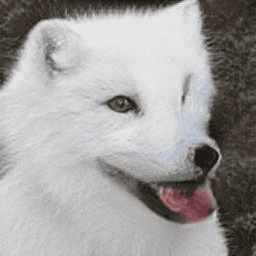}
\includegraphics[width=0.13\textwidth]{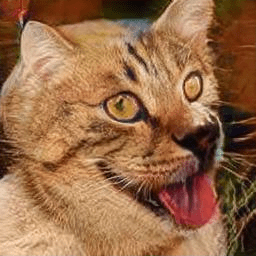}
\includegraphics[width=0.13\textwidth]{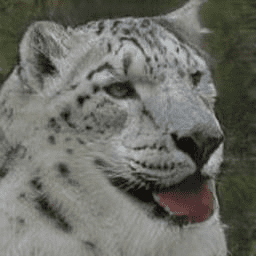}
\includegraphics[width=0.13\textwidth]{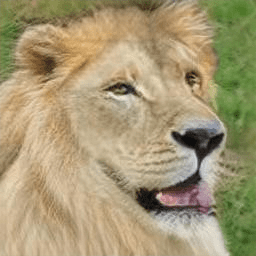}
\includegraphics[width=0.13\textwidth]{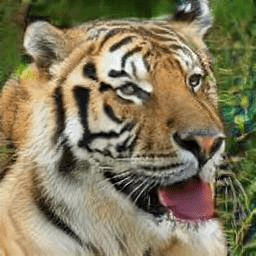} \\
\includegraphics[width=0.13\textwidth]{figure_sup/af_ref/con/4.png} & &
\includegraphics[width=0.13\textwidth]{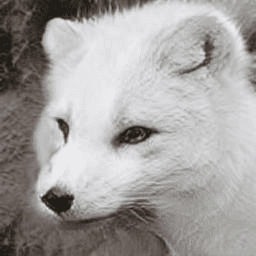}
\includegraphics[width=0.13\textwidth]{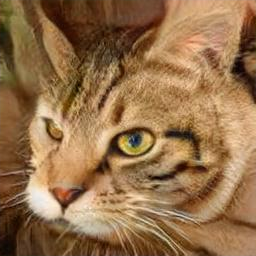}
\includegraphics[width=0.13\textwidth]{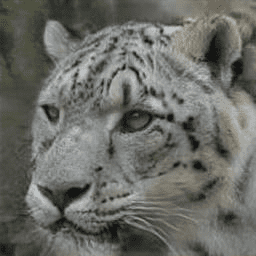}
\includegraphics[width=0.13\textwidth]{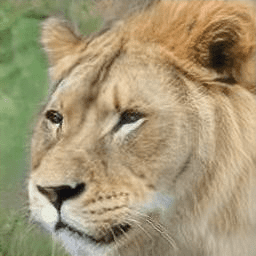}
\includegraphics[width=0.13\textwidth]{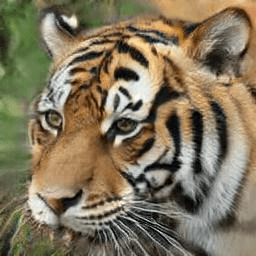} \\
\includegraphics[width=0.13\textwidth]{figure_sup/af_ref/con/5.png} & &
\includegraphics[width=0.13\textwidth]{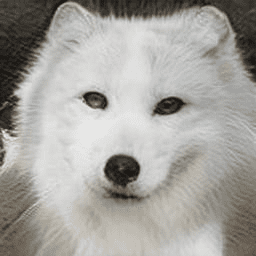}
\includegraphics[width=0.13\textwidth]{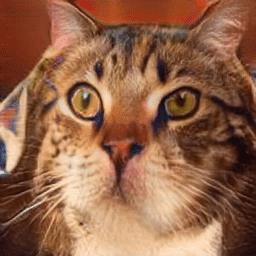}
\includegraphics[width=0.13\textwidth]{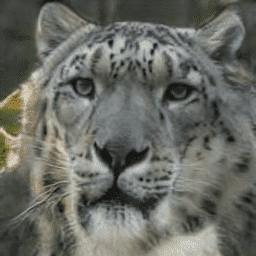}
\includegraphics[width=0.13\textwidth]{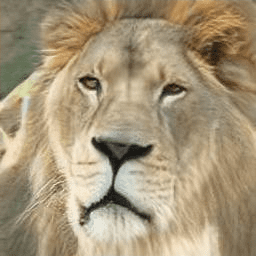}
\includegraphics[width=0.13\textwidth]{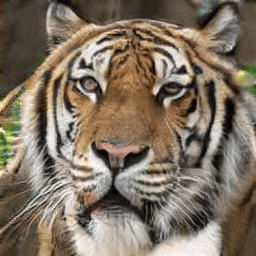} \\[10pt]

\end{tabular}
\end{center}
\vspace{-15pt}
\makebox[1\linewidth][c]{\textbf{Reference-guided image synthesis results}}
\caption{\textbf{Image translation results on Animal Faces-10.} Given domain descriptions are as follows: `appenzeller sennenhund', `beagle', `dandie dinmont terrier', `golden retriever', `malinois', `white fox', `tabby cat', `snow leopard', `lion', `tiger'.}
\label{fig:add_af}
\end{figure}

\begin{figure}
    \centering
\begin{center}
\begin{tabular}{ccc}
& \hspace{-10pt} \raisebox{0.11\height}{\rotatebox{90}{\hspace{13pt} Style}} 
& \includegraphics[width=0.13\textwidth]{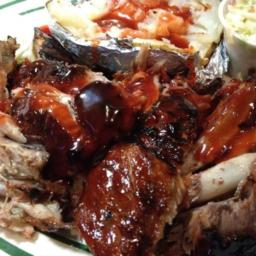}
\includegraphics[width=0.13\textwidth]{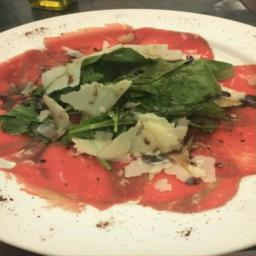}
\includegraphics[width=0.13\textwidth]{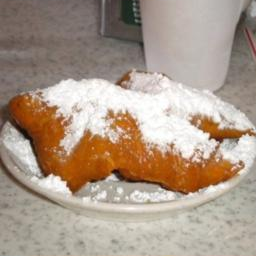}
\includegraphics[width=0.13\textwidth]{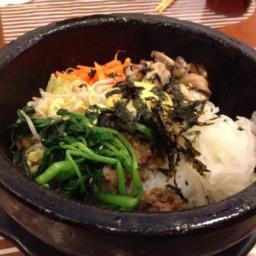}
\includegraphics[width=0.13\textwidth]{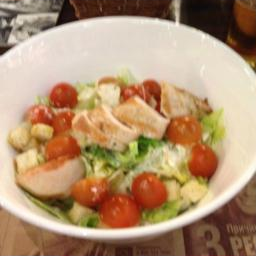} \\[-7pt]
Content \\ \hline \\[-5pt]
\includegraphics[width=0.13\textwidth]{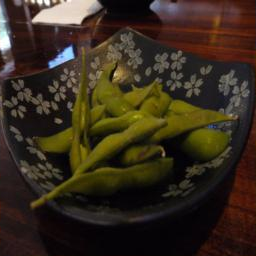} & &
\includegraphics[width=0.13\textwidth]{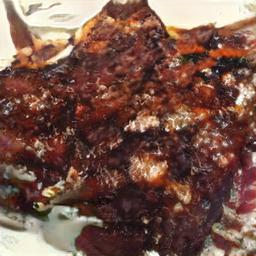}
\includegraphics[width=0.13\textwidth]{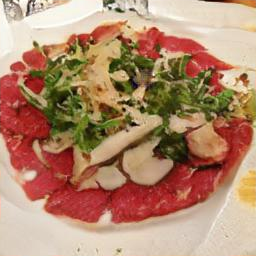}
\includegraphics[width=0.13\textwidth]{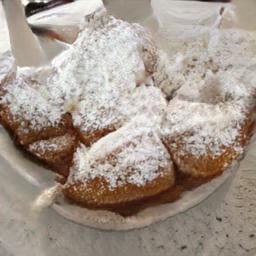}
\includegraphics[width=0.13\textwidth]{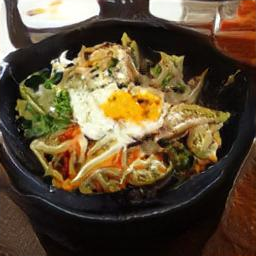}
\includegraphics[width=0.13\textwidth]{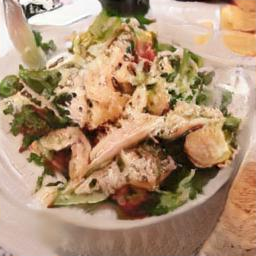} \\
\includegraphics[width=0.13\textwidth]{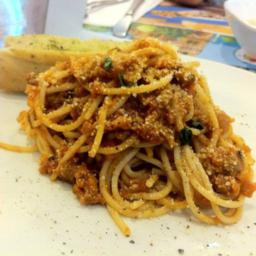} & &
\includegraphics[width=0.13\textwidth]{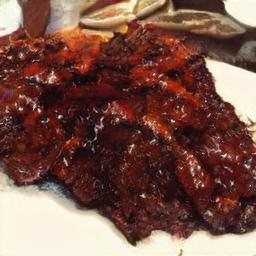}
\includegraphics[width=0.13\textwidth]{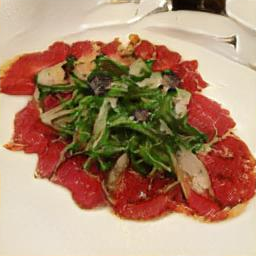}
\includegraphics[width=0.13\textwidth]{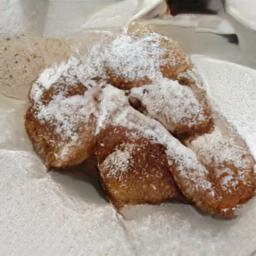}
\includegraphics[width=0.13\textwidth]{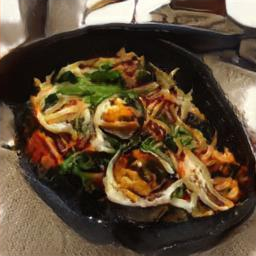}
\includegraphics[width=0.13\textwidth]{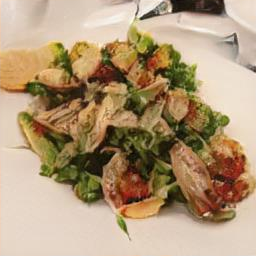} \\
\includegraphics[width=0.13\textwidth]{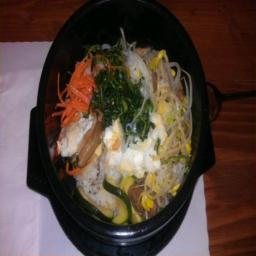} & &
\includegraphics[width=0.13\textwidth]{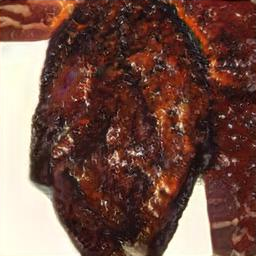}
\includegraphics[width=0.13\textwidth]{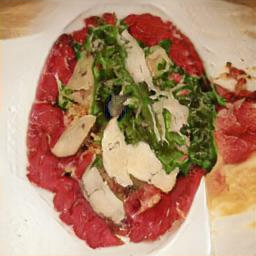}
\includegraphics[width=0.13\textwidth]{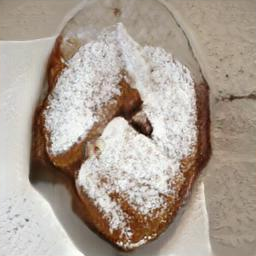}
\includegraphics[width=0.13\textwidth]{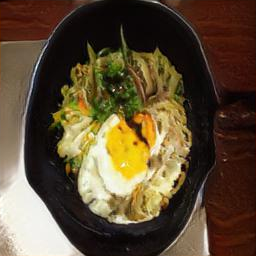}
\includegraphics[width=0.13\textwidth]{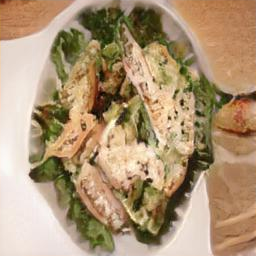} \\[10pt]

& \hspace{-10pt} \raisebox{0.11\height}{\rotatebox{90}{\hspace{13pt} Style}} 
& \includegraphics[width=0.13\textwidth]{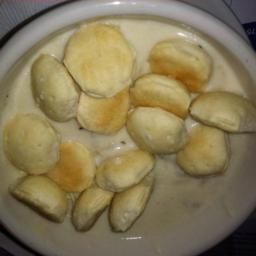}
\includegraphics[width=0.13\textwidth]{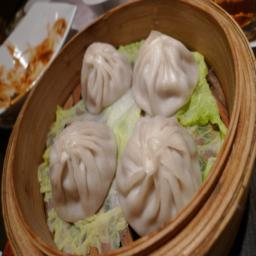}
\includegraphics[width=0.13\textwidth]{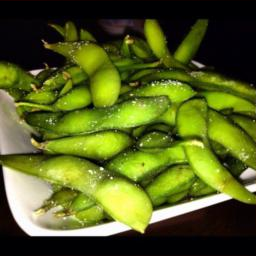}
\includegraphics[width=0.13\textwidth]{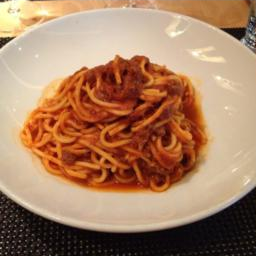}
\includegraphics[width=0.13\textwidth]{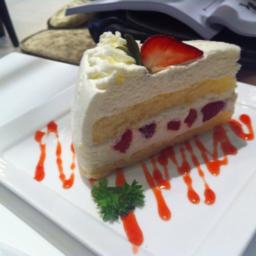} \\[-7pt]
Content \\ \hline \\[-5pt]
\includegraphics[width=0.13\textwidth]{figure_sup/fig6/b/src/food1.png} & &
\includegraphics[width=0.13\textwidth]{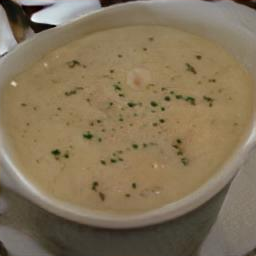}
\includegraphics[width=0.13\textwidth]{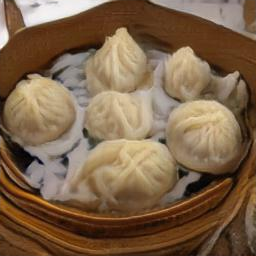}
\includegraphics[width=0.13\textwidth]{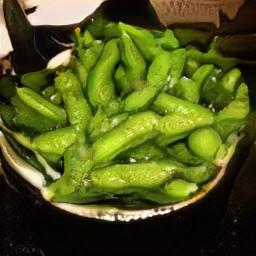}
\includegraphics[width=0.13\textwidth]{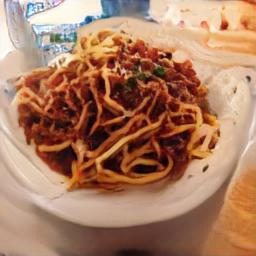}
\includegraphics[width=0.13\textwidth]{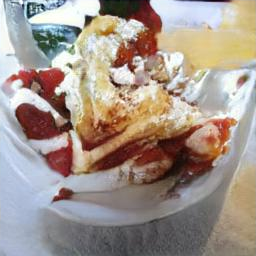} \\
\includegraphics[width=0.13\textwidth]{figure_sup/fig6/b/src/food2.png} & &
\includegraphics[width=0.13\textwidth]{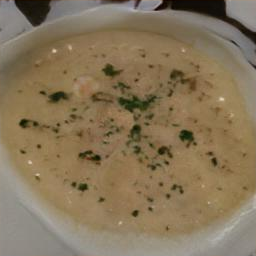}
\includegraphics[width=0.13\textwidth]{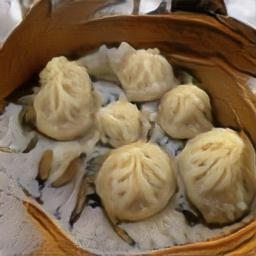}
\includegraphics[width=0.13\textwidth]{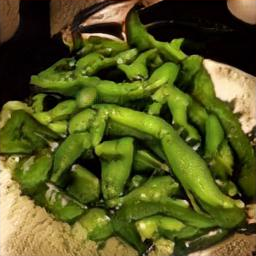}
\includegraphics[width=0.13\textwidth]{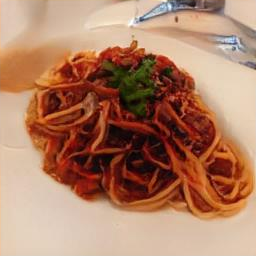}
\includegraphics[width=0.13\textwidth]{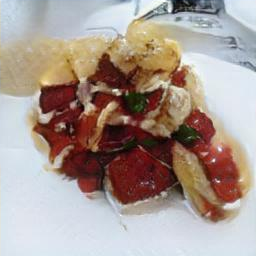} \\
\includegraphics[width=0.13\textwidth]{figure_sup/fig6/b/src/food3.png} & &
\includegraphics[width=0.13\textwidth]{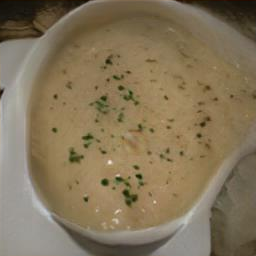}
\includegraphics[width=0.13\textwidth]{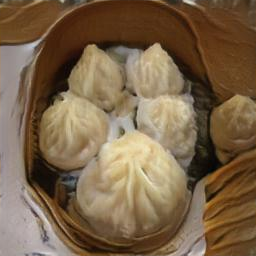}
\includegraphics[width=0.13\textwidth]{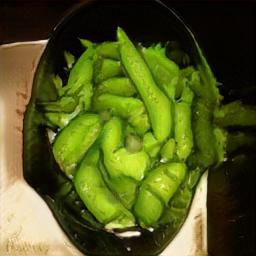}
\includegraphics[width=0.13\textwidth]{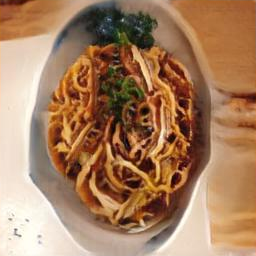}
\includegraphics[width=0.13\textwidth]{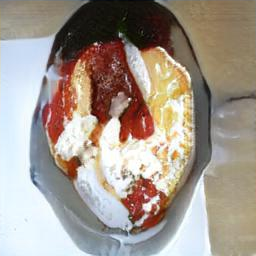} \\[10pt]
\end{tabular}
\end{center}
\vspace{-15pt}
\makebox[1\linewidth][c]{\textbf{Reference-guided image synthesis results}}
\caption{\textbf{Image translation results on Food-10.} Given domain descriptions are as follows: `baby back ribs', `beef carpaccio', `beignets', `bibimbap', `caesar salad', `clam chowder', `dumplings', `edamame', `spaghetti bolognese', `strawberry shortcake'.}
\label{fig:add_food}
\end{figure}

\begin{figure*}
\centering
\newcolumntype{M}[1]{>{\centering\arraybackslash}m{#1}}
\setlength{\tabcolsep}{1pt} 
\renewcommand{\arraystretch}{2} 
\begin{tabular}
{M{0.15\linewidth}M{0.02\linewidth}M{0.15\linewidth}M{0.15\linewidth}M{0.15\linewidth}M{0.15\linewidth}M{0.15\linewidth}}
Content && \makecell[c]{``bang'' \\ ``blond hair'' \\ ``smiling''} & \makecell[c]{ ``bang'' \\ ``black hair'' \\ ``smiling''} & \makecell[c]{``pale skin'' \\ ``rosy cheeks'' \\ ``lipstick''} & \makecell[c]{ ``male'' \\ ``mustach''\\ ``black hair''}  & \makecell[c]{``makeup'' \\ ``pale skin'' \\ ``black hair''} \\ \hline \\[-15pt]
{\includegraphics[width=\linewidth]{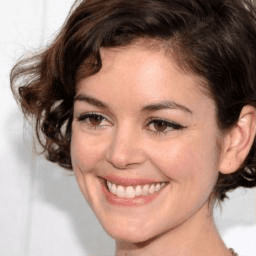}}\hfill & &
{\includegraphics[width=\linewidth]{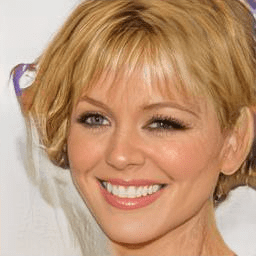}}\hfill  &
{\includegraphics[width=\linewidth]{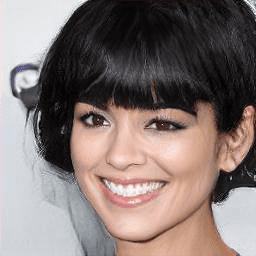}}\hfill  &
{\includegraphics[width=\linewidth]{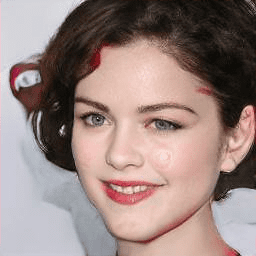}}\hfill  &
{\includegraphics[width=\linewidth]{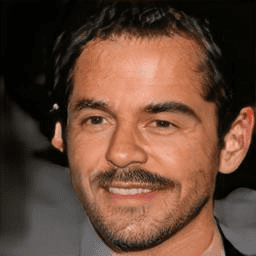}}\hfill  &
{\includegraphics[width=\linewidth]{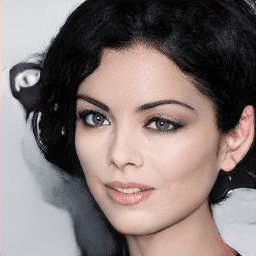}}\hfill \\[-3pt]
{\includegraphics[width=\linewidth]{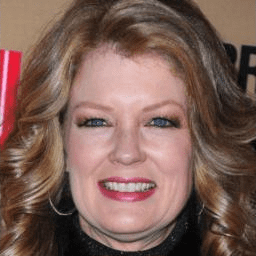}}\hfill & &
{\includegraphics[width=\linewidth]{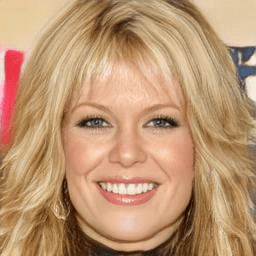}}\hfill  &
{\includegraphics[width=\linewidth]{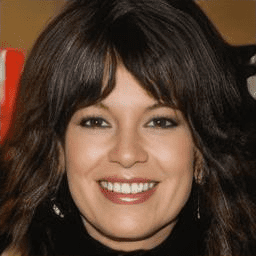}}\hfill  &
{\includegraphics[width=\linewidth]{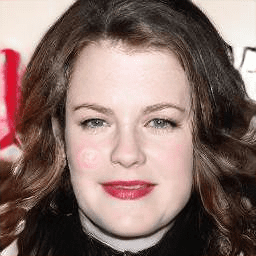}}\hfill  &
{\includegraphics[width=\linewidth]{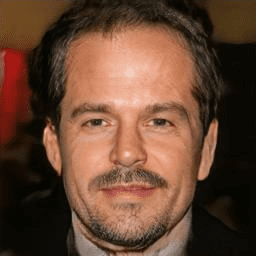}}\hfill  &
{\includegraphics[width=\linewidth]{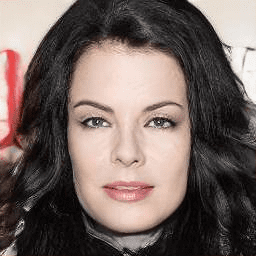}}\hfill \\[-3pt]
{\includegraphics[width=\linewidth]{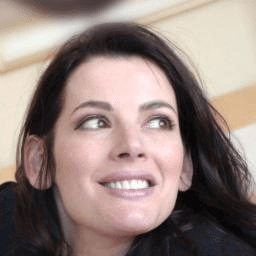}}\hfill & &
{\includegraphics[width=\linewidth]{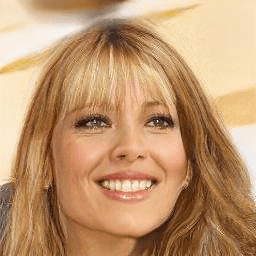}}\hfill  &
{\includegraphics[width=\linewidth]{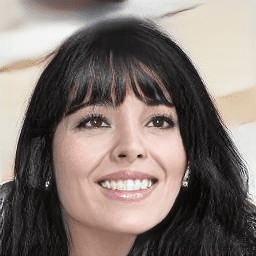}}\hfill  &
{\includegraphics[width=\linewidth]{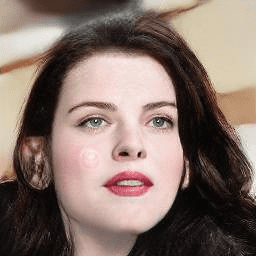}}\hfill  &
{\includegraphics[width=\linewidth]{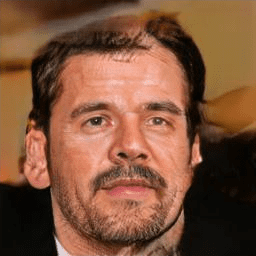}}\hfill  &
{\includegraphics[width=\linewidth]{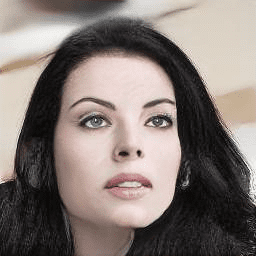}}\hfill  \\[-3pt]
{\includegraphics[width=\linewidth]{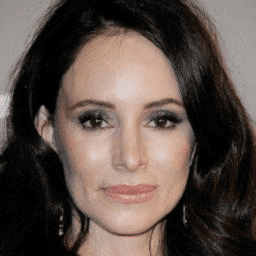}}\hfill & &
{\includegraphics[width=\linewidth]{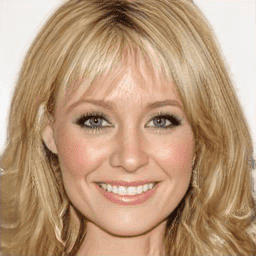}}\hfill  &
{\includegraphics[width=\linewidth]{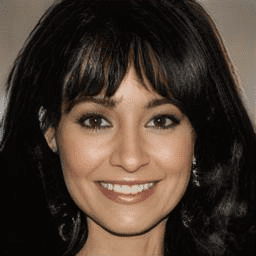}}\hfill  &
{\includegraphics[width=\linewidth]{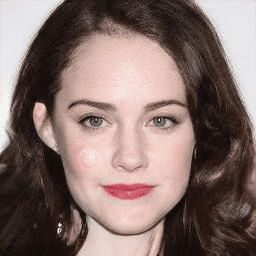}}\hfill  &
{\includegraphics[width=\linewidth]{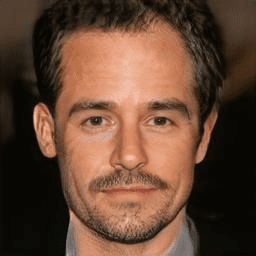}}\hfill  &
{\includegraphics[width=\linewidth]{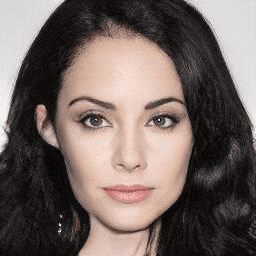}}\hfill  \\[-3pt]
{\includegraphics[width=\linewidth]{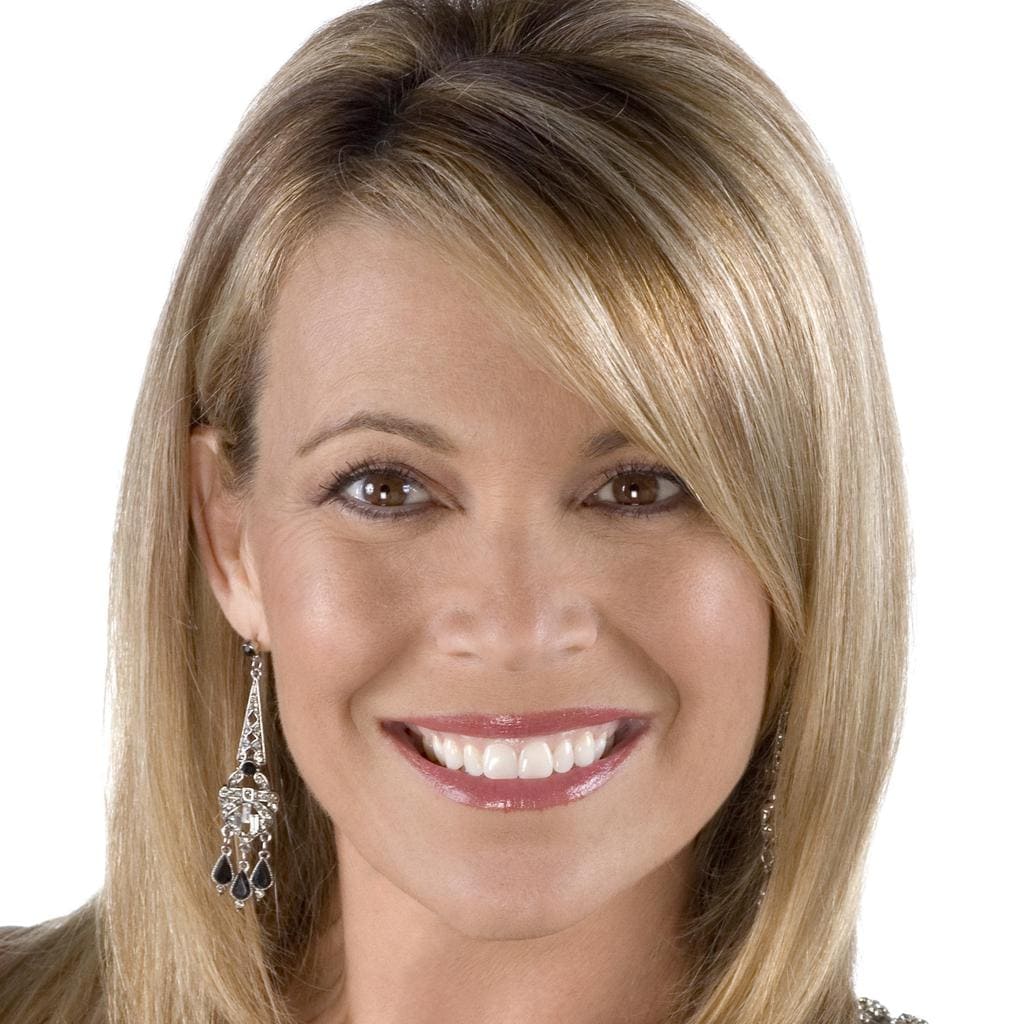}}\hfill & &
{\includegraphics[width=\linewidth]{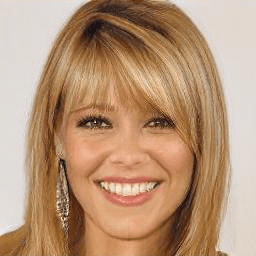}}\hfill  &
{\includegraphics[width=\linewidth]{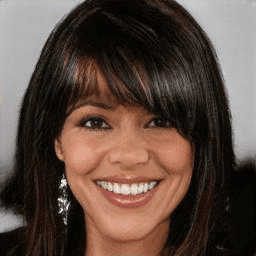}}\hfill  &
{\includegraphics[width=\linewidth]{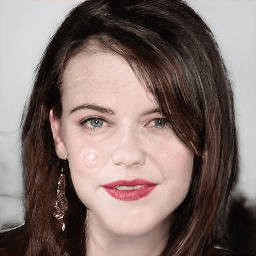}}\hfill  &
{\includegraphics[width=\linewidth]{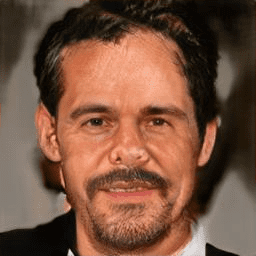}}\hfill  &
{\includegraphics[width=\linewidth]{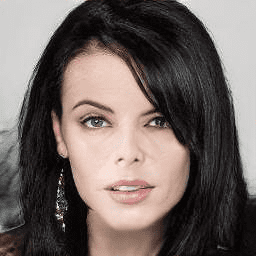}}\hfill  \\[-3pt]
{\includegraphics[width=\linewidth]{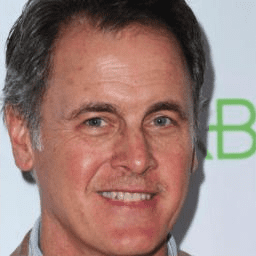}}\hfill & &
{\includegraphics[width=\linewidth]{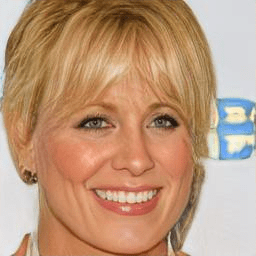}}\hfill  &
{\includegraphics[width=\linewidth]{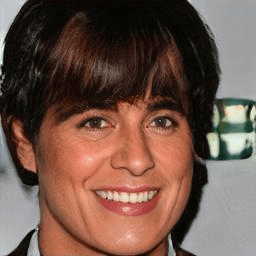}}\hfill  &
{\includegraphics[width=\linewidth]{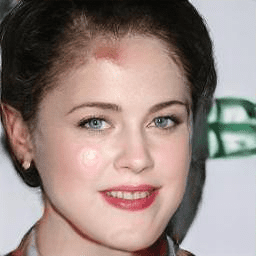}}\hfill  &
{\includegraphics[width=\linewidth]{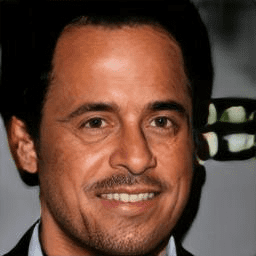}}\hfill  &
{\includegraphics[width=\linewidth]{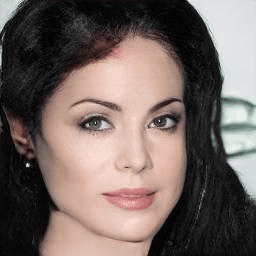}}\hfill  \\[-3pt]
\end{tabular}
    \caption{\textbf{Latent-guided diverse image synthesis results by our LANIT on CelebA-HQ.} Given domain descriptions are as follows: `bangs', `blond hair', `black hair',`smiling', `pale skin', `heavy makeup', `no beard', `rosy cheeks', `wearing lipstick', `male'.}
    \label{fig:add_celeba}
\end{figure*}

\begin{figure}
    \centering
\begin{center}
\begin{tabular}{ccc}
& \hspace{-10pt} \raisebox{0.11\height}{\rotatebox{90}{\hspace{13pt} Style}} 
& \includegraphics[width=0.15\textwidth]{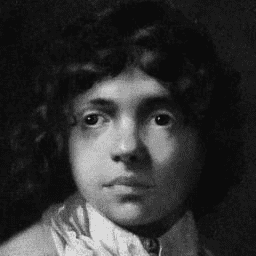}
\includegraphics[width=0.15\textwidth]{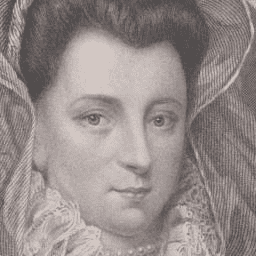}
\includegraphics[width=0.15\textwidth]
{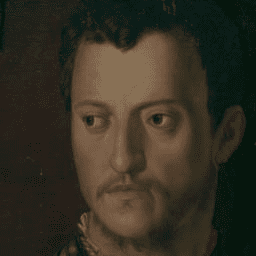}
\includegraphics[width=0.15\textwidth]{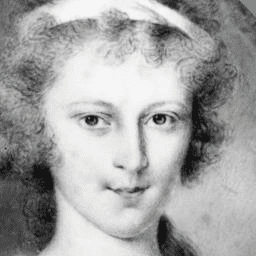}
\includegraphics[width=0.15\textwidth]{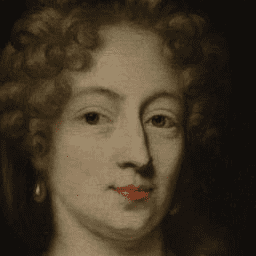} \\[-7pt]
Content \\ \hline \\[-5pt]
\includegraphics[width=0.15\textwidth]{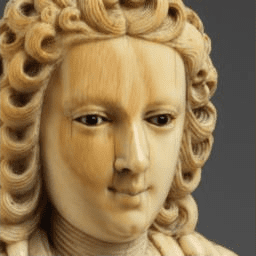} & &
\includegraphics[width=0.15\textwidth]{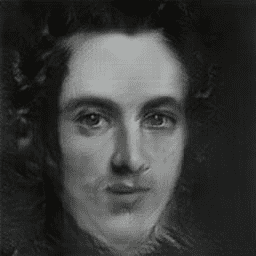}
\includegraphics[width=0.15\textwidth]{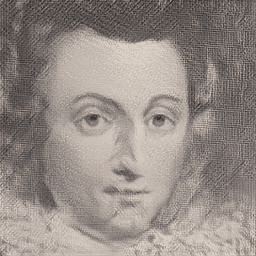}
\includegraphics[width=0.15\textwidth]{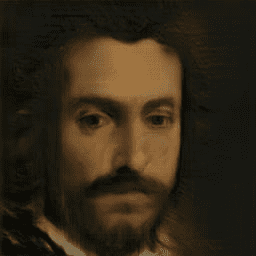}
\includegraphics[width=0.15\textwidth]{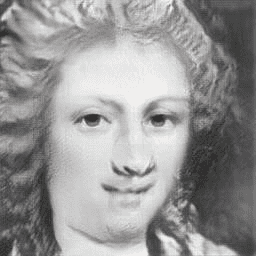}
\includegraphics[width=0.15\textwidth]{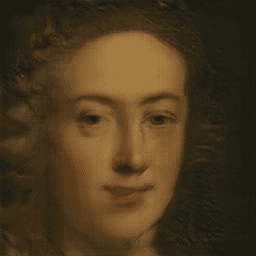} \\
\includegraphics[width=0.15\textwidth]{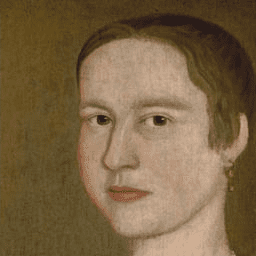} & &
\includegraphics[width=0.15\textwidth]{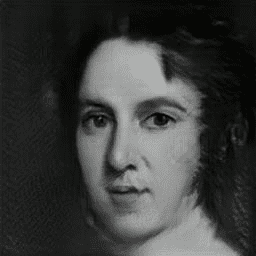}
\includegraphics[width=0.15\textwidth]{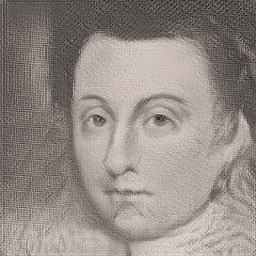}
\includegraphics[width=0.15\textwidth]{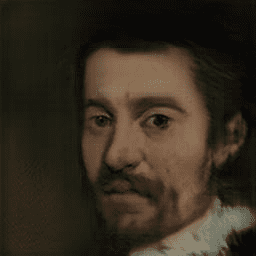}
\includegraphics[width=0.15\textwidth]{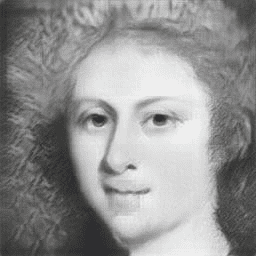}
\includegraphics[width=0.15\textwidth]{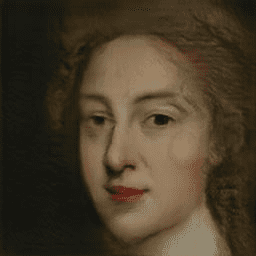} \\
\includegraphics[width=0.15\textwidth]{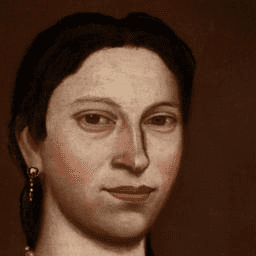} & &
\includegraphics[width=0.15\textwidth]{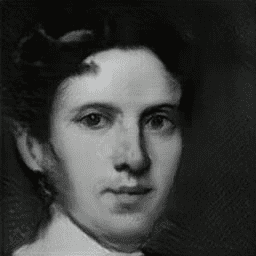}
\includegraphics[width=0.15\textwidth]{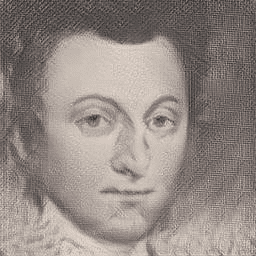}
\includegraphics[width=0.15\textwidth]{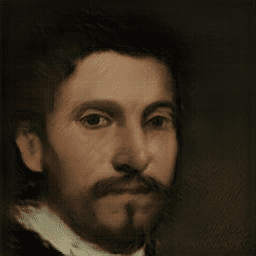}
\includegraphics[width=0.15\textwidth]{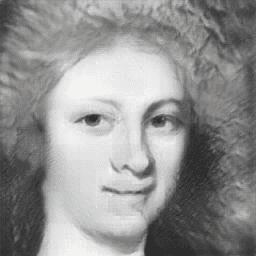}
\includegraphics[width=0.15\textwidth]{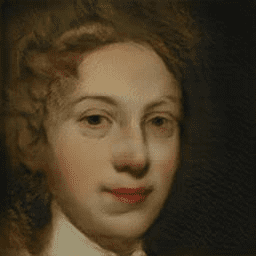} \\
\end{tabular}
\end{center}
    \caption{\textbf{Reference-guided image translation results by our LANIT on MetFace.} Given domain descriptions are as follows: ``oil painting'', ``grayscale'', ``black hair'', ``wavy hair'', ``male'', ``mustache'', ``smiling'', ``gray hair'', ``blonde hair'', ``sculpture''.}
    \label{fig:add_metface}
\end{figure}

\begin{figure}
    \centering
\begin{center}
\begin{tabular}{ccc}
& \hspace{-10pt} \raisebox{0.11\height}{\rotatebox{90}{\hspace{13pt} Style}} 
& \includegraphics[width=0.15\textwidth]{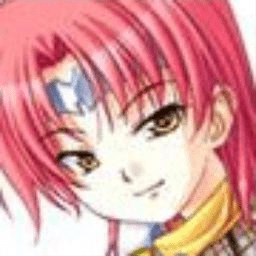}
\includegraphics[width=0.15\textwidth]{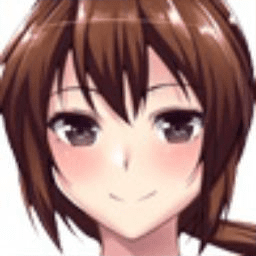}
\includegraphics[width=0.15\textwidth]{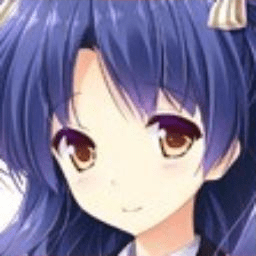}
\includegraphics[width=0.15\textwidth]{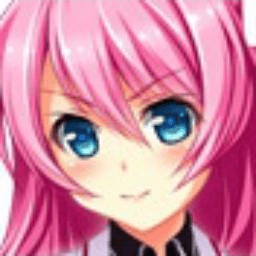}
\includegraphics[width=0.15\textwidth]{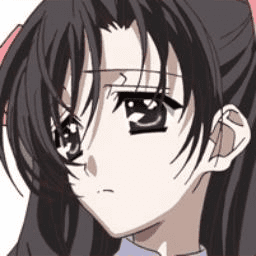} \\[-7pt]
Content \\ \hline \\[-5pt]
\includegraphics[width=0.15\textwidth]{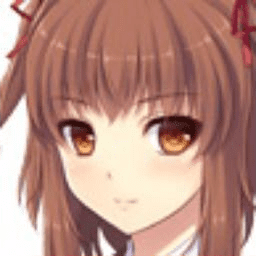} & &
\includegraphics[width=0.15\textwidth]{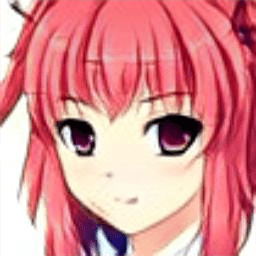}
\includegraphics[width=0.15\textwidth]{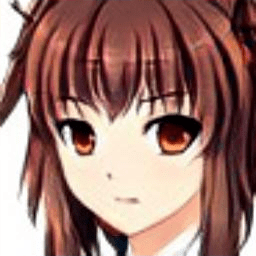}
\includegraphics[width=0.15\textwidth]{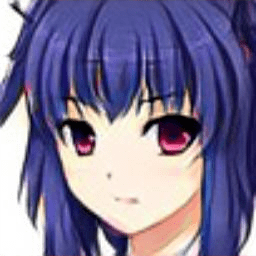}
\includegraphics[width=0.15\textwidth]{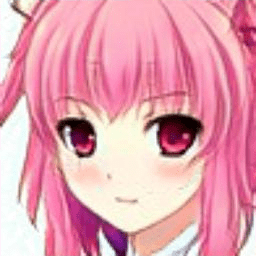}
\includegraphics[width=0.15\textwidth]{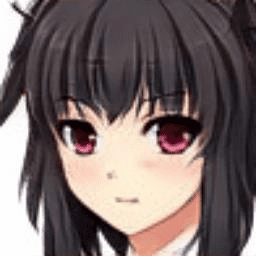} \\
\includegraphics[width=0.15\textwidth]{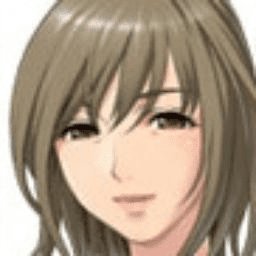} & &
\includegraphics[width=0.15\textwidth]{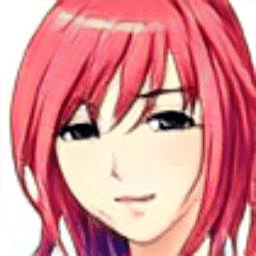}
\includegraphics[width=0.15\textwidth]{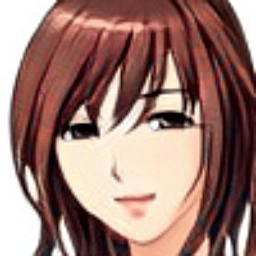}
\includegraphics[width=0.15\textwidth]{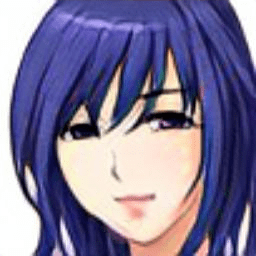}
\includegraphics[width=0.15\textwidth]{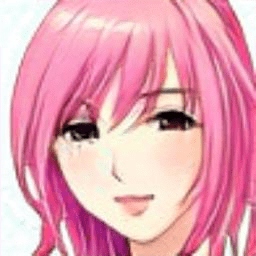}
\includegraphics[width=0.15\textwidth]{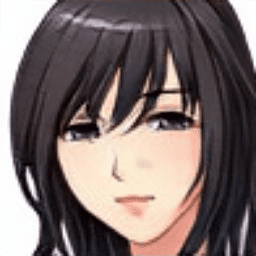} \\
\includegraphics[width=0.15\textwidth]{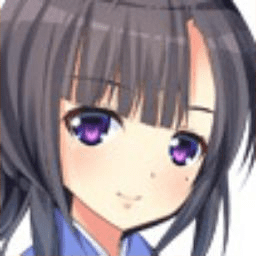} & &
\includegraphics[width=0.15\textwidth]{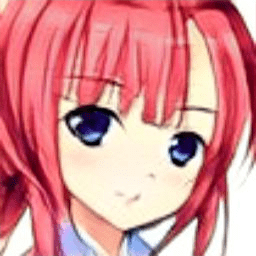}
\includegraphics[width=0.15\textwidth]{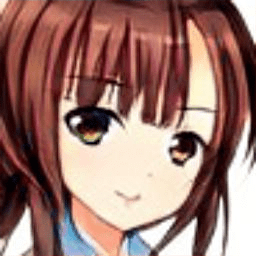}
\includegraphics[width=0.15\textwidth]{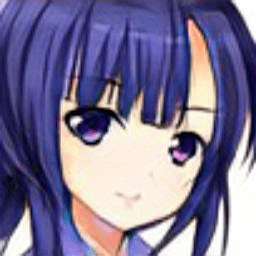}
\includegraphics[width=0.15\textwidth]{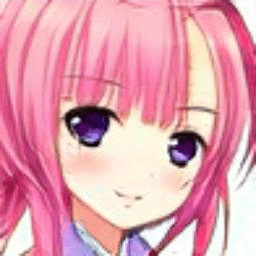}
\includegraphics[width=0.15\textwidth]{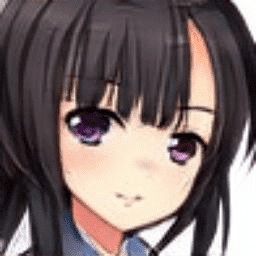} \\
\end{tabular}
\end{center}
\caption{\textbf{Reference-guided image translation results by our LANIT on Anime~\cite{chao2019/online}.} Given domain descriptions are as follows: ``brown hair'', ``red hair'', ``black hair'', ``purple hair'', ``blond hair'', ``blue hair'', ``pink hair'', ``silver hair'', ``green hair'', ``white hair''.}
\label{fig:add_anime}\vspace{-5pt}
\end{figure}

\begin{figure}
    \centering
\begin{center}
\begin{tabular}{ccc}
& \hspace{-10pt} \raisebox{0.11\height}{\rotatebox{90}{\hspace{13pt} Style}} 
& \includegraphics[width=0.15\textwidth]{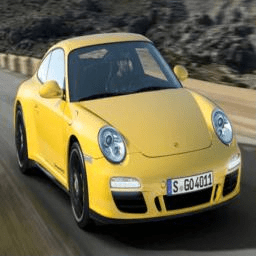}
\includegraphics[width=0.15\textwidth]{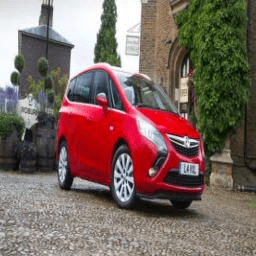}
\includegraphics[width=0.15\textwidth]{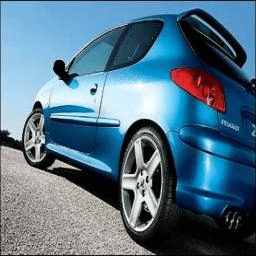}
\includegraphics[width=0.15\textwidth]{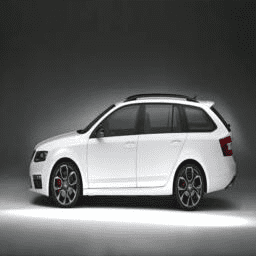}
\includegraphics[width=0.15\textwidth]{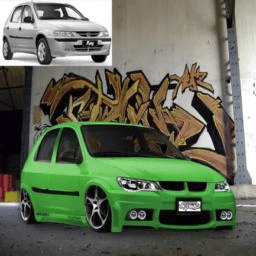} \\[-7pt]
Content \\ \hline \\[-5pt]
\includegraphics[width=0.15\textwidth]{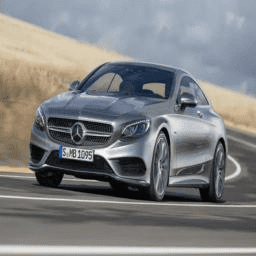} & &
\includegraphics[width=0.15\textwidth]{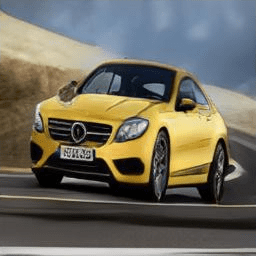}
\includegraphics[width=0.15\textwidth]{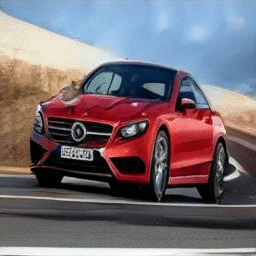}
\includegraphics[width=0.15\textwidth]{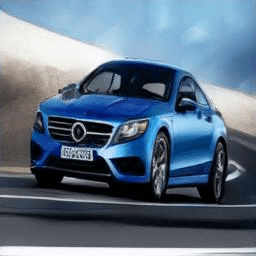}
\includegraphics[width=0.15\textwidth]{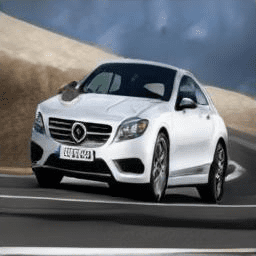}
\includegraphics[width=0.15\textwidth]{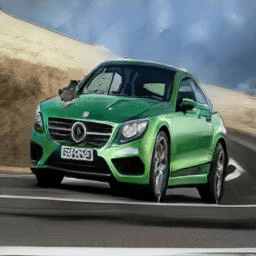} \\
\includegraphics[width=0.15\textwidth]{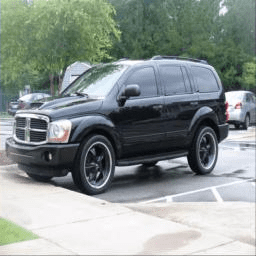} & &
\includegraphics[width=0.15\textwidth]{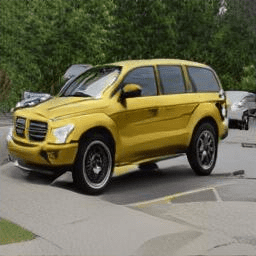}
\includegraphics[width=0.15\textwidth]{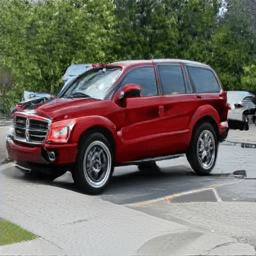}
\includegraphics[width=0.15\textwidth]{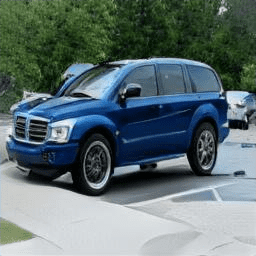}
\includegraphics[width=0.15\textwidth]{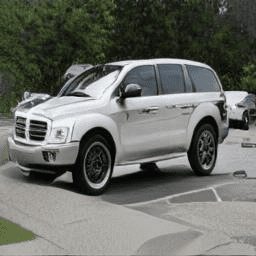}
\includegraphics[width=0.15\textwidth]{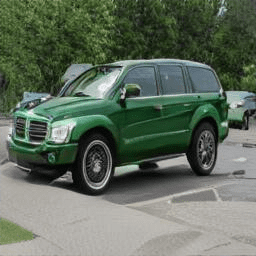} \\
\includegraphics[width=0.15\textwidth]{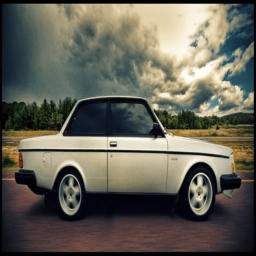} & &
\includegraphics[width=0.15\textwidth]{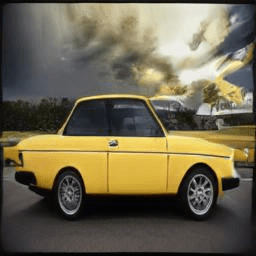}
\includegraphics[width=0.15\textwidth]{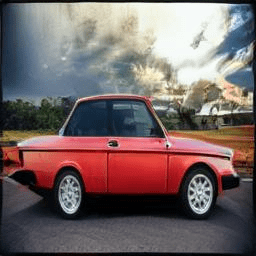}
\includegraphics[width=0.15\textwidth]{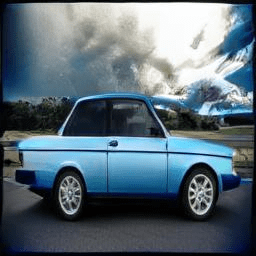}
\includegraphics[width=0.15\textwidth]{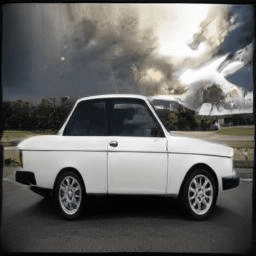}
\includegraphics[width=0.15\textwidth]{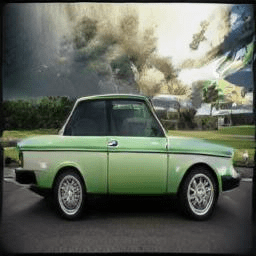} \\
\end{tabular}
\end{center}
\caption{\textbf{Reference-guided image translation results by our LANIT on LSUN-car.} Given domain descriptions are as follows: ``red color'', ``orange color'', `gray color'', ``blue color'', ``yellow color'', ``white color'', ``black color'', ``silver color'', ``green color'', ``pink color''.}
\label{fig:add_lsun}\vspace{-5pt}
\end{figure}

\begin{figure}
    \centering
\begin{center}
\begin{tabular}{ccc}
& \hspace{-10pt} \raisebox{0.11\height}{\rotatebox{90}{\hspace{13pt} Style}} 
& \includegraphics[width=0.15\textwidth]{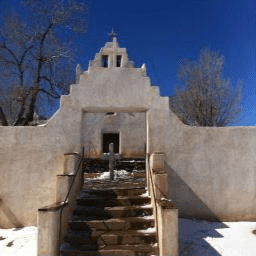}
\includegraphics[width=0.15\textwidth]{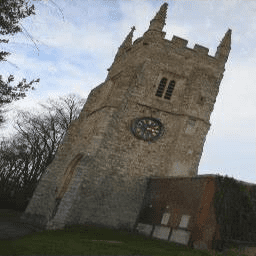}
\includegraphics[width=0.15\textwidth]{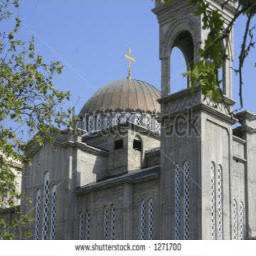}
\includegraphics[width=0.15\textwidth]{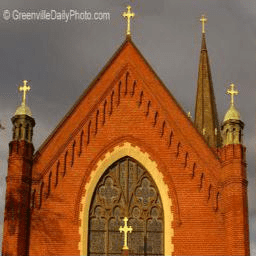}
\includegraphics[width=0.15\textwidth]{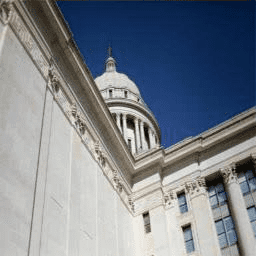} \\[-7pt]
Content \\ \hline \\[-5pt]
\includegraphics[width=0.15\textwidth]{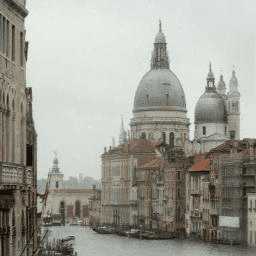} & &
\includegraphics[width=0.15\textwidth]{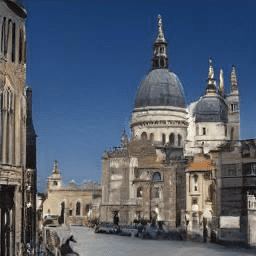}
\includegraphics[width=0.15\textwidth]{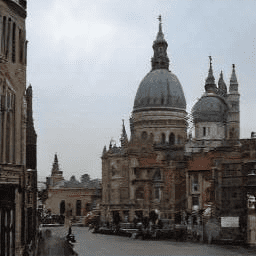}
\includegraphics[width=0.15\textwidth]{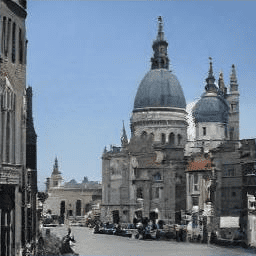}
\includegraphics[width=0.15\textwidth]{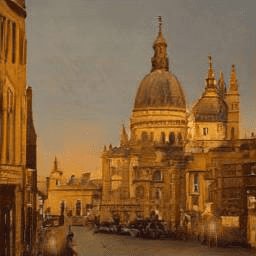}
\includegraphics[width=0.15\textwidth]{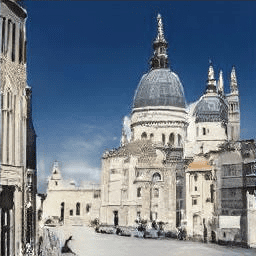} \\
\includegraphics[width=0.15\textwidth]{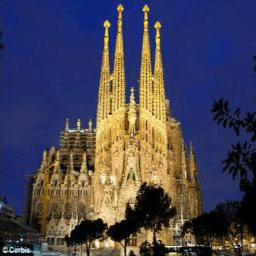} & &
\includegraphics[width=0.15\textwidth]{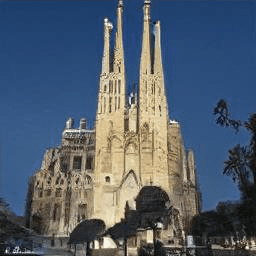}
\includegraphics[width=0.15\textwidth]{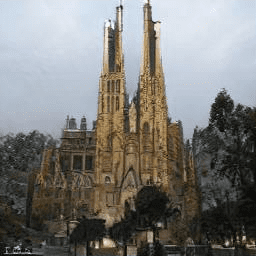}
\includegraphics[width=0.15\textwidth]{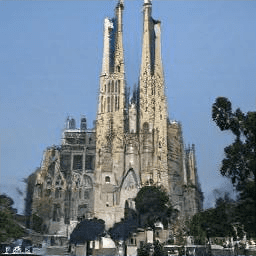}
\includegraphics[width=0.15\textwidth]{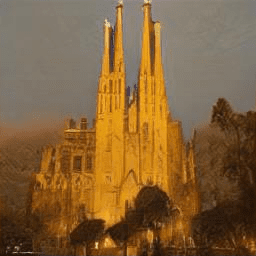}
\includegraphics[width=0.15\textwidth]{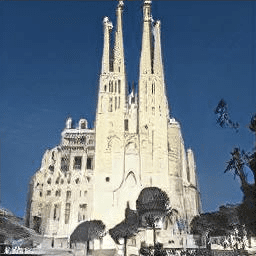} \\
\includegraphics[width=0.15\textwidth]{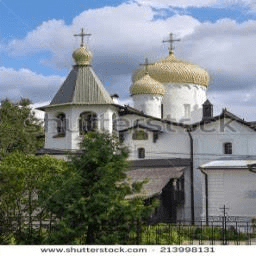} & &
\includegraphics[width=0.15\textwidth]{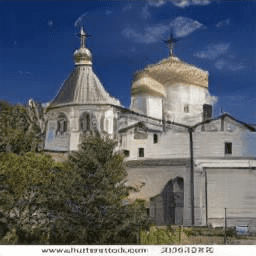}
\includegraphics[width=0.15\textwidth]{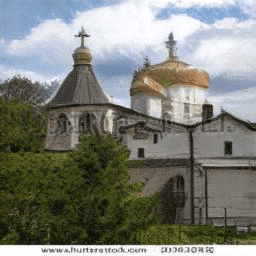}
\includegraphics[width=0.15\textwidth]{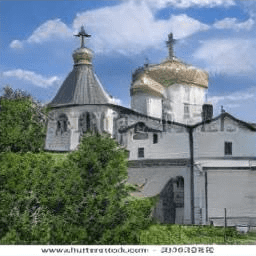}
\includegraphics[width=0.15\textwidth]{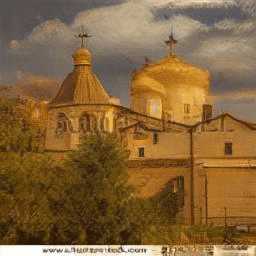}
\includegraphics[width=0.15\textwidth]{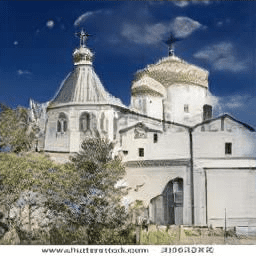} \\
\end{tabular}
\end{center}
\caption{\textbf{Reference-guided image translation results by our LANIT on LSUN-church.} Given domain descriptions are as follows:  ``at night'', ``at sunset'', ``in winter'', ``on cloudy day'', ``on sunny day'', ``with trees'', ``with a river''.}
\label{fig:add_lsun_church}\vspace{-5pt}
\end{figure}

In this section, we visualize additional visual results of our method in \figref{fig:add_af}, \figref{fig:add_food}, \figref{fig:add_celeba}, 
\figref{fig:add_metface}, \figref{fig:add_anime}, \figref{fig:add_lsun}, \figref{fig:add_lsun_church} on Animal Faces-10, Food-10, CelebA-HQ, MetFace, Anime, LSUN-car and LSUN-church datasets~\cite{liu2019few, bossard2014food, liu2015deep, chao2019/online, yu2015lsun}, including 
latent-guided diverse image synthesis and reference-guided image translation results. 
Thanks to our multi-hot labeling setting with the proposed prompt learning technique and adaptive thresholding with the base prompt, our mapping network and style encoder can produce the style vectors faithfully representing target multiple domain styles.

\clearpage

\section{User Study.}
\begin{figure}[h]
\centering
\includegraphics[width=60mm]{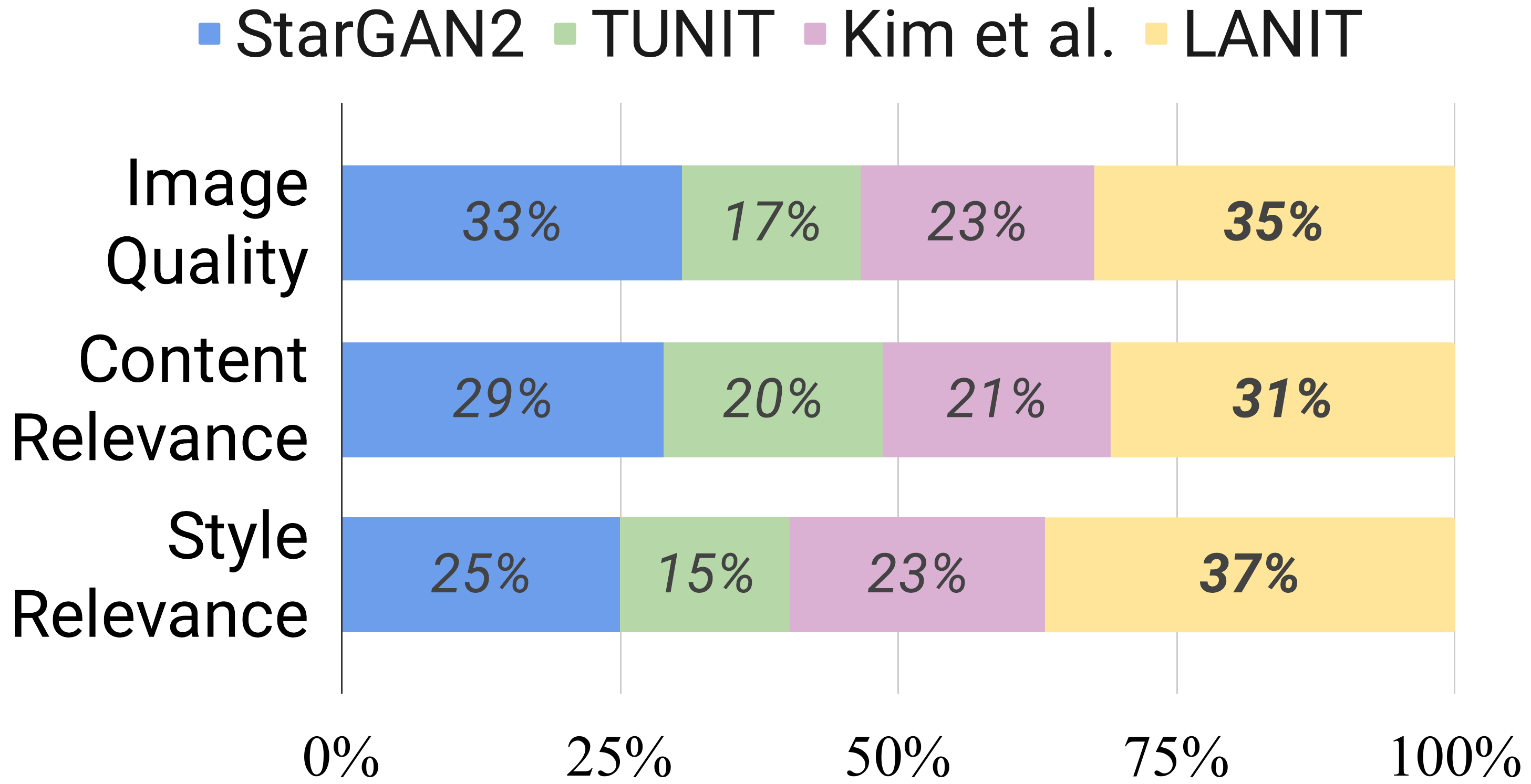}
   \caption{\textbf{User study results.}}
	\label{fig:user_study}\vspace{-10pt}
\end{figure}

Finally, we conducted a user study to evaluate the image quality, content preservation, and style consistency of LANIT compared to StarGAN2, TUNIT, and Kim \textit{et al}. 204 users were involved in this study, asked to answer 60 questions, each of which is generated from randomly sampled 20 images from CelebA-HQ. The examples of questions are as follows: ``Which results do you think have the highest quality/preserve content information such as pose/have similar style of the reference image?" \figref{fig:user_study} shows our LANIT achieves the top rank in all tasks.

\section{Limitations}
Although our LANIT shows outstanding performance on various benchmarks, LANIT inherits some problems from pretrained vision-language models. 
In specific, to overcome this, we suggest adaptive thresholding and prompt learning techniques and these techniques effectively boost the confidence of pseudo labels. However, these techniques still have limitations for getting accurate pseudo labels, which degrades the performance of the image-to-image translation framework.

\end{document}